# Color Fundus Photography Analysis: Co-evolution of Data, Preprocessing, and Modeling toward Multimodal AI

Yu Li[a,1], Wengan He[a,b,*,1], Wenhui Xu[b,1], Lihong Jiang[b,1], Fan Xiao[a], Zhuohang Huang[a], Yuanzhu Liang[a], Jiayi Liu[a], Yuxi Chen[a] and Yongsheng Luo[a,*]

[a]*Zhuhai College of Science and Technology, Zhuhai, Guangdong, China*
[b]*Zhuhai UNO technology Co., Ltd., Zhuhai, Guangdong, China*

ABSTRACT

Color Fundus Photography (CFP) serves as a primary, non-invasive diagnostic modality for high-throughput population screening of ophthalmic and systemic diseases. While existing comprehensive surveys offer valuable perspectives on task-specific algorithms, they remain limited in their analytical integration, often presenting enumerative summaries of datasets and auxiliary preprocessing rather than a systematic synthesis linking these engineering lifecycles to emerging algorithmic paradigms. This review addresses this critical gap by providing an integrated overview of CFP artificial intelligence through the continuous, tri-axial interplay of dataset evolution, preprocessing paradigms, and modeling frameworks. Our analysis reveals that the dataset landscape has expanded from early single-center cohorts with sparse, task-specific categorical labels to massive, multi-center registries characterized by cross-modal pairings and uncurated, heterogeneous longitudinal profiles. Concurrently, CFP preprocessing has transitioned from traditional auxiliary adjustments to neural network-driven data-engineering pipelines, hardware-aware token pruning, and advanced self-supervised tabular imputation for incomplete electronic health records (EHRs). Meanwhile, algorithmic backbones have evolved from inductive-biased local convolutional neural networks (CNNs) to unconstrained global reasoning foundation systems, linearized state space models (SSMs), and cross-modally aligned expert architectures. At the multimodal frontier, cutting-edge systems align intra-image ocular phenotypes with heterogeneous EHRs and longitudinal patient histories, effectively dismantling the systemic visual isolation inherent to unimodal computer vision. Ultimately, this review establishes that the upper bound of clinical diagnostic reliability is determined by collaborative optimization across data lifecycles, rigorous structural preprocessing, and multimodal context synthesis, offering a strategic roadmap to overcome unresolved challenges in cross-domain generalization and resource-constrained edge deployment.

## 1. Introduction

Fundus diseases are among the leading causes of blindness worldwide [1]. Epidemiological studies indicate that diseases such as diabetic retinopathy (DR), glaucoma, and age-related macular degeneration (AMD) continue to rise globally, driven by population aging and the increasing prevalence of diabetes, severely threatening visual health and quality of life [2–4]. The fundus not only serves as a key non-invasive window for observing blood vessels and nerves, but changes in its morphology and function are also closely associated with a range of systemic diseases, including hypertension, diabetes, and chronic kidney disease [1, 5, 6], thereby extending the value of fundus imaging data beyond the field of ophthalmology.

Among various fundus imaging techniques, Color Fundus Photography (CFP) is widely used for basic screening and auxiliary diagnosis due to its simplicity, non-invasiveness, and ability to quickly capture a large field of view (up to 150°-200°)[7, 8]. It has become widely adopted in screening-oriented settings in non-invasive imaging diagnosis systems [5, 6]. Compared to Optical Coherence Tomography (OCT) or Fundus Fluorescein Angiography (FFA), CFP is more cost-effective and easier to perform, making it particularly suitable for population screening in areas with limited medical resources [8]. Furthermore, the feasibility of non-mydriatic CFP using smartphones for DR screening has been validated, further strengthening the advantages of this technology in portable and large-scale screening [9].

*Corresponding author.
*Email address:* wenganhe@icloud.com (W. He)
[1]These authors contributed equally to this work.

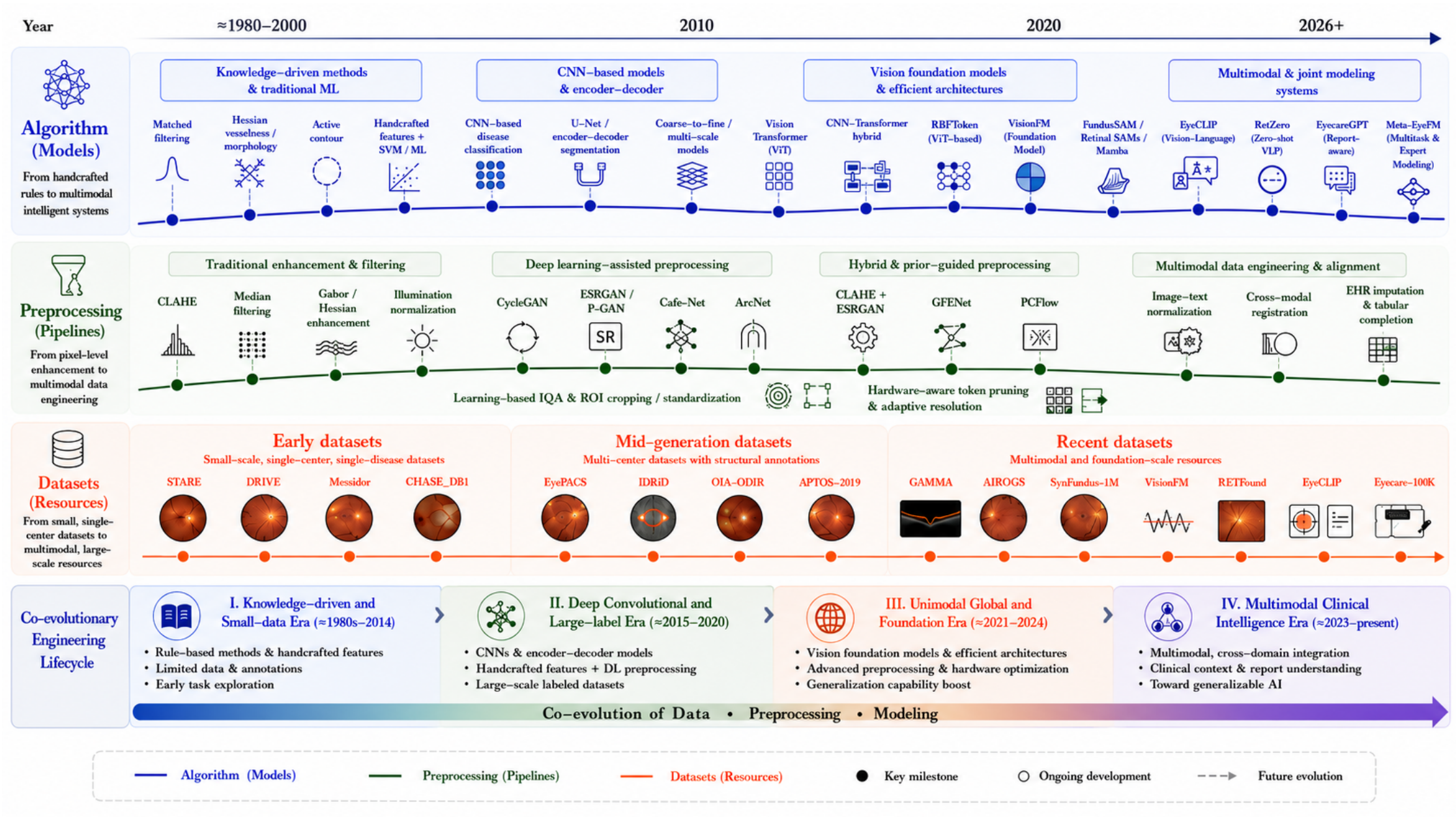

**Figure 1:** Evolution of Fundus Datasets, Preprocessing, and Algorithm.

However, practical challenges in CFP analysis—including low contrast, uneven illumination, color distortion, tissue occlusion, and specular reflection artifacts—often compromise the stability of automated analytical pipelines and the robustness of diagnostic models [10–12]. To surmount these obstacles, researchers have devoted substantial efforts to advancing intelligent CFP analysis by focusing on three core axes: data construction, preprocessing methodologies, and algorithm design. Driven by advances in medical imaging acquisition technology, image processing techniques, and artificial intelligence, substantial progress has been achieved across all three axes. From the perspective of dataset construction, the field has evolved from early small-scale resources primarily characterized by single-disease labels or single-structure annotations (e.g., STARE [13, 14], Messidor[15]) toward datasets featuring the coexistence of multiple annotation types, including image-level, pixel-level, and structure-level annotations (e.g., IDRiD [16], LMOD [17]). Subsequently, metadata and longitudinal follow-up information have been incorporated to capture disease progression and inter-individual variability (e.g., SMDG-19 [18]). Building upon these advances, several recent datasets have further integrated multimodal information, forming large-scale, multi-center, and multimodal data resources (e.g., Eyecare-100K [19], SynFundus-1M [20]), thereby providing more diverse training signals and improving cross-scenario generalization performance. In terms of preprocessing,, methodologies have progressed from traditional filtering and enhancement methods (such as CLAHE [21], median filtering(MF) [12, 22]) to data-driven learning-based approaches (such as CycleGAN [23, 24]), and has further formed hybrid frameworks that combine the advantages of both [25–27], partially mitigating image-quality variations and improving input consistency for downstream models. At the algorithmic level, analytical paradigms have gradually evolved from traditional machine learning [28, 29] to deep models represented by convolutional neural networks (CNNs) [30, 31] and Vision Transformer (ViT) [32], and have further expanded to foundational segmentation models (such as Segment Anything Model (SAM) [33]), multimodal large models including large image-based multimodal models (LMMs)) [34] and vision-language models (VLMs)[19, 20, 35]. These advancements have improved reported performance on several benchmark tasks such as disease classification, detection and segmentation, while also making notable progress in interpretability, interactivity, and clinical adaptability. The evolution of CFP data, preprocessing methods, and algorithm models along these three axes is illustrated in Figure 1.

With respect to the aforementioned topics, existing review studies have evolved from task-specific summaries of conventional CFP analysis methods to broader surveys of deep learning-based and multimodal CFP analysis. Early work by Fraz et al. [36] focused primarily on retinal blood vessel segmentation, systematically reviewing traditional vessel extraction methodologies in two-dimensional CFPs, including filtering-based, model-based, tracking-based, and early machine learning approaches. Although this survey provided an important methodological foundation for CFP analysis, its scope was largely restricted to vessel segmentation and predated the rapid development of modern CNNs, Vision Transformers, and foundation models. Subsequently, Litjens et al. [37] reviewed the rise of deep learning in medical image analysis more broadly, covering classification, detection, segmentation, registration, and other tasks across different medical imaging modalities. However, because this review was not specifically centered on CFP, it did not provide a detailed synthesis of fundus-specific datasets, preprocessing pipelines, disease-oriented tasks, or ophthalmic clinical scenarios. Li et al. [38] further narrowed the focus to CFP analysis and summarized deep learning applications in lesion segmentation, biomarker segmentation, disease diagnosis, and image synthesis, while also organizing a set of publicly available datasets. This work represented a major shift from traditional feature engineering toward data-driven CFP analysis; nevertheless, its discussion was still mainly organized around established deep learning tasks and CNN-dominated paradigms, with limited coverage of more recent developments such as diffusion models, lightweight deployment strategies, prompt-free and prompt-based foundation models, and large vision-language models. More recently, Sheng et al. [39] extended the discussion to multimodal retinal image analysis by categorizing datasets according to imaging modalities and reviewing supported tasks, methods, and evaluation metrics. This survey reflects the increasing importance of multimodal data and benchmarking in retinal AI research. However, its emphasis is mainly on multimodal-in-image retina datasets and task-method organization, rather than on the historical evolution of fundus datasets, the role of preprocessing strategies, or the interaction between data characteristics and algorithmic generalization. In parallel, Dharmaseelan et al. [40] investigated retinal image registration methods from a highly task-specific perspective, comparing several deep learning-based registration pipelines under consistent experimental settings. While such work is valuable for understanding registration performance, its scope is necessarily narrow and does not address broader CFP analysis tasks such as classification, detection, segmentation, synthesis, and foundation-model-based analysis.

Taken together, these reviews show a clear progression from traditional retinal structure analysis to deep learning, fundus-specific applications, multimodal retina analysis, and task-specific registration studies. Despite these contributions, existing reviews remain limited in their analytical integration. At the dataset level, most studies mainly provide enumerative summaries of public CFP datasets, with insufficient synthesis of their evolution in task coverage, annotation granularity, data scale, and modality composition. In particular, recent datasets have expanded from conventional CFP datasets for vessel segmentation, lesion detection, and disease grading to large-scale pretraining datasets, multimodal ophthalmic datasets, and image-text datasets; however, this transition has not been systematically organized in previous reviews. In terms of preprocessing, existing surveys usually mention quality control, normalization, enhancement, cropping, and registration only as auxiliary procedures, with limited discussion of learning-based or neural network-driven preprocessing methods. At the algorithmic level, prior reviews still mainly emphasize CNN-based methods or early Vision Transformer variants, whereas recent paradigms such as self-supervised learning, contrastive learning, diffusion models, foundation models, and vision-language or large medical multimodal models have not been sufficiently synthesized. Therefore, a unified review linking dataset evolution, preprocessing strategies, and algorithmic development remains lacking. Our review addresses this gap by providing an integrated synthesis of CFP analysis with particular attention to datasets, preprocessing, and emerging algorithmic paradigms.

To address the aforementioned research gaps, this review provides a systematic overview of advances in deep learning for CFP analysis from three complementary perspectives: datasets, preprocessing, and algorithms. It examines the intrinsic relationships among these components and analyzes their combined impact on model generalization and clinical applicability. Specifically, this review establishes a structured review framework spanning data, preprocessing, and algorithmic dimensions, tracing the developmental trajectories of relevant methods and highlighting representative works. In addition, by synthesizing existing studies, it summarizes and distills several unresolved common challenges in the field, including multi-center variability and annotation quality constraints at the data level; the relatively high computational cost of certain preprocessing methods, along with risks of artifact introduction and over-enhancement at the preprocessing level; and persistent limitations at the algorithmic level related to cross-domain generalization, uncertainty modeling, multimodal fusion, and engineering-oriented deployment. These analyses aim to provide reference directions for future research and practical translation into real-world clinical settings.

The remainder of this review is organized as follows. Section 2 presents a stage-wise review of CFP datasets, systematically examining the characteristics and limitations of early, intermediate, and contemporary datasets, and outlining an evolutionary framework for dataset development. Section 3 provides a comprehensive review of preprocessing techniques for CFP, categorizing them into traditional preprocessing methods, deep learning–assisted

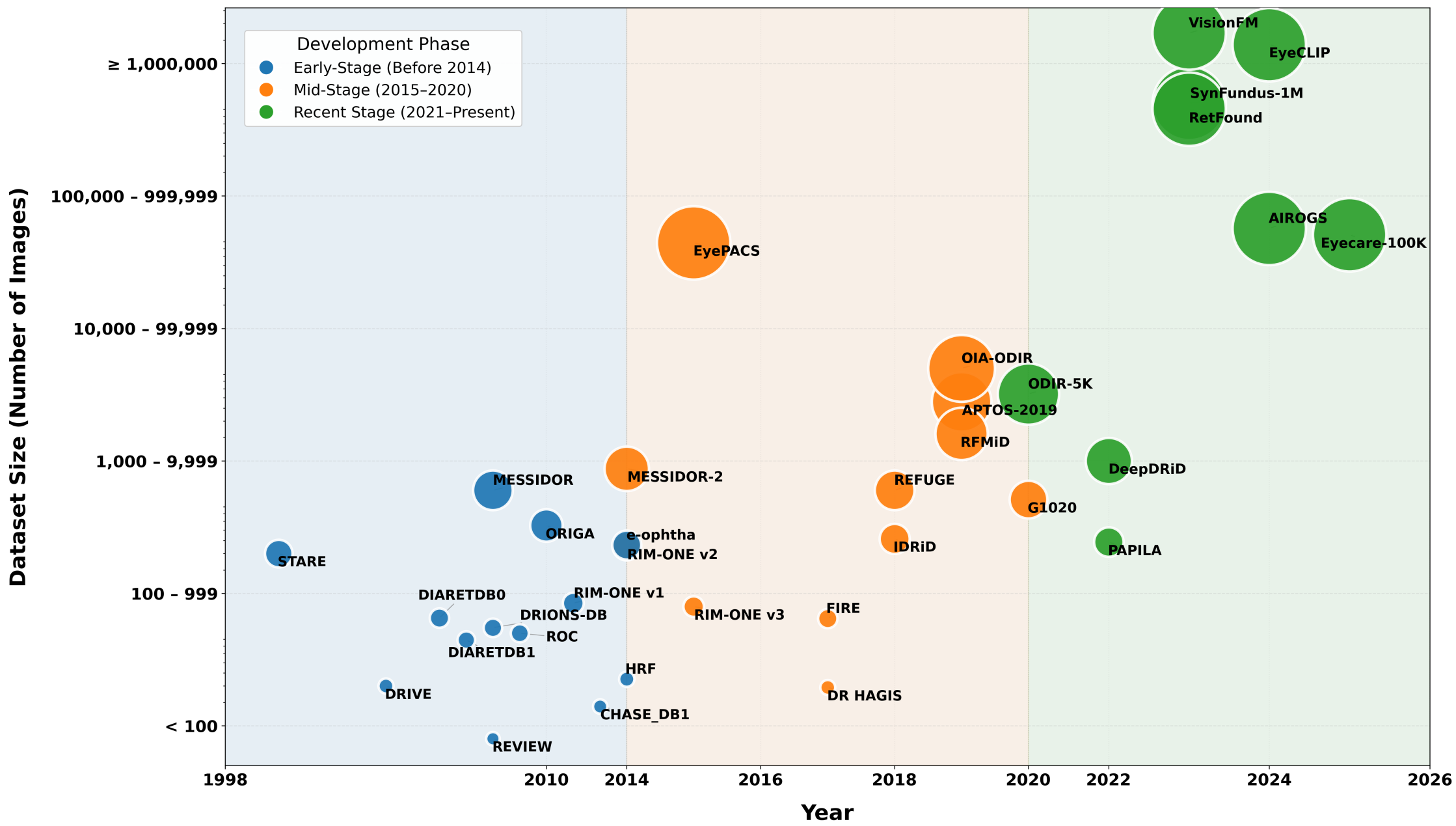

*Notes: The horizontal axis denotes the year in which each dataset was made publicly available, while the vertical axis represents the dataset size (number of images). Each circle corresponds to a fundus image dataset; the color indicates its development phase, and the circle size is proportional to the dataset scale.*

**Figure 2:** Evolution of Fundus Datasets.

preprocessing approaches, and hybrid preprocessing strategies that integrate traditional and deep learning techniques; the characteristics and limitations of each category are analyzed, and their impacts on classification, detection, and segmentation tasks are evaluated. Section 4 reviews the evolution of algorithmic models, ranging from early machine learning methods to the application of CNNs, Transformers, and multimodal learning models, and discusses their influence on CFP analysis as well as future development trends. Finally, Section 5 summarizes the review and outlines directions for future research.

## 2. Datasets

As a core non-invasive imaging modality in ophthalmic diagnosis, the development of intelligent analysis techniques for CFP is highly dependent on the availability of high-quality, large-scale datasets with multi-task annotations. To visually characterize the evolutionary trajectory of public datasets, figure 2 presents a visual summary of representative datasets released between 1998 and 2025, with time on the horizontal axis and dataset scale on the vertical axis. This visualization highlights the expansion of datasets from hundreds to hundreds of thousands of images and the accompanying shift in task formulation from single-structure segmentation to disease grading, multi-task learning, and multimodal integration.

Within this overarching framework, we further summarize the detailed attributes of publicly available datasets in Table 1 (the complete dataset list is shown in Appendix 1), including sample size, task type, annotation content, and label format, thereby providing a detailed reference basis for the subsequent stage-wise analysis.

**Table 1**
Summary of the Dataset (Representative subset).

| No. | Dataset | Size | Task | Annotation | Label | Year | Ref |
|---|---|---|---|---|---|---|---|
| 1 | STARE | 400 | Vessel segmentation, multi-disease classification | Vessel contour; DR, glaucoma, and other diseases (13 classes) | Image-level, Pixel-level | 2000 | [13, 14] |
| 2 | DRIVE | 40 | Vessel segmentation | Vessel contour | Pixel-level | 2004 | [28] |
| 3 | Messidor | 1.2K | DR grading, DME risk assessment | DR grade; DME risk | Image-level | 2008 | [15] |
| 4 | DIARETDB0 | 130 | DR classification, lesion detection | MA, HE, EX, SE references; DR label | Image-level | 2006 | [41, 42] |
| 5 | DIARETDB1 | 89 | DR classification, lesion detection | MA, HE, EX, SE references; DR label | Polygon-level, Image-level | 2007 | [42, 43] |
| 6 | CHASE_DB1 | 28 | Vessel segmentation | Vessel contour | Pixel-level | 2012 | [44] |
| 7 | HRF (High-Resolution Fundus) | 45 | Vessel segmentation, OD localization | Vessel contours; OD location and diameter | Pixel-level, Bounding box-level (circular) | 2013 | [45] |
| 8 | Diabetic Retinopathy Detection (EyePACS/Kaggle) | 88,696 | DR grading | Five-level DR grades | Image-level | 2015 | [46] |
| 9 | FIRE (Fundus Image Registration Dataset) | 129 | Image registration | Corresponding control point pairs | Point-level | 2017 | [47] |
| 10 | DR HAGIS | 39 | Vessel segmentation | Vessel contours | Pixel-level | 2017 | [48] |
| 11 | IDRiD (Indian Diabetic Retinopathy Image Dataset) | 516 | Lesion segmentation, OD segmentation, fovea detection, DR grading; DME grading | MA, HE, EX, SE contours, OD contour, disc center, fovea localization, DR and DME grades | Image-level, Pixel-level, Point-level | 2018 | [16] |
| 12 | APTOS-2019 Blindness Detection | 5,590 | DR grading | Five-level DR grades | Image-level | 2019 | [49] |
| 13 | OIA-ODIR | 10,000 | Multi-disease classification | Normal, DR, glaucoma, cataract, AMD, HR, myopia, and other diseases | Image-level | 2019 | [50] |
| 14 | AIROGS (Artificial Intelligence for Robust Glaucoma Screening) | 113,893 | Glaucoma grading | Referable glaucoma, non-referable glaucoma, ungradable | Image-level | 2024 | [51] |
| 15 | GAMMA (Glaucoma Grading from Multi-Modality Images Challenge) | 300 pairs (fundus + OCT) | Glaucoma grading; OD and OC segmentation; fovea localization | Normal, early glaucoma, progressive glaucoma; fovea coordinates; OD and OC contours | Image-level, Pixel-level, Point-level | 2022 | [52] |
| 16 | OCTA-Fundus-DR | 222 (111 OCTA and 111 fundus) | DR grading | No DR, mild DR, moderate DR | Image-level | 2023 | [53] |
| 17 | SynFundus-1M | 1,000,000 | Foundation model pretraining; multi–disease diagnosis | Natural language diagnostic descriptions | Image-text pair; Image-level | 2023 | [20] |
| 18 | Eyecare-100K | 102,000 image–text pairs | Visual question answering; report generation; diagnosis and grading; lesion localization | Classification/grading labels; QA pairs | Image-text pair | 2025 | [19] |
| 19 | EyeCLIP Pretraining Data | 2,777,593 images ; 11,180 reports | Vision–language pretraining | Mostly unlabeled images; angiography reports with structured keywords | Image-text pair | 2024 | [35] |
| 20 | RETFound Pretraining Data | Fundus: 904,170; OCT: 736,442 | Retinal foundation model pretraining | Unlabeled | None | 2023 | [54] |
| 21 | VisionFM Pretraining Data | 3,396,918 | General ophthalmic foundation model pretraining | Unlabeled | None | 2023 | [55] |

*Abbreviations:* DME, diabetic macular edema; HR, hypertensive retinopathy; MA, microaneurysms; HE, hemorrhages; EX, exudates; SE, soft exudates; OD, optic disc; OC, optic cup.

A synthesis of the datasets summarized in Table 1 reveals pronounced differences across different time periods along three key dimensions: sample scale, task objectives, and annotation strategies. Early datasets were generally limited in size and focused on pixel-level annotations of basic anatomical structures, such as retinal vessels. Since 2015, driven by the substantially increased demand for large-scale training data in deep learning methods, fundus image datasets have rapidly expanded to the order of tens of thousands of images. During this period, task emphasis gradually shifted toward image-level annotations centered on disease screening and grading, while multi-disease recognition and finer-grained lesion- and structure-level annotations began to be introduced. In more recent years, these trends have further accelerated, with datasets increasingly emphasizing broader multi-disease coverage, multi-source information integration, and applications in more complex clinical scenarios, forming the basis for the stage-wise analysis presented in the following sections.

## 2.1. Development of Datasets

Based on the stage-wise characteristics of dataset development, fundus image datasets are categorized into three periods: the early stage (before 2014), the intermediate stage (2015 to 2020), and the recent stage (2021 to the present).

### *2.1.1. Early Stage*

Fundus image datasets released prior to 2014 are characterized by small sample sizes, single-structure annotations, and a primary focus on single diseases. These early repositories served as foundational validation resources for traditional image processing and early machine learning algorithms, with research tasks heavily concentrated on retinal vessel segmentation, optic disc localization, and preliminary screening for diabetic retinopathy (DR). Figure 3 presents representative examples from the DRIVE dataset, providing an intuitive illustration of the original fundus image appearance and the corresponding manually annotated vessel masks characteristic of this stage. Most data repositories from this foundational stage were derived from isolated screening programs or single hospital-based settings, resulting in constrained population coverage and centralized demographic distributions. In terms of disease scope, the data landscape primarily gravitated toward DR—exemplified by cohorts such as Messidor [15], DIARETDB0 [41, 42], and DIARETDB1 [42, 43]—while a distinct subset emphasized vascular structure modeling and geometric analysis, including STARE [13, 14], DRIVE [28], CHASE_DB1 [44], and HRF [45].

Regarding annotations, early datasets predominantly relied on manual labeling by a small group of ophthalmic experts, ensuring basic reliability through double reading or multi-expert consensus strategies. However, annotation formats varied widely across different repositories—encompassing bitmap masks, circular bounding boxes, and point-based coordinate labels—which introduced severe engineering friction for cross-dataset transfer learning and joint multi-cohort training. Furthermore, systematic alignment with contemporary unified clinical standards, such as the International Clinical Diabetic Retinopathy (ICDR) scale or the Early Treatment Diabetic Retinopathy Study (ETDRS) guidelines, was largely absent, and structured patient-level clinical metadata (e.g., BMI, blood glucose levels, and blood pressure longitudinal tracks) were rarely integrated. Data cleaning procedures were also seldom explicitly documented, typically relying implicitly on basic image quality thresholds inherent to clinical acquisition devices, with only a minimal fraction enforcing rigorous quality filtering or field-of-view standardization. To preserve the analytical fluidity of this section while providing exhaustive structural parameters—including specific sample sizes, resolution ranges, and modality configurations—a comprehensive matrix detailing the individual profiles of these foundational datasets is synthesized in Appendix 2. Synthesizing this multi-era paradigm shift, while these early data repositories played a crucial pioneering role in structural modeling and early task exploration, they exhibited severe inherent limitations in sample scale, disease and structural coverage, annotation consistency, clinical information availability, and data standardization.

### *2.1.2. Intermediate Stage*

As the architectural advantages of convolutional neural networks (CNNs) in medical image analysis became increasingly evident, early datasets designed primarily for localized, small-scale algorithmic validation grew fundamentally inadequate to satisfy the high-capacity training requirements and hunger for statistical consistency inherent to deep architectures. Consequently, the period between 2015 and 2020 witnessed a major paradigm shift, marked by the development of intermediate-stage fundus image repositories characterized by accelerated sample volumes, diversified task topologies, and expanded disease coverage. Compared with the preceding era, dataset capacities underwent an order-of-magnitude expansion, transitioning from hundreds of rudimentary images to massive repositories containing tens of thousands of samples, though localized small- and medium-scale cohorts continued to coexist for specialized downstream verification.

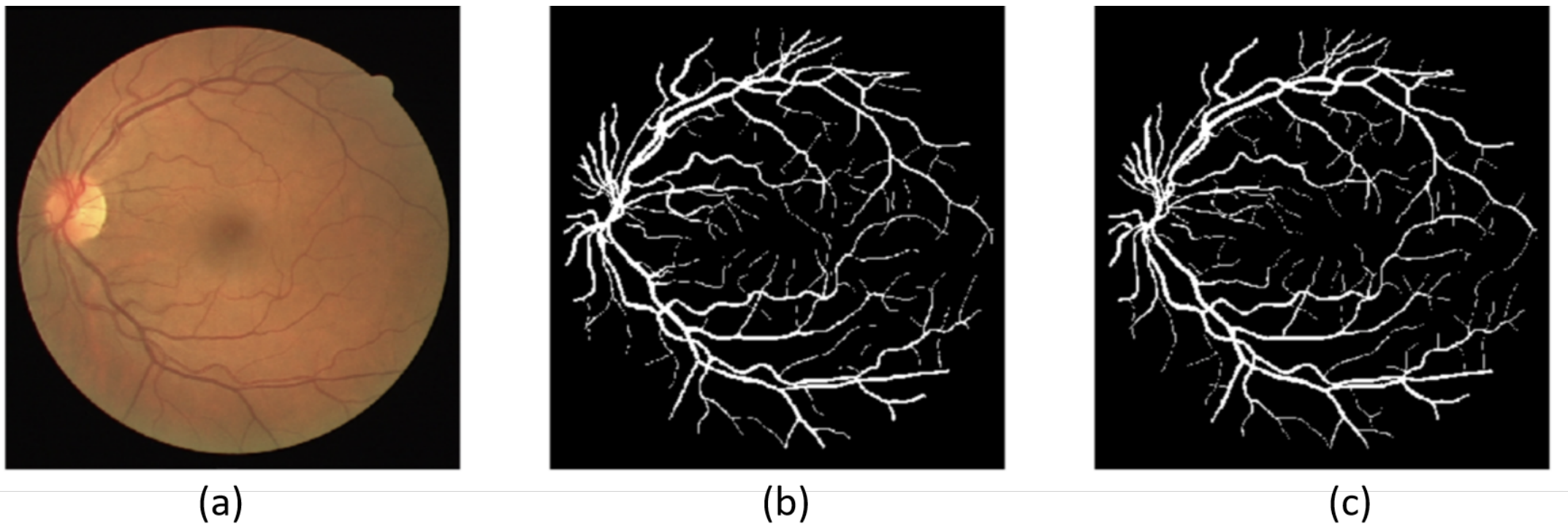

**Figure 3:** Example images from the DRIVE dataset [28]. (a) Original RGB fundus image; (b) Manual vessel annotation by Expert 1; (c) Manual vessel annotation by Expert 2.

In terms of task taxonomies, these intermediate-stage datasets primarily prioritized image-level classification and staging to provide robust data foundations for screening and diagnostic deep networks. Concurrently, more granular annotation schemes tailored for object detection, localized bounding, and pixel-level parsing were progressively integrated into mainstream registries. For instance, the IDRiD dataset [16] advanced beyond standard diagnostic classification by offering both pixel-level lesion segmentation masks and spatial coordinates for target detection of vital anatomical landmarks, such as the macular center, as illustrated in Fig.4. Meanwhile, highly specialized, task-specific cohorts were strategically retained to benchmark niche mathematical challenges, including DR HAGIS [48] for automated vascular tortuosity analysis and FIRE [47] for multimodal retinal image registration. Regarding clinical coverage, while diabetic retinopathy (DR) remained the predominant diagnostic pivot—with pivotal cohorts like APTOS-2019 [49] and EyePACS [46] providing extensive, multi-class severity grading (see Fig.5)—the data scope began transitioning from single-disease isolation toward multi-disease diagnostic formulations. This multi-label evolution is exemplified by the OIA-ODIR dataset [50], which comprehensively encapsulates diverse ocular pathologies across a single cohort, including pathological myopia, glaucoma, cataracts, and comorbid diabetic conditions (see Fig.6).

For a highly structured architectural breakdown of this transitional era, including granular cross-tabulations of sensor modalities, sample scales, and multi-task annotation properties, readers are referred to the exhaustive taxonomy matrix synthesized in Appendix 3. Overall, while this intermediate epoch marked a critical evolutionary leap from unconstrained methodological exploration to large-scale data support for deep learning, severe limitations persisted regarding inter-observer annotation consistency, the availability of holistic clinical metadata, and cross-dataset evaluation standardization. These systemic constraints introduced latent domain shifts and generalization bottlenecks, explicitly laying the groundwork for the subsequent era of generalized, strictly standardized, and cross-modally aligned registries.

### *2.1.3. Recent Stage*

Since 2021, the construction of retinal fundus image datasets has undergoes a radical paradigm shift, transitioning from isolated image repositories into highly complex, multidimensional, and interconnected data organization ecosystems. Driven by the training demands of foundational AI models, real-world clinical datasets from this contemporary epoch exhibit unprecedented expansion in terms of sample scale, demographic representation, and cross-center heterogeneity. Strikingly, population screening cohorts such as AIROGS [51] have scaled beyond the critical threshold of one hundred thousand samples for the first time, consolidating data across more than 500 global screening centers to mitigate geographical bias and enhance cross-domain resilience. Concurrently, the programmatic configuration of disease taxonomies has evolved from single-condition isolation toward dense multi-disease diagnostic matrices. This multi-label maturation is epitomized by the EyeCLIP dataset [35], which encompasses 17 distinct ophthalmic and systemic conditions within a unified framework, effectively facilitating the learning of shared pathological representations.

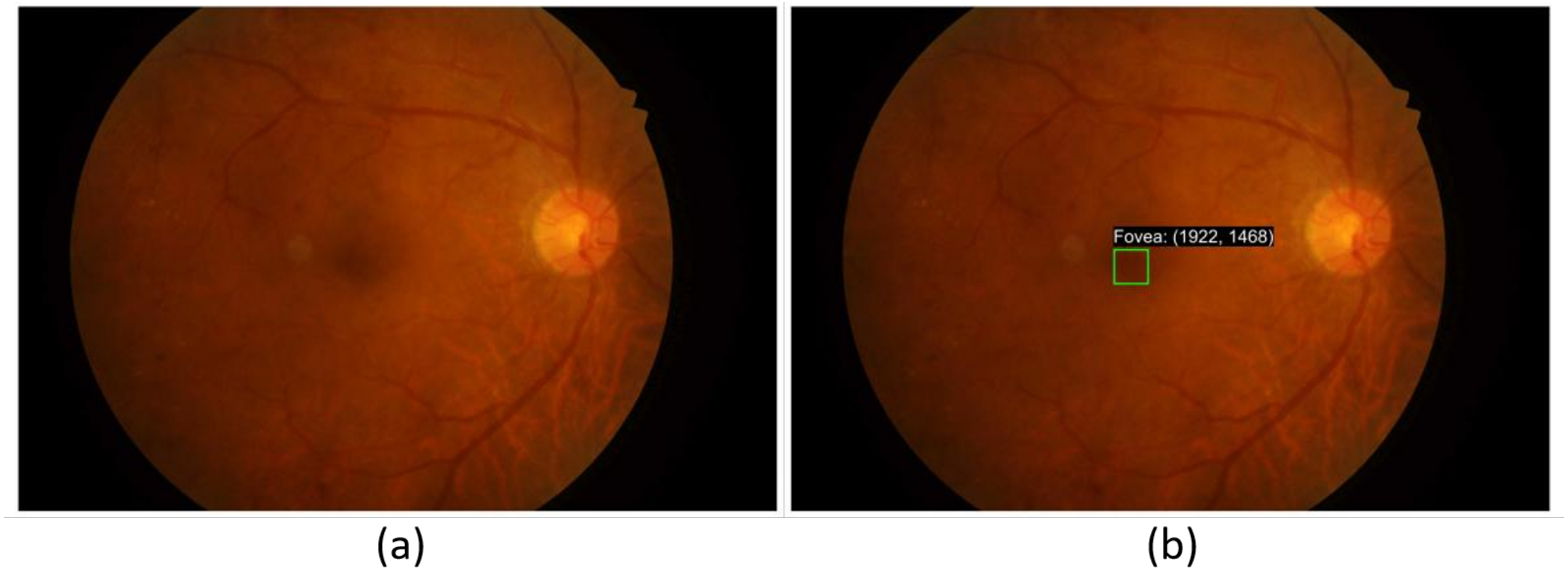

**Figure 4:** Example of foveal center detection in the IDRiD dataset [16]. (a) Original; (b) Annotated.

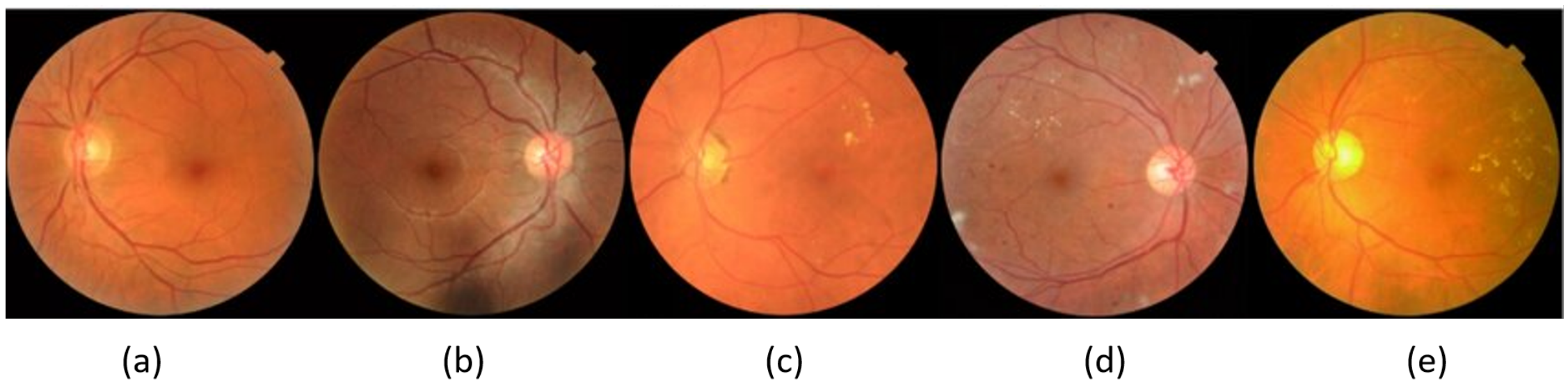

**Figure 5:** Examples from the EyePACS dataset [46]. (a) No DR; (b) Mild; (c) Moderate; (d) Severe; (e) PDR (Proliferative DR).

To dismantle the systemic information bottlenecks and visual isolation inherent to single-modality clinical reasoning, the contemporary frontier has witnessed an explosion of multimodal and cross-modally aligned registries. These advanced data ecosystems are split into two primary integration trajectories: first, the co-indexed synthesis of 2D retinal surface topographies with 3D structural tomographies, such as optical coherence tomography (OCT), optical coherence tomography angiography (OCTA), and fundus fluorescein angiography (FFA), as represented by the OCTA-Fundus-DR [53] and GAMMA [52] cohorts; second, the strategic pairing of high-resolution images with unstructured textual data—including clinical narratives, expert diagnostic reports, and longitudinal electronic health records (EHRs)—as exemplified by Eyecare-100K [19], RETFound [54], and VisionFM [55]. This structural shift promotes the evolution of ophthalmic data foundations toward multi-task generative paradigms (see Fig.7 and Fig.8). Furthermore, generative synthetic and pretraining-oriented data generators, such as SynFundus-1M [20], have emerged to alleviate the acute scarcity of rare pathological tokens, providing entirely new synthesized data channels for self-supervised pretraining.

Correspondingly, at the data curation level, recent registries have instituted highly formalized, automated processing pipelines encompassing image quality artifact screening, advanced deduplication, and multi-tier cryptographic anonymization. However, the introduction of non-image textual data has introduced severe methodological vulnerabilities. While multimodal frameworks have begun integrating natural language processing (NLP) filters for textual denoising and structural normalization, these pipelines remain predominantly ad-hoc and heuristic. Regarding annotations and systemic quality control, while these datasets enforce multi-expert consensus and rigorous arbitration mechanisms to sustain basic visual label reliability, extending such quantitative quality control to non-image modalities remains an unresolved bottleneck. Constrained by the immense semantic complexity of medical text and prohibitive annotation overheads, current frameworks rely heavily on sampling-based manual reviews or downstream outcome-level validation, lacking the systematized, quantitative inter-observer consistency metrics that standardize contemporary image annotations. Moreover, according to publicly available documentation, a substantial majority of recent cohorts only state in principle that their annotations adhere to widely accepted clinical conventions, whereas only a select few explicitly delineate their standard criteria, such as the strict adherence of OCTA-Fundus-DR to the Early Treatment Diabetic Retinopathy Study (ETDRS) guidelines.

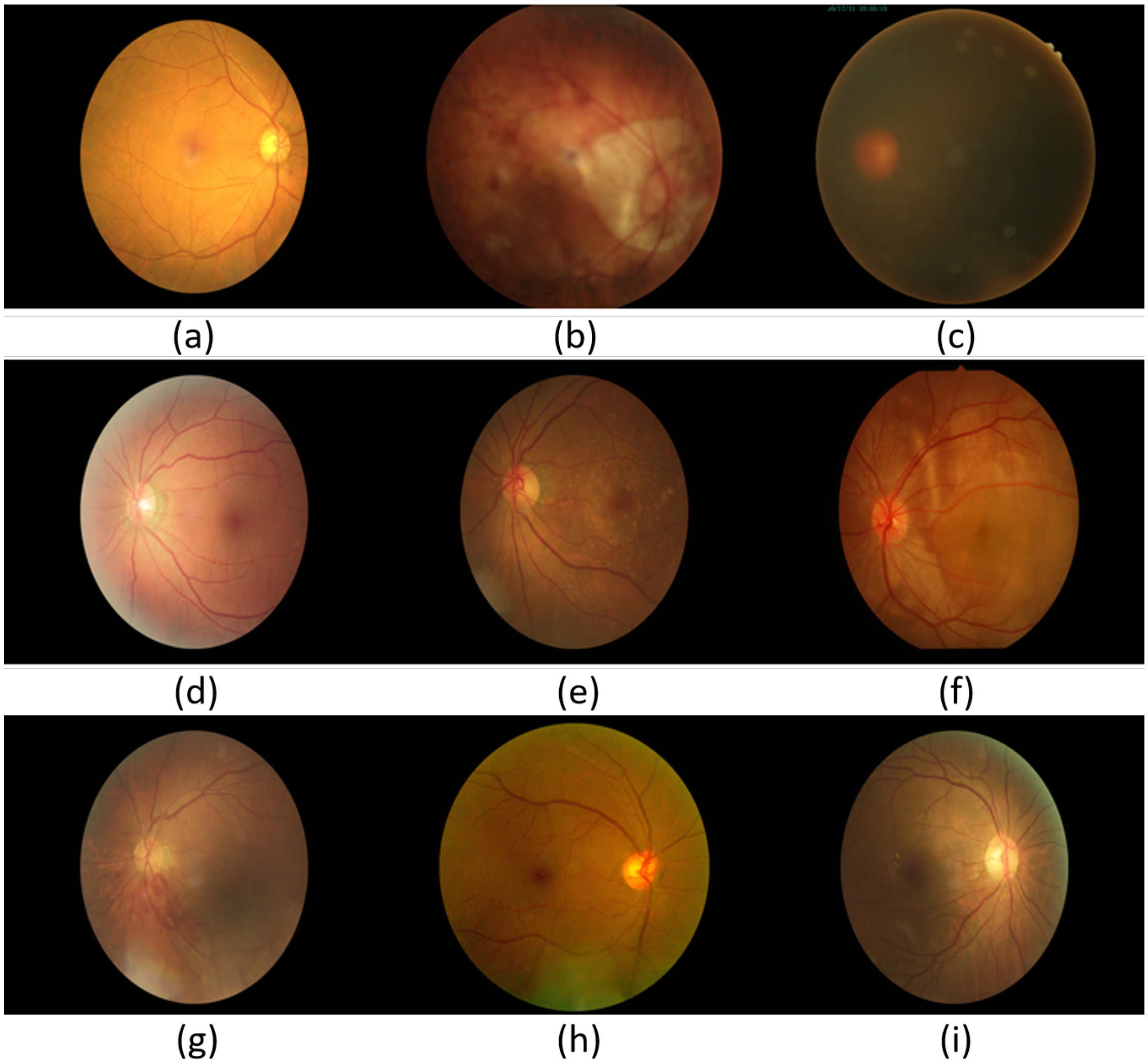

**Figure 6:** Examples of fundus image classification in the OIA-ODIR datase [50]. (a) Normal; (b) Pathological Myopia; (c) Cataract; (d) DR; (e) AMD; (f) Central Retinal Artery Occlusion; (g) Branch Retinal Artery Occlusion; (h) Glaucoma; (i) HR.

To bridge the conceptual synthesis of this cutting-edge landscape with its exhaustive engineering parameters, a comprehensive multidimensional metadata matrix encapsulating sample scale, modal combinations, and clinical standard designations across these recent registries is cataloged in Appendix 4. Taken together, the continuous, tri-axial evolution of dataset landscapes from pre-2014 small-scale structural validation, through the 2015-2020 deep learning data support, and culminating in the post-2021 multimodal foundation ecosystems, underscores a fundamental truth in ophthalmic artificial intelligence: the performance upper bound of downstream diagnostic systems is intrinsically governed by the scale, structural diversity, and clinical standardization of the underlying data foundation. This evolutionary trajectory establishes that resolving cross-modal uncertainty propagation and enforcing hardware-aware data-engineering pipelines represent the critical prerequisites for the next generation of generalizable, deployable expert architectures.

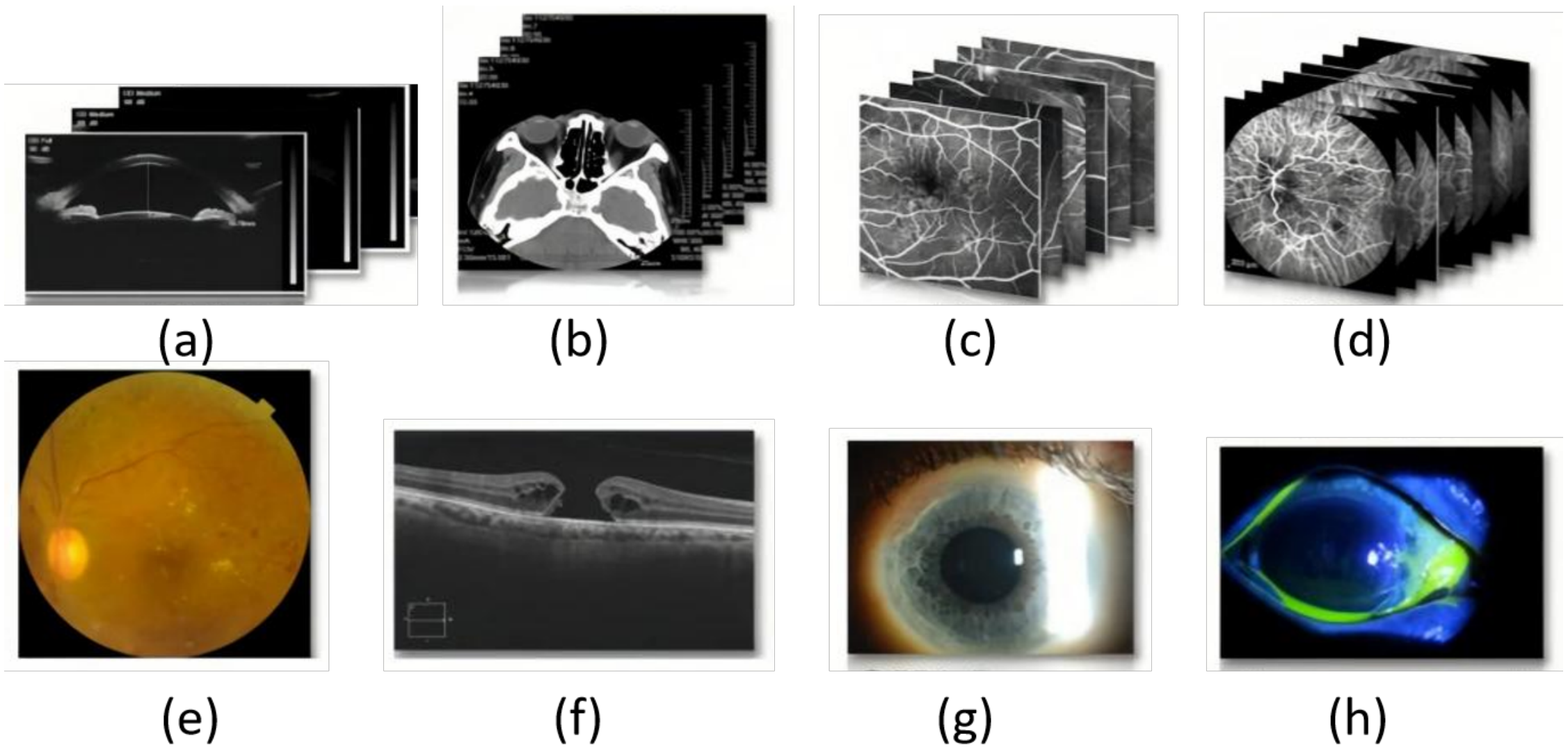

**Figure 7:** Examples of multimodal data in the Eyecare-100K dataset [19]. (a) UBM; (b) CT; (c) FFA; (d) ICGA; (e) Fundus; (f) OCT; (g) Slit-Lamp; (h) Fluorescein staining.

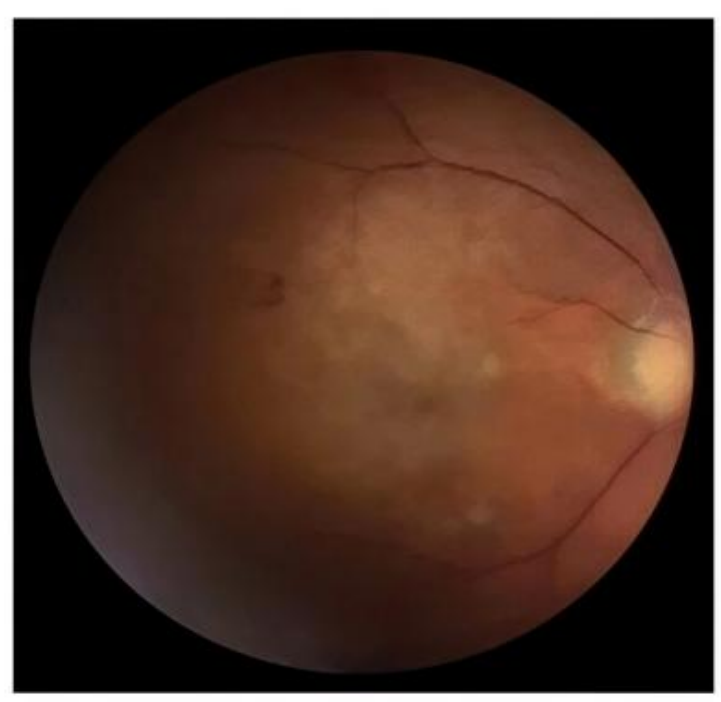

**Figure 8:** Examples of image–text pairs in the Eyecare-100K dataset [19].

## 2.2. Limitations and Future Directions

An analysis of these three stages indicates that retinal fundus image datasets have undergone substantial technological advancement and task expansion. From early-stage resources characterized by single diseases, single tasks, and limited scale, they have evolved into contemporary datasets featuring multi-disease coverage, multi-task design, and large-scale data volumes, with increasing integration of multimodal information. This evolution has laid a solid foundation for the training and validation of deep learning models in retinal image analysis.

Despite recent advances in dataset scale and annotation quality, current retinal fundus datasets still face notable limitations in supporting real-world clinical applications, particularly with respect to annotation consistency, semantic and modality coverage, data distribution, and image quality.

### *2.2.1. Limitations*

**Annotation Consistency and Label Reliability**

Annotation consistency and label reliability constitute major challenges for current public fundus image datasets [56, 57]. Discrepancies in annotation specifications across different datasets, together with subjective differences among experts in the recognition of pathological features and in the application of decision thresholds, may lead to systematic inconsistencies in annotation outcomes. For example, in the Retinopathy Online Challenge(ROC) dataset, different experts exhibit varying sensitivity levels in detecting MA [58]; similarly, substantial inter-grader variability has been reported in the delineation of the cup-disc boundary, which in turn leads to increased variability in cup-to-disc ratio (CDR) measurements [59]. In addition, some datasets provide insufficient transparency regarding annotation workflows, lacking systematic cross-validation and arbitration procedures. These deficiencies compromise annotation consistency and label reliability, introducing uncertainty into model training and evaluation[56].

With the emergence of multimodal ophthalmic datasets, label reliability has encountered new challenges. For instance, textual labels derived from large language models without rigorous human oversight are susceptible to hallucination-induced errors [19, 35]. Moreover, the lack of structured alignment and a unified labeling framework across different modalities further constrains the applicability of such datasets in multimodal tasks [54, 55].

**Semantic Information and Modality Coverage**

When extending toward complex clinical tasks, publicly accessible datasets remain insufficient in terms of semantic richness and modality coverage[60]. Most currently available public datasets provide only images and basic labels, while lacking comprehensive patient-level systemic health indicators—such as body mass index (BMI), blood pressure, and blood glucose—as well as longitudinal follow-up information[61]. This limitation hampers the development of association modeling that bridges retinal findings with systemic health conditions. Among publicly accessible datasets, TREND is one of the few that incorporates partial clinical parameters, though its sample size remains limited [62]. Furthermore, while certain multimodal datasets provide OCT or FFA images, the number of jointly annotated multimodal datasets is limited, which constrains research on cross-modal relationships.[63].

**Data Distribution and Imaging Biases**

Fundus image datasets commonly exhibit multiple sources of bias, including imbalanced disease class distributions, heterogeneous geographic origins, and variations in imaging devices, all of which substantially impair the generalization capability of learning-based models. In most datasets, healthy cases or common conditions (e.g., early-stage DR) dominate the sample composition, whereas severe pathologies and rare diseases are markedly underrepresented [35], which results in a pronounced degradation of recognition performance for these classes. In parallel, anatomical variations across populations from different regions, which manifest as differences in optic disc size and retinal vascular morphology, introduce additional domain shifts [64, 65]. Discrepancies among imaging devices from different manufacturers, including variations in spatial resolution, spectral response, and noise characteristics, further alter lesion appearance and visual contrast , which can enhance model generalization across diverse devices but also increase complexity in specific device-specific application scenarios[66].

#### *2.2.2. Future Directions*

To address the aforementioned challenges, the construction of fundus photograph datasets requires systematic upgrades across annotation frameworks, modality coverage, and data distribution..

In terms of annotation, a unified annotation standard should be established to clearly define disease grading, structural boundaries, and lesion definitions. Consistency in annotation can be strengthened through expert consensus and AI-assisted verification, providing reliable supervisory signals for downstream learning models.

Regarding information structure and modality coverage, structured collection of clinical metadata can be leveraged to build associative databases linking fundus findings with systemic health, while promoting spatial registration and semantic alignment between fundus images and heterogeneous data sources such as OCT and clinical reports, thereby improving the consistency between imaging data and corresponding textual descriptions and providing high-quality supervisory signals for multimodal pretraining models.

At the level of sample distribution and data sources, a combination of clinically targeted data acquisition and algorithmic strategies, such as difficult case sampling and synthetic data augmentation, should be adopted to mitigate class imbalance and improve model performance on rare lesions and clinically high risk targets. Meanwhile, establishing collaborative networks across multiple clinical centers and imaging devices can expand coverage across diverse populations and acquisition systems, thereby reducing generalization risks induced by geographic and device related biases.

## 3. Preprocessing Techniques for Color Fundus Images

Fundus color images are inevitably affected by imaging conditions and inter-subject variability, and therefore often suffer from quality degradations such as insufficient contrast, non-uniform illumination, noise contamination, and artifact occlusion. These issues substantially constrain the reliability of a wide range of downstream tasks, such as clinical diagnosis and lesion detection. From the perspective of technological evolution, preprocessing methods for fundus color images have undergone a progressive transition from hand-crafted, rule-driven techniques to data-driven approaches, recently to hybrid frameworks that integrate both paradigms, and ultimately toward multi-modal heterogeneous data preparation that abstracts preprocessing beyond the visual domain.

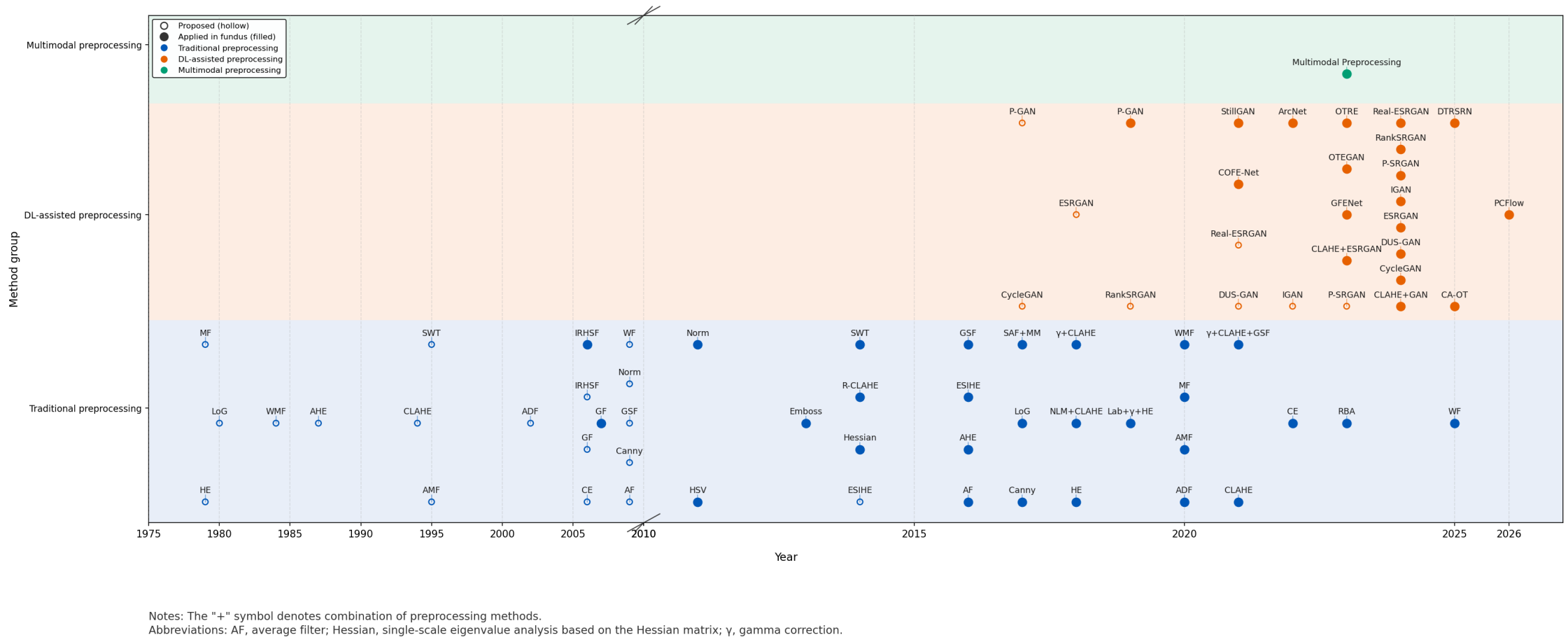

**Figure 9:** Evolution of preprocessing methods for fundus color images.

As illustrated in Figure 9, the evolution diagram uses chronological progression as the horizontal axis, showing dominant preprocessing strategies across periods and their transitions, while the vertical axis categorizes different types of preprocessing methods. It can be observed that early studies primarily relied on conventional enhancement techniques such as histogram equalization(HE), filtering, and morphological operations. Subsequently, learning-based models were gradually introduced to address challenges posed by non-uniform illumination, complex noise, and artifacts. In recent years, a clear trend has emerged toward hybrid approaches that combine prior rules with deep learning models, setting the stage for contemporary visual-language-EHR expert systems that demand advanced text normalization and self-supervised tabular imputation. To sustain the overarching theme of tri-axial co-evolution while respecting the methodological diversity of signal processing, this chapter prioritizes a feature-centric taxonomy based on intrinsic algorithmic attributes rather than a rigid chronological partitioning. This design preserves the structural granularity of individual operators—such as multi-scale filtering and cross-modal imputation—while mapping their systemic transitions to the corresponding downstream algorithmic eras within each section and the concluding synthesis.

## 3.1. Traditional Preprocessing Methods

Prior to the widespread adoption of deep learning, traditional image enhancement techniques were the predominant approaches for fundus image preprocessing [12, 67, 68]. These methods are primarily based on manually designed mathematical models and empirically tuned parameters [69], aiming to improve image quality and highlight structural features. Over time, they have formed a set of technical pathways including contrast enhancement, brightness adjustment, smoothing and denoising, as well as structural and edge enhancement [12, 70, 71]. Owing to their low computational complexity, independence from annotated data, and strong interpretability, traditional preprocessing methods were extensively applied in early fundus image analysis and clinical workflows, and they constitute an important foundation for subsequent automated analysis approaches. In the following, traditional preprocessing techniques are introduced in four subsections, covering their underlying principles, representative applications, and performance characteristics.

It should be noted that fundus image preprocessing pipelines are often accompanied by data augmentation operations to improve model generalization, with common strategies including geometric transformations (e.g., horizontal flipping and random rotation), image composition methods (e.g., mosaic), and pixel masking techniques (e.g., GridMask). As this review focuses on preprocessing techniques aimed at image quality optimization and feature extraction, these data augmentation methods are not included in the present analysis.

#### 3.1.1. Contrast and Brightness Adjustment Methods

Within traditional preprocessing methods, contrast and brightness adjustment plays a key role in optimizing the gray-level distribution and luminance uniformity of fundus images[72]. This category includes contrast enhancement methods such as histogram equalization (HE) and its variants (AHE, CLAHE)[73], luminance adjustment strategies like gamma correction or Retinex method[72], and fundamental operations including channel extraction (CE), normalization (Norm), and color correction[74].

Global HE is a classical technique for image contrast enhancement. Although it can improve overall contrast, it often leads to gray-level distortion and loss of fine details in color fundus images [73]. In 1987, Pizer et al. proposed adaptive histogram equalization (AHE), which enhances contrast by performing HE over local regions. However, in medical images, AHE tends to amplify noise in homogeneous areas [73]. In low-contrast or low-illumination regions, it may further reduce the visibility of fine vessels and even misclassify noise as small lesions [68]. Subsequently, contrast-limited adaptive histogram equalization (CLAHE) was introduced by incorporating a clipping threshold to suppress excessive local histogram amplification, thereby achieving a better balance between contrast enhancement and noise control. Further extensions of CLAHE include Rayleigh-distribution-based CLAHE(R-CLAHE) [75], and exposure-based sub-image histogram equalization (ESIHE, also referred to as EB-SIHE) [68]. These approaches introduce statistical distribution constraints, or exposure-aware mechanisms, enabling the enhancement strength to be dynamically adjusted according to local luminance levels. This design enhances details in dark regions while suppressing over-enhancement in bright regions, thereby improving overall luminance balance and structural readability.

In addition to contrast enhancement, luminance adjustment strategies constitute a crucial component of fundus image preprocessing. Common approaches include gamma correction, which performs nonlinear mapping of pixel intensities to enhance dark regions and improve overall brightness uniformity [72], and Retinex-based methods, which decompose images into illumination and reflectance components, suppress uneven lighting, and enhance structural details [72].

Subsequently, fundamental operations include channel extraction (CE), normalization (Norm), and color correction. Previous studies have shown that the green channel provides more stable contrast for vessels and exudates, and using only the green channel can improve vessel detection rates on certain datasets[12, 25], although such single-channel strategies may weaken multispectral complementary information and are unfavorable for hemorrhagic lesions that rely on features from the red channel [12]. Normalization reduces luminance variations caused by differences in imaging devices and exposure conditions by unifying pixel value scales, thereby improving model training stability [67]. Color and luminance correction methods compensate for imaging condition variability through brightness-contrast perturbation, adjustments in the HSV color space, or independent multi-channel correction, enhancing model adaptability to multi-source fundus data. Nevertheless, inappropriate parameter settings may introduce color distortion or attenuate discriminative features [76–78].

To further illustrate the visual differences among various contrast enhancement strategies when applied to the same fundus image, Figure 10 presents a comparative visualization of the processing results obtained using several representative methods.

#### 3.1.2. Smoothing and Denoising Methods

In traditional fundus image preprocessing, smoothing and denoising methods primarily aim to suppress imaging noise and artifacts so as to improve the stability and robustness of subsequent vessel or lesion analysis. The core challenge lies in effectively reducing noise while preserving high-frequency details such as fine vessels and lesion boundaries. According to the processing domain, these methods can be broadly categorized into spatial-domain filtering and frequency-domain filtering approaches.

Spatial-domain filtering operates directly on pixel neighborhoods in the pixel domain and can be further divided into linear and nonlinear filtering methods. Linear filtering reduces spatial noise through linear transformations, with representative techniques including mean filtering, Gaussian filtering(GSF), and Wiener filtering(WF). Mean filtering achieves rapid smoothing by averaging local neighborhoods but tends to blur critical details such as vessel edges and optic disc boundaries [79]. Gaussian filtering introduces distance-based weighting, resulting in more stable smoothing performance; however, it may still attenuate fine anatomical structures [80]. Wiener filtering adaptively adjusts the degree of smoothing based on local statistical characteristics, offering a better trade-off between noise suppression and detail preservation [79, 81].

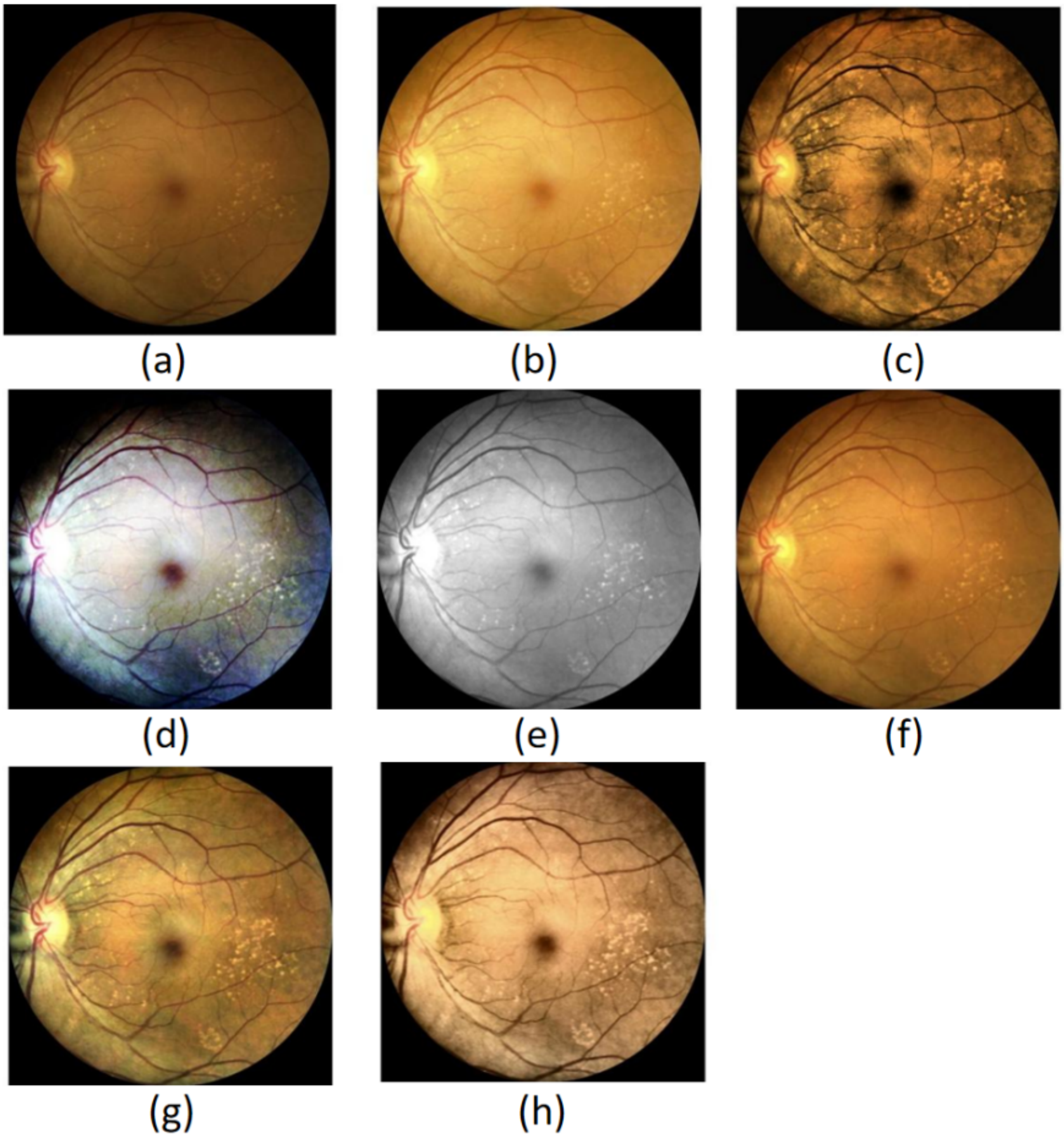

**Figure 10:** Comparison of methods for adjusting contrast and brightness effects. (a) Original image; (b) HE; (c) AHE; (d) ESIHE; (e) Green channel; (f) Normalization; (g) CLAHE; (h) R-CLAHE.

Among nonlinear filtering methods, median filtering [22] is effective in removing impulse (salt-and-pepper) noise while maintaining edge sharpness, but it falters when the probability of impulse noise occurrence becomes high[82]. Representative improved variants include weighted median filtering(WMF) [83], adaptive median filtering(AMF)[82], and anisotropic diffusion filtering(ADF) [84]. These methods enhance structural preservation by introducing weighting mechanisms, adaptive window sizes, or diffusion control strategies. However, these nonlinear filtering methods may still introduce detail blurring or structural attenuation under high noise conditions, and their performance often depends on appropriate parameter or window-size selection.

Frequency-domain filtering emphasizes frequency and directional characteristics to enhance oriented fine structures. For instance, the Gabor filter(GF) proposed by Daugman enhances local feature representation using multi-orientation and multi-frequency kernels [85]. Subsequent multi-directional Gabor fusion methods integrate responses from different orientations via directional score fusion, improving vessel continuity and traceability in crossing and bifurcation regions [86]. In addition, some studies have introduced multiscale transform-domain denoising strategies. For example, approaches based on the stationary wavelet transform (SWT) [87] employ shift-invariant decomposition and threshold-based reconstruction to mitigate vessel artifacts induced by conventional wavelet transforms and to achieve stable suppression of mixed noise. Despite these advantages, these methods remain sensitive to frequency, orientation, or threshold parameter selection, potentially attenuating fine structural details under different imaging conditions.

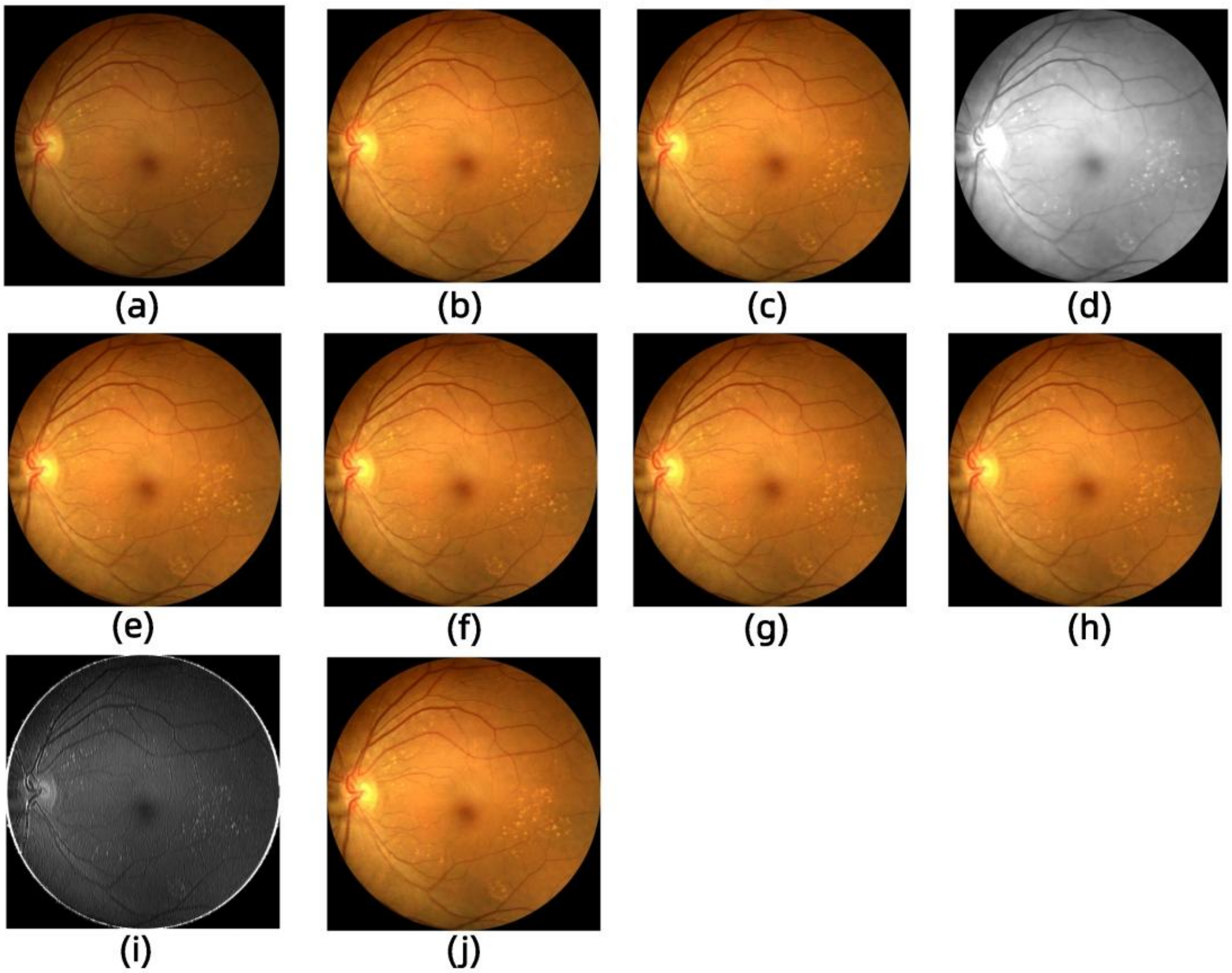

**Figure 11:** Comparison of smoothing and denoising methods. (a) Original image; (b) Mean filter; (c) Gaussian filter; (d) Wiener filter; (e) Median filter; (f) Weighted median filter; (g) Adaptive median filter; (h) Anisotropic diffusion filter; (i) Gabor filter; (j) Stationary wavelet transform.

To provide an intuitive comparison of the effects of different smoothing and denoising methods on noise suppression and structural preservation in fundus images, Figure 11 compares the processing results of several representative algorithms.

### *3.1.3. Structure and Edge Enhancement Methods*

Structure and edge enhancement methods aim to reinforce the spatial morphology and boundary information of target structures such as blood vessels and lesions, thereby improving the separability between regions of interest and the background[88, 89].

Early studies predominantly relied on gradient operators and differential operators to achieve edge enhancement, such as the Laplacian of Gaussian (LoG) and the Canny operator, which extract continuous edges via zero-crossing detection and gradient constraints, respectively [69]. In addition, some studies approached the problem from the perspective of second-order structural modeling by introducing vessel enhancement methods based on the eigenvalue analysis of the Hessian matrix, which emphasize the ridge-like structures of retinal vessels by characterizing their local curvature properties [90]. Meanwhile, other works incorporated structure visualization and directional enhancement strategies; for example, Emboss filtering highlights spatial relief features by generating topographic maps that illustrate elevations and depressions in the retinal posterior pole [91]. However, these traditional structure and edge enhancement methods are often sensitive to noise, illumination variations, and parameter settings, which may lead to pseudo-edge generation, weak responses to low-contrast vessels, or loss of fine structural details. To provide an intuitive comparison of the effects of different structure and edge enhancement methods on reinforcing vessel and lesion contours in fundus images, Figure 12 presents the processing results of several representative algorithms.

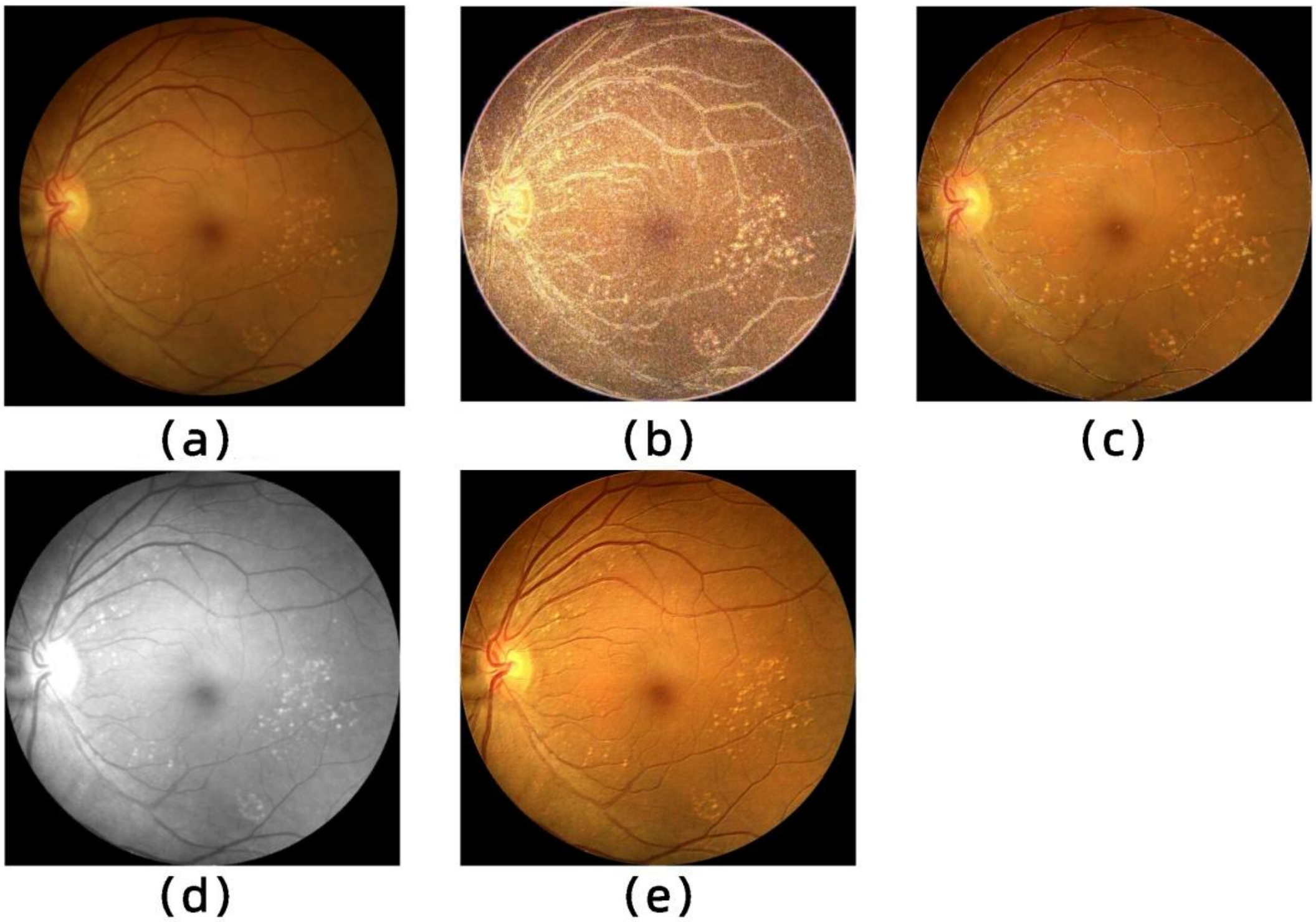

**Figure 12:** Comparison of structure and edge enhancement methods. (a) Original image; (b) LoG; (c) Canny operator; (d) Multiscale vessel enhancement based on the Hessian matrix; (e) Embossing.

### *3.1.4. Combination strategies of traditional preprocessing*

Given the limitations of individual traditional preprocessing methods, some studies have combined multiple enhancement, denoising, and structural processing strategies to improve overall preprocessing performance. To address insufficient brightness distribution and contrast, Gupta et al. combined adaptive gamma correction with quantile-weighted histogram equalization in the Lab color space to maintain color consistency [92]; Zhou et al. adopted a cascade of gamma correction and CLAHE on the luminance channel to enhance low-illumination details [93].

Meanwhile, several studies jointly applied contrast enhancement and noise removal to improve the stability of enhancement results. Desiani et al. combined luminance adjustment and contrast enhancement methods such as gamma correction and CLAHE with median or Gaussian filtering to suppress noise amplification [94]. Subudhi et al. employed a serial pipeline of median filtering, resizing, grayscale conversion, and PSO-optimized enhancement to achieve simultaneous noise suppression and vessel structure reinforcement [95]. Shailesh Kumar [96], Badgujar [97], and Naveed [98] applied denoising filters such as Non-Local Adaptive Fuzzy Switching Median Filter(NAFSM), Block-Matching and 3D Filtering(BM3D), or non local means(NLM) in conjunction with histogram equalization methods including HE, ADHE, CLAHE, or ESIHE to enhance low-intensity structures and improve structural continuity while suppressing mixed noise. In addition, Khawaja et al. combined probabilistic patch-based denoising with SVD and CLAHE enhancement to suppress amplified noise and improve retinal vessel detail preservation [99].

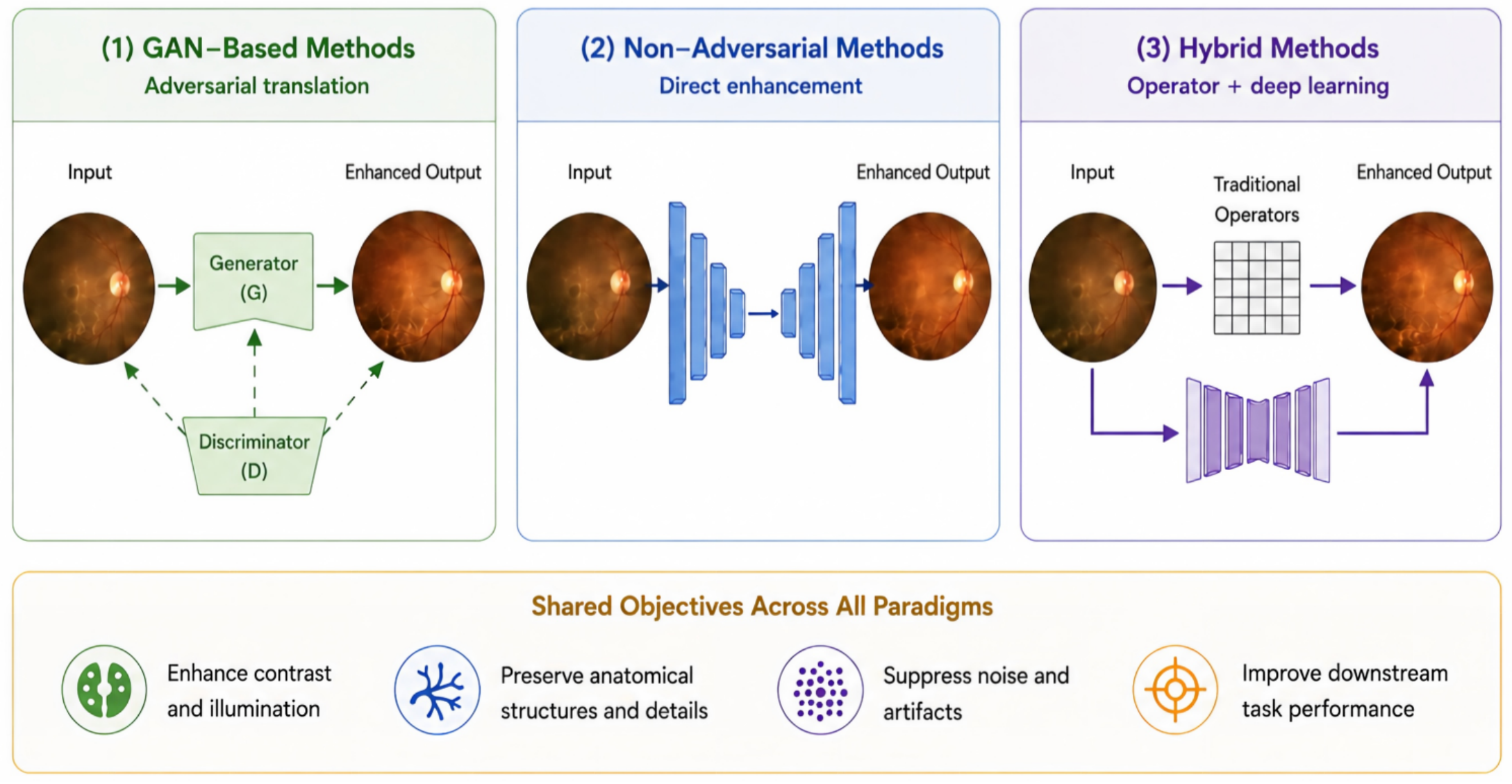

**Figure 13:** Illustration of Deep Learning–Assisted Preprocessing Methods.

These hybrid strategies partially compensate for the limitations of individual preprocessing methods by jointly improving contrast enhancement, noise suppression, and structural preservation. However, their performance still depends on appropriate parameter coordination and may vary across different imaging conditions and lesion distributions.

## 3.2. Deep Learning–Assisted Preprocessing Methods

Traditional preprocessing methods commonly utilize manually designed operators and fixed parameter settings. Although effective in specific scenarios, they often suffer from limited adaptability, parameter sensitivity, and detail loss under complex imaging conditions[74, 100]. In contrast, deep learning–based preprocessing methods can automatically learn complex intensity distributions and structural representations from large-scale fundus datasets, enabling more robust enhancement and detail preservation through end-to-end optimization [38, 80, 101]. Accordingly, deep learning–assisted preprocessing methods are categorized in this review into GAN-based methods, non-adversarial methods, and hybrid preprocessing methods integrating traditional operators with deep learning frameworks, as shown in Figure 13.

### *3.2.1. GAN–Based Preprocessing Methods*

Generative adversarial networks (GANs), through the adversarial training mechanism between generators and discriminators, have demonstrated strong capabilities in complex image domain translation tasks, and are particularly well suited for applications such as artifact removal, super-resolution reconstruction, and distribution correction in fundus imaging [23, 24, 102]. In fundus color image preprocessing, CycleGAN [24] and its variant StillGAN [103] enable cross-domain mapping without paired data and have been widely applied to image quality enhancement and domain adaptation tasks; these methods effectively suppress common artifacts such as haze, eyelash occlusion, and specular reflections, while largely preserving key anatomical structures including blood vessels and the optic disc. Building upon this framework, subsequent studies have introduced optimal transport constraints and structural consistency regularization to further improve the stability of distribution migration and the fidelity of anatomical structures [104, 105].

With respect to adversarial learning–based preprocessing, Progressive GAN with Triplet Loss (P-GAN) further extends GANs to high-magnification super-resolution reconstruction of fundus images; by adopting progressive training and incorporating triplet loss, it effectively alleviates detail blurring in high-scale super-resolution, and the generated high-resolution images outperform multiple traditional super-resolution methods in downstream tasks such as microaneurysm detection [106]. Meanwhile, several general GAN-based super-resolution approaches, including ESRGAN, P-SRGAN, RankSRGAN, DUS-GAN, IGAN, and Real-ESRGAN, have also been introduced into fundus image enhancement studies. Jia et al. conducted a systematic comparison of these methods and proposed a GAN-based super-resolution framework guided by vascular structural priors, aiming to mitigate vessel detail blurring and structural distortion commonly observed in fundus scenarios [102]. In addition, by refining loss functions and training strategies, CycleGAN has been further enhanced for stable cross-domain mapping, enabling unlabeled source-domain fundus images to be transformed into the target-domain distribution and thereby serving as a data distribution correction step prior to diagnosis [23].

#### *3.2.2. Non-Adversarial Preprocessing Methods*

Non-adversarial deep learning–based preprocessing methods, including CNN- and transformer-based architectures, have demonstrated strong capability in suppressing noise and preserving structural details in fundus images [101, 107, 108].

In terms of noise suppression, Shen et al. proposed cofe-Net, which constructs a degradation model to correct uneven illumination while preserving retinal anatomical structures [108]. Lee et al. developed an attention-based U-Net framework for fundus image enhancement, in which paired low-quality and high-quality images were constructed and multiple degradation types, including blur, haze, and low illumination, were simulated to improve the quality of fundus images under complex degradation conditions [107]. Ye et al. employed Retinex theory to decompose images and optimize the illumination component, effectively addressing non-uniform illumination while simultaneously improving downstream vessel segmentation performance [101].

Regarding structural preservation, Pazo et al. introduced a dual-Transformer residual network, DTRSRN, which integrates global contextual modeling with fine-grained feature extraction, thereby enhancing vascular structure preservation in super-resolution reconstruction [109]. Li et al. designed an unsupervised framework termed ArcNet, which simulates degradation processes, incorporates high-frequency structural constraints, and combines domain adaptation strategies to achieve annotation-free restoration of cataract-affected fundus images [110]. In addition, Vasa et al. proposed an optimal transport-based retinal image enhancement paradigm in deep feature space, which effectively suppresses noise while minimizing over-correction of pathological regions and anatomical structures [111].

### 3.3. Hybrid Preprocessing Methods

As shown in Appendix E, single traditional preprocessing techniques exhibit limited generalization capability under complex degradation scenarios. In contrast, purely deep learning–based enhancement models provide stronger nonlinear modeling capacity but typically rely heavily on large-scale datasets and substantial computational resources. To balance structural priors with data-driven learning advantages, recent studies have explored hybrid preprocessing strategies that integrate traditional image processing techniques with deep learning models.

One category involves serial hybrid architectures. For instance, Alwakid et al. combined CLAHE with ESRGAN for image enhancement in DR grading tasks before feeding the output into a classification network [112]. Similarly, Bhoopal et al. employed a serial approach using a joint CLAHE and GAN preprocessing pipeline, where CLAHE enhances local contrast and the GAN further restores fine details [113].

Another category consists of embedded hybrid enhancement models. Within this group, Xu et al. proposed PCFlow, which embeds traditional high-frequency filtering operators into a deep learning network to extract retinal structures as constraints, utilizing invertible pyramid normalizing flows to achieve end-to-end hierarchical enhancement in the frequency domain [114]. Additionally, Li et al. introduced GFENet, which extracts High-Frequency Maps (HFM) via Gaussian high-pass filtering and embeds them into a deep enhancement network as frequency-domain self-supervised signals [115].

### 3.4. Multi-modal Preprocess

**Characteristics and Architectural Evolution**

The paradigmatic shift of fundus color image (CFP) analysis from isolated visual feature extraction toward multimodal collaborative intelligence—as prominently evidenced by heterogeneous cohorts like GAMMA [52] , OCTA-Fundus-DR [53], and Eyecare-100K [19]—has fundamentally expanded the mathematical definition and engineering execution of preprocessing workflows. In this advanced data landscape, preprocessing has transcended conventional single-modality pixel manipulation, evolving into a multi-axial framework tasked with standardizing cross-modal information.

For million-scale, self-supervised retinal foundation architectures such as RETFound [54] and VisionFM [55], explicit handcrafted illumination correction and heuristic contrast enhancements have largely been internalized through massive pretraining, shifting the primary visual preprocessing pivot toward structural, hardware-aware input preparation. This includes localized patch tokenization, dynamic resolution scaling, and token pruning designed to bypass the quadratic computational complexity inherent to self-attention mechanisms without sacrificing micro-pathological sensitivity. Concurrently, the inclusion of text-paired expert diagnostic reports and report-paired corpora, exemplified by SynFundus-1M [20] and EyeCLIP [35], has integrated natural language processing (NLP) heuristics directly into the data preparation lifecycle. Here, preprocessing demands rigorous clinical text normalization, vocabulary filtering, and prompt template crafting to transmute highly unstructured, free-text clinical narratives into structured, semantic tokens compatible with downstream cross-modal alignment layers.

**Emerging Trends**

Beyond text-image pairing, contemporary multi-modal engineering increasingly gravitates toward two sophisticated trends: cross-modal spatial synchronization and self-supervised tabular imputation. The integration of multi-modality structural imaging (e.g., fundus photographs combined with 3D-OCT or OCTA [53] necessitates precise spatial co-registration, shifting preprocessing toward rigid and non-rigid anatomical alignment protocols to preserve physical and geometric consistency during multi-source feature fusion. Furthermore, as fundus expert systems increasingly assimilate longitudinal electronic health records (EHRs) and systemic clinical risk factors, preprocessing must confront the acute sparsity and irregular sampling inherent to real-world clinical variables. The emerging trend in processing these tabular streams involves abandoning brittle, heuristic mean/median fillers in favor of self-supervised masked modeling and generative adversarial imputation networks, thereby empowering the data foundation to natively sustain clinical uncertainty without degrading diagnostic sensitivity.

**Persistent Bottlenecks and Methodological Tensions**

Despite its theoretical potential, this multi-modal preprocessing paradigm exposes critical methodological vulnerabilities that constrain real-world clinical translation. A primary bottleneck stems from the lack of reporting standards and unified criteria for clinical text and tabular curation. Ophthalmic text normalization heavily relies on ad-hoc, center-specific NLP rules, which inadvertently introduces severe semantic domain shifts and restricts the cross-center generalizability of multimodal models compared to their unimodal visual counterparts.

Most fundamentally, a severe tension persists regarding uncalibrated uncertainty propagation. Sub-optimal data imputation or flawed prompt templates applied to highly sparse EHR data or low-quality visual inputs often introduce artificial semantic noise and structural artifacts. These upstream preprocessing imperfections are catastrophically amplified across heterogeneous fusion layers, inevitably propagating into uninterpretable cross-modal hallucinations and compromised decision transparency. Because current evaluation protocols lack quantitative metrics to assess the independent robustness and reproducibility of the preprocessing pipeline itself, this high-dimensional data preparation phase remains a "black box," severely impeding the sustainable deployment of multimodal ophthalmic AI in resource-constrained primary screening workflows.

## 3.5. Limitations and Future Directions

### *3.5.1. Limitations*

To provide a rigorous engineering appraisal for real-world expert system deployment, the intrinsic vulnerabilities of current color fundus photography (CFP) preprocessing paradigms are synthesized across three core engineering dimensions: parameter volatility, structural distortion, and cross-modal uncertainty cascades.

**Parameter Volatility and Domain Adaptability in Classical and Hybrid Schemes**

Traditional mathematical operators, including contrast-limited adaptive histogram equalization (CLAHE) and Gabor filtering, alongside their serial hybrid configurations, suffer from an acute dependency on empirical parameter configurations such as clip thresholds, window sizes, and directional orientations. Because these parameters are typically optimized on specific, homogeneous datasets, their performance deteriorates sharply when confronted with multi-center clinical data characterized by non-uniform illumination and sensor-specific variations. Inappropriate tuning frequently triggers gray-level distortions or amplifies background noise, which downstream models misinterpret as micro-pathological lesions. This parameter fragility has recently escalated into the multimodal domain. Current text normalization and prompt template engineering for unstructured expert reports rely heavily on ad-hoc, center-specific natural language processing (NLP) heuristics. These brittle text-filtering rules fail to accommodate semantic variations across multi-center electronic health records (EHRs), introducing severe linguistic domain shifts that undermine system stability.

**Structural Distortion, Artificial Artifacts, and Black-Box Obfuscation in Deep Enhancements**

While data-driven models offer powerful nonlinear mapping for artifact removal and illumination correction, they introduce substantial risks of structural distortion. Generative adversarial networks (GANs) operating without rigorous geometric regularizations are prone to hallucinating or obliterating fine-grained pathological structures, such as early-stage microaneurysms or subtle drusen clusters, in pursuit of macro-level visual smoothness. Furthermore, deep filtering and super-resolution architectures often blur vessel boundaries in low-contrast peripheral retinal regions. Crucially, as preprocessing moves from explicit pixel manipulation to implicit deep feature space optimization, such as token pruning or neural Retinex decomposition, the preprocessing phase itself effectively becomes a black box. Existing evaluation frameworks rely on unstandardized, heterogeneous visual metrics like peak signal-to-noise ratio (PSNR) and structural similarity index measure (SSIM) that correlate poorly with downstream diagnostic accuracy, rendering the independent robustness of the data preparation phase unquantifiable and uninterpretable to clinicians.

**Heterogeneous Noise and Cascading Uncertainty Propagation in Multimodal Curation**

The cutting-edge transition to multimodal ophthalmic expert systems has shifted the preprocessing bottleneck from single-modality pixels to high-dimensional heterogeneous data synchronization. This multi-modal curation vulnerability is fundamentally driven by a dual-faceted engineering friction across distinct clinical spaces.

On the spatial dimension, anatomical co-registration discrepancies severely compromise visual-structural alignment. The rigid and non-rigid alignment protocols required to synchronize 2D fundus photographs with 3D structural modalities—such as optical coherence tomography (OCT) or optical coherence tomography angiography (OCTA)—frequently disintegrate when confronting severe pathological distortions, thereby injecting geometric alignment noise into the visual stream. Concurrently, on the clinical dimension, acute tabular sparsity and irregular sampling undermine the systemic fidelity of longitudinal patient histories within electronic health record (EHR) cohorts. The deployment of brittle, heuristic mean or median imputation during data preparation introduces artificial statistical artifacts that actively dissociate data distributions from true pathophysiological correlations.

Crucially, when these uncalibrated visual, structural, and tabular discrepancies intersect within downstream heterogeneous fusion layers, the localized noises undergo catastrophic compound amplification. This systemic error propagation ultimately culminates in uninterpretable cross-modal hallucinations within large multimodal models (LMMs), fundamentally dismantling the clinical trustworthiness of the expert system.

#### *3.5.2. Future Directions*

To bridge the gap between theoretical algorithmic capacity and high-throughput primary care deployment, future research must transition from post-hoc, isolated input fixes to a unified, hardware-aware, and causally-aligned preprocessing framework along three strategic trajectories.

**Adaptive Multi-Axial Conditioning and Standardized Clinical NLP Pipelines**

Future preprocessing paradigms must discard empirical parameter tuning in favor of self-adaptive, data-driven image enhancement strategies. By embedding reinforcement learning or continuous optimization loops, the enhancement intensity, contrast thresholds, and localized regions-of-interest (ROIs) should automatically calibrate according to the degradation coefficients of the input image. In parallel, this standardized adaptability must extend to non-image modalities. The ophthalmic community must establish unified, open-source benchmarks for clinical text normalization and medical prompt engineering. Developing robust, standardized medical NLP preprocessing pipelines is an absolute prerequisite to dismantle semantic domain barriers and unlock generalizable multimodal synthesis across heterogeneous global registries.

**Precision Pathological Preservation and Robust Self-Supervised Tabular Imputation**

Achieving an absolute balance between noise suppression and high-frequency structural fidelity remains paramount. Future architectures should integrate explicit anatomical structural priors, such as vessel topology graphs or optic disc boundary constraints, directly into deep enhancement and spatial co-registration pipelines to prevent generative hallucinations. To resolve the acute sparsity of longitudinal clinical variables, data preparation frameworks must abandon heuristic statistical imputation. Instead, research should leverage advanced self-supervised masked modeling and generative adversarial imputation networks natively designed for tabular streams. This methodology will empower the data foundation to handle highly incomplete clinical data streams gracefully, isolating structural noise and sustaining clinical uncertainty before the data enters downstream cross-modal reasoning backbones.

**Lightweight, Hardware-Aware Preprocessing for Resource-Constrained Edge Deployment**

The prohibitive computational footprint and memory overhead imposed by processing ultra-high-resolution fundus images alongside long-form textual reports represent a massive deployment barrier. Future expert system engineering must optimize joint preprocessing-modeling workflows to facilitate edge-device translation in community screening.

Rather than treating efficiency as a post-hoc compression step, lightweight, unsupervised visual corrections must be embedded directly into edge-computing hardware or camera firmware.

More importantly, hardware-aware edge preprocessing modules, such as automated micro-lesion ROI cropping and dynamic multi-resolution scaling, must work in tight synergy with token pruning, dynamic sparse attention, and linearized sequence scan topologies like State Space Models (SSMs) and Mamba. By passing raw high-resolution images through a lightweight ROI crop or high-pass filter before sending the streamlined tokens to a linear Mamba backbone, the processing complexity can be reduced from a quadratic O(N2)scaling to a linear O(N) relationship relative to the sequence length. This holistic coordination will drastically lower engineering overhead, ensuring sustainable, high-throughput deployment in low-resource primary care screening scenarios.

## 4. Methodological Evolution of CFP Analysis Algorithms

The evolution of CFP analysis algorithms cannot be thoroughly understood as an isolated algorithmic phenomenon; rather, it represents a deep paradigm shift from hand-crafted prior modeling to data-driven latent representation learning [7, 116], and from vision-centric localized feature extraction to global contextual reasoning and clinical intelligence [54, 117–119]. Crucially, this algorithmic trajectory has been continuously mirrored by, and entangled with, a parallel evolution in image and data preparation strategies. To systematically untangle the co-evolutionary dependencies across these methodological epochs, this section structures the field into four distinct paradigm shifts, analyzed through the holistic lens of joint data-preprocessing-algorithm pipelines:

1.The Knowledge-Driven Paradigm: Characterized by explicit mathematical modeling and handcrafted prior descriptors [7], where algorithms were strictly dependent on rigid, deterministic pixel-level preprocessing configurations—such as contrast-limited adaptive histogram equalization (CLAHE), illumination correction, and binary field-of-view masking—to suppress imaging noise and guarantee feature stability.

2.The Deep Convolutional Paradigm: Institutionalizing end-to-end convolutional neural networks (CNNs) for robust representation learning [116], which simultaneously shifted the preprocessing axis from manual pixel manipulation to scalable, neural network-driven data engineering, including automated image quality assessment (IQA) screening networks, localized region-of-interest (ROI) detection, and extensive online geometric and semantic data augmentations.

3.The Unimodal Global and Foundation Paradigm: Overcoming the intrinsic inductive biases of localized receptive fields via Vision Transformers (ViTs) and State Space Models (SSMs) [54, 119, 120], a stage wherein traditional explicit image-enhancement preprocessing became progressively internalized and mitigated through massive self-supervised pretraining, while forcing a preprocessing pivot toward hardware-aware patch tokenization, dynamic resolution scaling, and token-pruning pipelines.

4.The Multimodal Joint Modeling Paradigm: Oriented toward real-world clinical decision-making through the explicit alignment of multi-source heterogeneous evidence [117, 118], which fundamentally expanded the definition of preprocessing beyond the visual domain to encompass structured diagnostic text standardization, prompt template crafting, and advanced missing-value tabular imputation for electronic health records (EHRs).

To systematically illustrate this technological progression, Figure 14 presents the evolutionary roadmap of CFP analysis methods, spanning from knowledge-driven approaches and convolutional neural networks to vision-based foundation models and multimodal medical AI systems. By decoding each milestone through its representative architectures, data dependencies, and operational workflows, this comprehensive analysis aims to uncover the systemic engineering logic that has propelled CFP analysis from isolated "perceptual image enhancement" to holistic, deployment-ready "clinical reasoning expert systems."

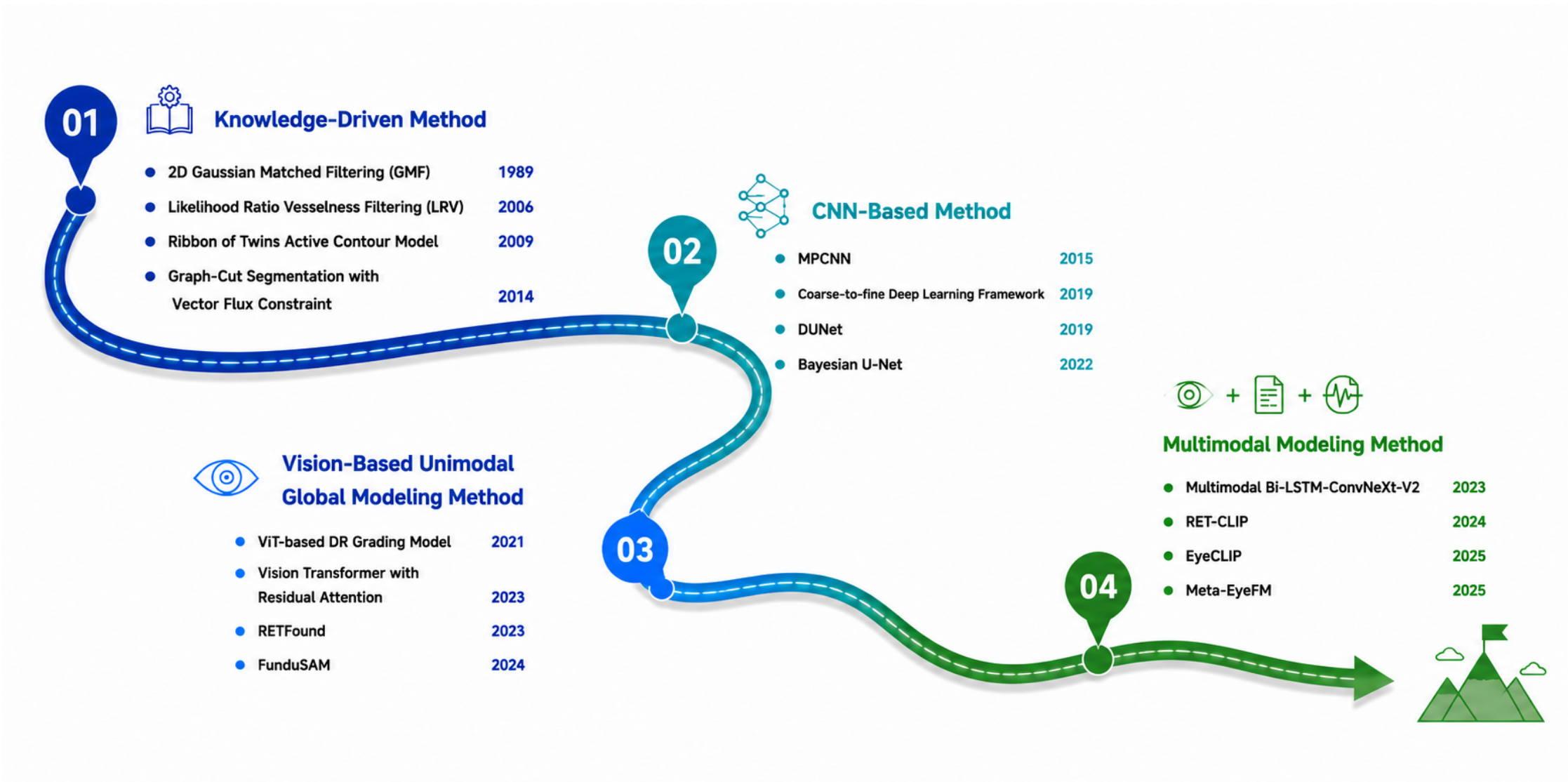

**Figure 14:** Evolution of CFP Analysis Algorithm.

Table 2: Representative Algorithms in the Evolution of CFP Analysis

| Method | Task | Target | Preprocessing | Dataset | Performance |
|---|---|---|---|---|---|
| 2D Gaussian Matched Filtering [121] | Detection | Retinal vessels | Mean filtering for noise suppression; field-of-view masking; automatic threshold segmentation using between-class variance maximization | DRIVE | Acc = 0.877, AUC = 0.788 |
| Mathematical Morphology with Curvature Evaluation [122] | Segmentation | Retinal vessels | Noise suppression; connectivity-based morphological reconstruction; intensity normalization | STARE | Sen = 0.675, Spe = 0.957, AUC = 0.904 |
| Ridge-Based Vessel Segmentation Using Profile-Based Measurement [28] | Segmentation | Retinal vessels | Green-channel intensity extraction; local mean-variance normalization; multi-scale Gaussian derivative computation | Utrecht; Hoover | Utrecht: Acc = 0.944, AUC = 0.952; Hoover: Acc = 0.952, AUC = 0.961 |
| Ribbon of Twins Active Contour Model [123] | Segmentation | Retinal vessels | Morphological centerline initialization; removal of spurious branches and junction artifacts | DRIVE; STARE; REVIEW | DRIVE: Sen = 0.732; HRIS: vessel width error = 0.420 pixels |
| Graph-Cut Segmentation with Vector Flux Constraint [124] | Segmentation | Retinal vessels and optic disc | Intensity inversion; adaptive histogram equalization; morphological opening; distance-transform-based vector field initialization | DRIVE; STARE | DRIVE: Sen = 0.751, Acc ≈ 0.947; STARE: Sen = 0.789, Acc = 0.944, FPR = 0.037 |
| Wavelet Transform with k-Nearest Neighbor Classification [125] | Classification | Diabetic retinopathy | Median filtering; adaptive histogram equalization; green-channel extraction; optic disc removal | DIARETDB1 | Acc = 0.950, Sen = 0.926, Spe = 0.876, Precision = 0.961 |
| Line Operator with Linear SVM Classification [126] | Segmentation | Retinal vessels | Inverted green-channel extraction; field-of-view correction | DRIVE; STARE | DRIVE: AUC = 0.963, Acc = 0.960; STARE: AUC = 0.968, Acc = 0.965 |
| Multiscale Gabor Filtering with GMM Segmentation [127] | Segmentation | Retinal vessels | Green-channel extraction; contrast enhancement | DRIVE | AUC = 0.974, Sen = 0.965, Spe = 0.971 |
| Fully Connected Conditional Random Field Segmentation [128] | Segmentation | Retinal vessels | NR | DRIVE; STARE; CHASE_DB1; HRF | DRIVE: Sen = 0.790, Spe = 0.968; STARE: Sen = 0.768, Spe = 0.974; CHASE_DB1: Sen = 0.728, Spe = 0.951; HRF: Sen = 0.787, Spe = 0.958 |
| Ensemble Classification-Based Vessel Segmentation [129] | Segmentation | Retinal vessels | Green-channel extraction; contrast normalization | DRIVE; STARE | DRIVE: AUC = 0.975, Acc = 0.948, Sen = 0.741, Spe = 0.981; STARE: AUC = 0.977, Acc = 0.953, Sen = 0.755, Spe = 0.976 |
| W-Net [130] | Segmentation | Optic disc and optic cup | Region-of-interest cropping; contrast enhancement using CLAHE; Gaussian filtering; polar coordinate transformation | REFUGE; DRISHTI-GS; RIM-ONE r3; Private dataset | REFUGE: OD (OE = 0.068, F1 = 0.964, Spe = 0.986, Sen = 0.974, Acc = 0.982), OC (OE = 0.178, F1 = 0.900, Spe = 0.995, Sen = 0.905, Acc = 0.987); DRISHTI-GS: OD (OE = 0.054, F1 = 0.971, Spe = 0.994, Sen = 0.958, Acc = 0.981), OC (OE = 0.233, F1 = 0.858, Spe = 0.983, Sen = 0.925, Acc = 0.972); RIM-ONE r3: OD (OE = 0.049, F1 = 0.975, Spe = 0.986, Sen = 0.981, Acc = 0.984), OC (OE = 0.237, F1 = 0.863, Spe = 0.986, Sen = 0.895, Acc = 0.972); Private dataset: OD (OE = 0.051, F1 = 0.974, Spe = 0.990, Sen = 0.967, Acc = 0.983), OC (OE = 0.255, F1 = 0.843, Spe = 0.989, Sen = 0.879, Acc = 0.976) |
| Coarse-to-fine deep learning framework [131] | Segmentation | Optic disc | Removal of non-retinal background; field-of-view cropping; image normalization; data augmentation using image flipping and rotation | CFI; DIARETDB0; DIARETDB1; DRIONS-DB; DRIVE; MESSIDOR; ORIGA | Overall (n = 2978): IoU = 0.891, DSC = 0.939, Acc = 0.970, Sen = 0.944 |

Table 2: Representative Algorithms in the Evolution of CFP Analysis (continued)

| Method | Task | Target | Preprocessing | Dataset | Performance |
|---|---|---|---|---|---|
| Bayesian U-Net [132] | Segmentation | Optic disc | Pseudo-label generation using Hough transform; Gaussian filtering; morphological opening; Canny edge detection; morphological dilation; zero padding and resizing to 640 × 640 | DRISHTI-GS; RIM-ONE; REFUGE validation set; REFUGE test set | DRISHTI-GS: Acc = 0.997, Sen = 0.983, Spe = 0.997, IoU = 0.895, DSC = 0.944, HD = 3.615, ASD = 3.394; RIM-ONE: Acc = 0.992, Sen = 0.974, Spe = 0.992, IoU = 0.783, DSC = 0.876, HD = 4.993, ASD = 8.886; REFUGE validation set: Acc = 0.997, Sen = 0.965, Spe = 0.997, IoU = 0.829, DSC = 0.903, HD = 4.043, ASD = 7.343; REFUGE test set: Acc = 0.997, Sen = 0.973, Spe = 0.997, IoU = 0.819, DSC = 0.896, HD = 4.049, ASD = 6.954 |
| SEGAN [133] | Segmentation | Retinal vessels | Intensity normalization; dataset-specific resolution adjustment using downsampling; spatial padding; data augmentation using rotation and flipping | DRIVE; STARE; CHASE_DB1; HRF | DRIVE: Sen = 0.829, Spe = 0.981, Precision = 0.840, Acc = 0.956, AUC = 0.983, G = 0.902, F1 = 0.835; STARE: Sen = 0.881, Spe = 0.978, Precision = 0.795, Acc = 0.967, AUC = 0.986, G = 0.928, F1 = 0.836; CHASE_DB1: Sen = 0.844, Spe = 0.978, Precision = 0.801, Acc = 0.963, AUC = 0.987, G = 0.908, F1 = 0.822; HRF: Sen = 0.831, Spe = 0.973, Precision = 0.812, Acc = 0.956, AUC = 0.969, G = 0.899, F1 = 0.821 |
| DA-Res2UNet [134] | Segmentation | Retinal vessels | Removal of non-retinal borders; gamma correction; grayscale conversion; CLAHE; Z-score normalization | DRIVE; CHASE_DB1; STARE | DRIVE: Acc = 0.970, Spe = 0.985, Sen = 0.820, AUC = 0.987, MCC = 0.812, F1 = 0.827; CHASE_DB1: Acc = 0.977, Spe = 0.989, Sen = 0.808, AUC = 0.991, MCC = 0.804, F1 = 0.816; STARE: Acc = 0.976, Spe = 0.989, Sen = 0.814, AUC = 0.988, MCC = 0.823, F1 = 0.835 |
| ViT-based DR grading model [135] | Classification | DR | Histogram equalization; image translation augmentation; random rotation | Kaggle Diabetic Retinopathy Detection | Acc = 0.914, Sen = 0.926, Spe = 0.977, AUC = 0.988, F1 = 0.923, Precision = 0.928, QWK = 0.935 |
| Swin-MCSFNet [136] | Classification | Fundus image quality | Removal of non-retinal borders; image resizing to 256 × 256; random vertical flipping; random horizontal flipping; random rotation from −180° to 180°; intensity normalization | EyeQ; OIA-ODIR; EyePACS | EyeQ: Acc = 0.877, F1 = 0.776, Precision = 0.792, Recall = 0.766, AUC (micro) = 0.930, AUC (Good) = 0.960, AUC (Usable) = 0.810, AUC (Reject) = 0.960; OIA-ODIR: Acc = 0.798, F1 = 0.644, Precision = 0.639, Recall = 0.665; EyePACS: Acc = 0.749, F1 = 0.627, Precision = 0.616, Recall = 0.664 |
| RETFound [54] | Classification / Prediction | Ophthalmic and systemic diseases | NR | APTOS-2019; IDRiD; MESSIDOR-2; Glaucoma Fundus; PAPILA; Retina; JSIEC; OCTID; UK Biobank; MEH-AlzEye | DR classification: APTOS-2019 (AUC = 0.943), IDRiD (AUC = 0.822), MESSIDOR-2 (AUC = 0.884); Wet AMD 1-year conversion prediction: CFP (AUC = 0.862), OCT (AUC = 0.799); 3-year systemic disease prediction: myocardial infarction (CFP AUC = 0.737), heart failure (CFP AUC = 0.794), ischemic stroke (CFP AUC = 0.754), Parkinson's disease (CFP AUC = 0.669) |
| FunduSAM [137] | Segmentation | Optic disc and optic cup | Image centering; optic-disc ROI cropping; alignment; polar coordinate transformation; inverse polar-coordinate transformation of output mask | REFUGE | REFUGE: Optic disc (DSC = 0.961, IoU = 0.882), optic cup (DSC = 0.867, IoU = 0.789) |
| Serp-Mamba [120] | Segmentation | Retinal vessels | Exclusion of invalid regions using official masks; uniform resizing to 1024 × 1024 | PRIME-FP20; MU-VS Center A; MU-VS Center B | PRIME-FP20: DSC = 0.690, IoU = 0.529, MCC = 0.684, BM = 0.658; MU-VS Center A: DSC = 0.611, IoU = 0.440, MCC = 0.601, BM = 0.584; MU-VS Center B: DSC = 0.571, IoU = 0.401, MCC = 0.563, BM = 0.539 |

Table 2: Representative Algorithms in the Evolution of CFP Analysis (continued)

| Method | Task | Target | Preprocessing | Dataset | Performance |
|---|---|---|---|---|---|
| FLAIR [138] | Detection / Classification | Retinal pathology and disease categories | Image resizing to 512 × 512; zero padding for rectangular images; intensity normalization to [0, 1]; data augmentation using horizontal flipping, random rotation within [-5°, 5°], scaling within [0.9, 1.1], and color jittering | MESSIDOR; FIVES; REFUGE; 20×3; ODIR200×3; MMAC; ACRIMA; DeepDRiD; MMAC-B; RP-MHL-2; CAT-MYA-2 | Domain-shift zero-shot: MESSIDOR (ACA = 0.604, Kappa = 0.772), FIVES (ACA = 0.735), REFUGE (ACA = 0.920); unseen-category zero-shot: 20×3 (ACA = 0.983), ODIR200×3 (ACA = 0.667), MMAC (ACA = 0.400); linear-probe ACA: MESSIDOR = 0.719, FIVES = 0.879, REFUGE = 0.843, 20×3 = 1.000, ODIR = 0.935, MMAC = 0.740, average = 0.852 |
| RetiZero [118] | Classification / Retrieval | Common and rare fundus diseases | Text standardization to a unified prompt template; retention of only text precisely matched to color fundus photographs; exclusion of blurred, overexposed, underexposed, montage, local-field, monochromatic, and externally annotated images; label denoising | EYE-15; EYE-52; H1; H2; H3; rH1; rH2; rH3; SIC; SUSTech; SiMES; SINDI | Zero-shot classification: EYE-15 (Top-1/Top-3/Top-5 = 0.442/0.702/0.840), EYE-52 (Top-1/Top-3/Top-5 = 0.360/0.626/0.756); retrieval: EYE-15 (Top-1/Top-3/Top-5 = 0.854/0.928/0.950), EYE-52 (Top-1/Top-3/Top-5 = 0.726/0.843/0.886); internal-domain AUC: H1/H2/H3 = 0.997/0.980/0.993; few-shot AUC = 0.967/0.859/0.942; cross-domain internal AUC: rH1/rH2/rH3 = 0.998/0.986/0.990; external test AUC ≥ 0.912; clinical assistance accuracy: 0.552–0.628 |
| DeepDR Plus [139] | Prediction | DR progression risk | Exclusion of ungradable images; image resizing to 512 × 512 for pretraining and fine-tuning; extraction of two 512 × 512 crops per iteration; data augmentation using random compression, random blurring, brightness jittering, contrast jittering, random gamma correction, additive Gaussian noise, and random rotation | DRPS; ECHM; WTHM; NDSP; CUHK-STDR; PUDM; SEED; SiDRP; BJHC | Internal validation: fundus model (C-index = 0.823), combined model (C-index = 0.833), metadata model (C-index = 0.696); external cohorts: fundus model (C-index = 0.786–0.802); 1-year progression prediction: internal fundus model (C-index = 0.823–0.862, IBS = 0.049–0.161), external cohorts (C-index = 0.804–0.837, IBS = 0.066–0.170); 1-year AUC: internal = 0.826–0.865, external = 0.722–0.863; $R^2$ = 0.678; mean individualized screening interval = 31.970 months; screening frequency reduction = 0.625; VTDR delayed detection rate = 0.002 |
| FedMHA-DR [140] | Classification | DR | Image resizing to 224 × 224; minority-class augmentation using rotation, flipping, and mild color jittering; additional augmentation during training using rotation within ±15°, flipping, and mild brightness and contrast adjustment; class-weighted loss; per-round feature regeneration and normalization for electronic health record variables | RetinaMNIST | AUC = 0.764, F1 = 0.711, Precision = 0.705, Recall = 0.718 |
| Time-Series VLM for DR Forecasting [141] | Prediction | Referable DR or maculopathy progression | Use of current macula-centered fundus image; automatic generation of twenty semantically equivalent prompt templates; random sampling of historical sequence length; appending prior best-corrected visual acuity and previous diabetic retinopathy grade when available | Internal hold-out test set | 1-year: AUC = 0.691, Acc = 0.631, F1 = 0.365, Sen = 0.725, Spe = 0.535; 2-year: AUC = 0.674, Acc = 0.611, F1 = 0.357, Sen = 0.609, Spe = 0.591; 3-year: AUC = 0.671, Acc = 0.610, F1 = 0.382, Sen = 0.595, Spe = 0.543 |

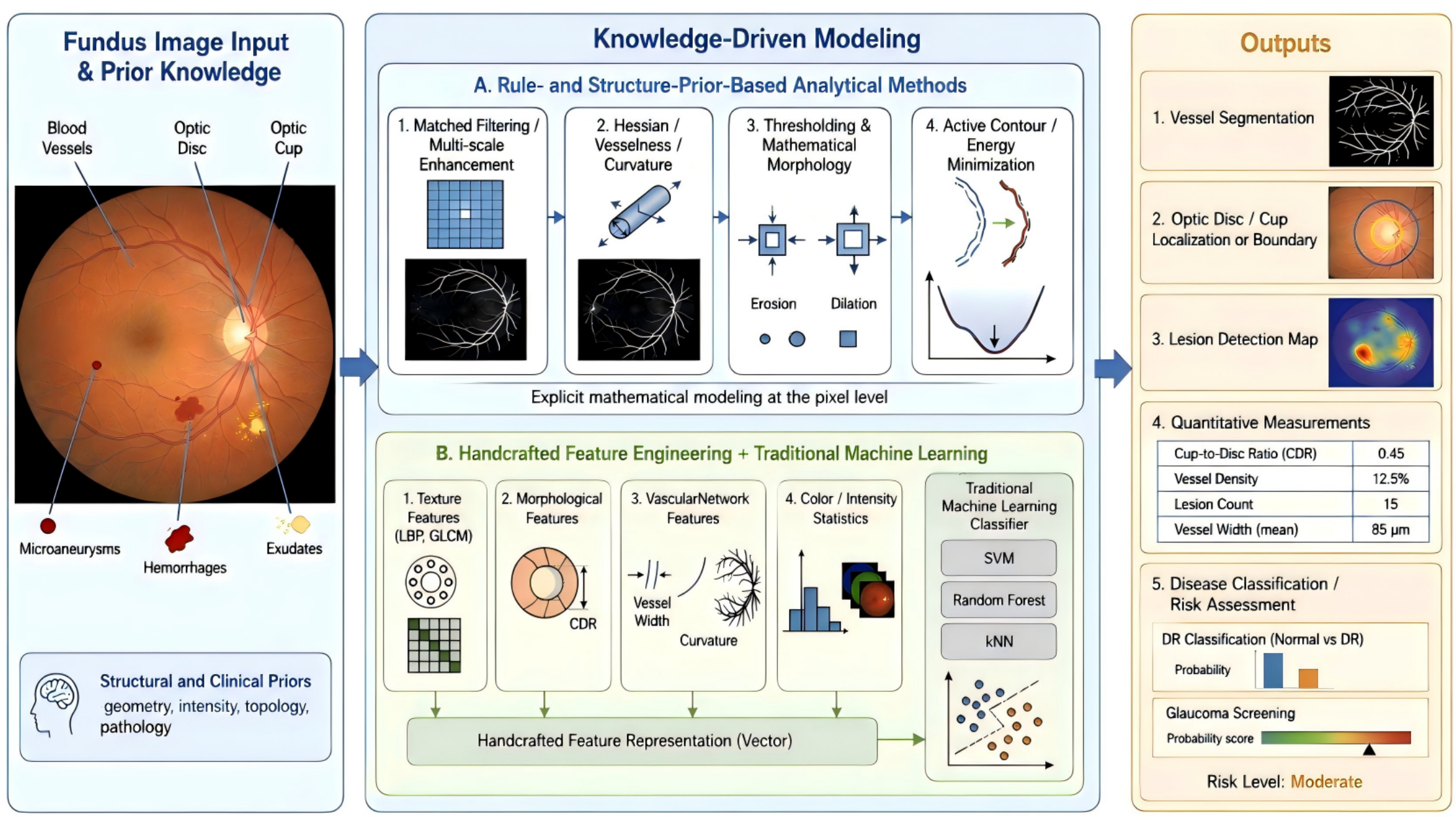

**Figure 15:** Knowledge-driven CFP Analysis Pipeline.

## 4.1. Knowledge-Driven Methods

Prior to the widespread adoption of deep learning approaches, CFP analysis predominantly relied on knowledge-driven analytical paradigms, as summarized in previous surveys and reviews [7, 142, 143]. In this stage, ophthalmic expert knowledge was incorporated into algorithm design by translating retinal anatomical and pathological priors into explicit rules or computable features. For retinal vessel analysis, early methods modeled vessels using matched filtering [14, 144], ridge- or profile-based representations [28], mathematical morphology [122, 145], vesselness or curvature-related responses [146, 147], and active contour models [123]. Optic disc localization and segmentation were often based on brightness, shape, and morphological priors [124, 148]. These studies were mainly evaluated on early public datasets, with primary attention devoted to vessel segmentation on DRIVE, STARE, REVIEW, and CHASE_DB1 [28, 123, 145, 149], optic disc localization and segmentation on DRIVE and DIARETDB1 [124, 148], and limited morphological quantification tasks such as vessel width measurement on REVIEW [123]. Overall, although these methods offered clear interpretability and well-defined mechanisms, their performance was often sensitive to handcrafted parameter settings, image quality, illumination variation, and acquisition conditions [145, 147, 150].

**Rule- and Structure-Prior-Based Analytical Methods**

Early studies predominantly adopted explicit mathematical modeling approaches to directly characterize the geometric morphology, intensity distribution, and topological properties of fundus structures [14, 122, 143, 145]. Representative techniques include vessel enhancement methods based on matched filtering or multi-scale enhancement [14, 144, 147], structure extraction approaches grounded in thresholding and mathematical morphology [14, 145], as well as vessel-like structure detection methods incorporating Hessian-based or curvature-related information [122, 144, 146]. These methods typically operate at the pixel level and do not rely on large-scale annotated data for training, thereby demonstrating clear feasibility in early research settings with limited samples [14, 121, 122, 145]; however, they are inherently sensitive to image quality, parameter configurations, and acquisition conditions, and often struggle to maintain robust performance in scenarios involving complex noise, low contrast, or variations in imaging conditions [146, 147].

**Methods Based on Handcrafted Feature Engineering and Traditional Machine Learning**

With the introduction of statistical learning theory, research gradually evolved toward a two-stage analytical framework of “feature extraction–classifier learning" [126, 142, 151]. Under this paradigm, handcrafted features—such as texture descriptors, morphological characteristics, color statistics, and vascular network representations—are first extracted from CFP [125, 142, 152], and subsequently fed into support vector machines, ensemble learning models, or other conventional classifiers to accomplish disease recognition and structural discrimination [125, 126, 153]. Compared with purely rule-based methods, this approach integrates explicit feature representation with statistical classifiers, thereby advancing tasks such as glaucoma screening, diabetic retinopathy detection, and vascular analysis[125, 129, 151–154]; however, its feature design remains heavily constrained by task-specific definitions and empirical assumptions, resulting in limited robustness and generalization across varying data distributions and imaging conditions [155].

**Stage Summary and Methodological Limitations**

Knowledge-driven methods for CFP analysis historically established a foundational cornerstone in early research, providing the structural templates and methodological basis for vessel segmentation, lesion characterization, and quantitative morphometry [7, 121, 142, 143]. However, their core bottleneck stems from a profound dependence on handcrafted priors and an absolute, deterministic reliance on manual pixel-level preprocessing paradigms. Fixed morphological rules and manually engineered feature descriptors are inherently insufficient to capture the intense non-linear complexity and heterogeneous manifestations of retinal pathologies, particularly when lesions present with high variability in scale, contrast, morphology, and imaging contexts [151]. Consequently, the operational success of these traditional frameworks was entirely entangled with brittle, expert-designed pre-processing stages—including global illumination correction, localized contrast enhancement via CLAHE, and binary field-of-view masking—along with dataset-specific heuristic parameter tuning [145, 147, 150]. This tight coupling between hand-crafted preprocessing and rigid algorithmic execution rendered the entire pipeline highly vulnerable to image quality degradation and data distribution shifts. These coupled limitations, in turn, defined a restrictive methodological boundary, directly catalyzing the subsequent paradigm shift toward data-driven systems capable of transferring processing burdens from explicit pixel manipulation to learning hierarchical, discriminative representations directly from raw visual pixels [116, 151].

## 4.2. CNN Based Methods

Against this background, convolutional neural networks (CNNs) were systematically introduced into fundus color image analysis with the advancement of computational resources, the accumulation of task-specific public datasets, and the successful transfer of deep learning techniques from natural image analysis to medical imaging [116, 151]. These studies were evaluated on representative datasets corresponding to different algorithmic tasks, including disease recognition and lesion detection [156–158], retinal vessel segmentation on DRIVE, STARE, CHASE_DB1, and HRF [159–161], and optic disc/cup segmentation on MESSIDOR, ORIGA, REFUGE, DRISHTI-GS, and RIM-ONE [130, 162, 163]. Methodologically, CNNs introduced an end-to-end representation learning paradigm, in which the traditionally separated stages of handcrafted feature extraction and prediction were integrated within a unified deep network architecture [116, 156, 157]. Concurrently, this algorithmic evolution catalyzed a programmatic shift in preprocessing paradigms, transitioning from rigid manual pixel tuning to highly scalable, learning-based data engineering pipelines. Rather than relying on manually designed low-level features, CNN-based models learned discriminative visual representations directly from raw pixels through supervised hierarchical convolutional encoding [156, 157, 164]. To sustain the training convergence and robustness of these data-hungry backbones, contemporary preprocessing established systematized pipelines encompassing deep-learning-based automated image quality assessment (IQA) to filter motion artifacts, localized region-of-interest (ROI) detection to standardize visual inputs, and extensive online geometric and color-space data augmentations to explicitly counteract severe class imbalances. This multi-axial shift marks a fundamental transition in CFP analysis from knowledge-driven feature engineering to data-driven representation learning, and establishes a unified data-preprocessing-algorithmic foundation for subsequent architectural innovations focusing on network depth, multi-scale representation, pixel-level segmentation, and deployment efficiency [116, 164, 165].

**Deep Representation Learning and Residual Learning**

In the early phase of applying CNNs to fundus color image analysis, studies mainly examined whether deep networks could learn clinically discriminative representations from fundus images without relying on handcrafted features, particularly in diabetic retinopathy and glaucoma recognition [156–158], as summarized in subsequent reviews [116]. With the development of deeper CNN architectures in general computer vision, fundus studies increasingly adopted deep convolutional backbones for disease recognition and lesion analysis [156–158, 164]. However, increasing network depth also introduced optimization difficulties, motivating the use of residual learning [31], channel recalibration mechanisms, and more stable modular or compound-scaled designs [166, 167]. These architectural advances improved the trainability and representational capacity of CNNs, and established reusable backbone architectures for subsequent CFP analysis tasks, including disease classification, lesion analysis, and structure segmentation [164, 165].

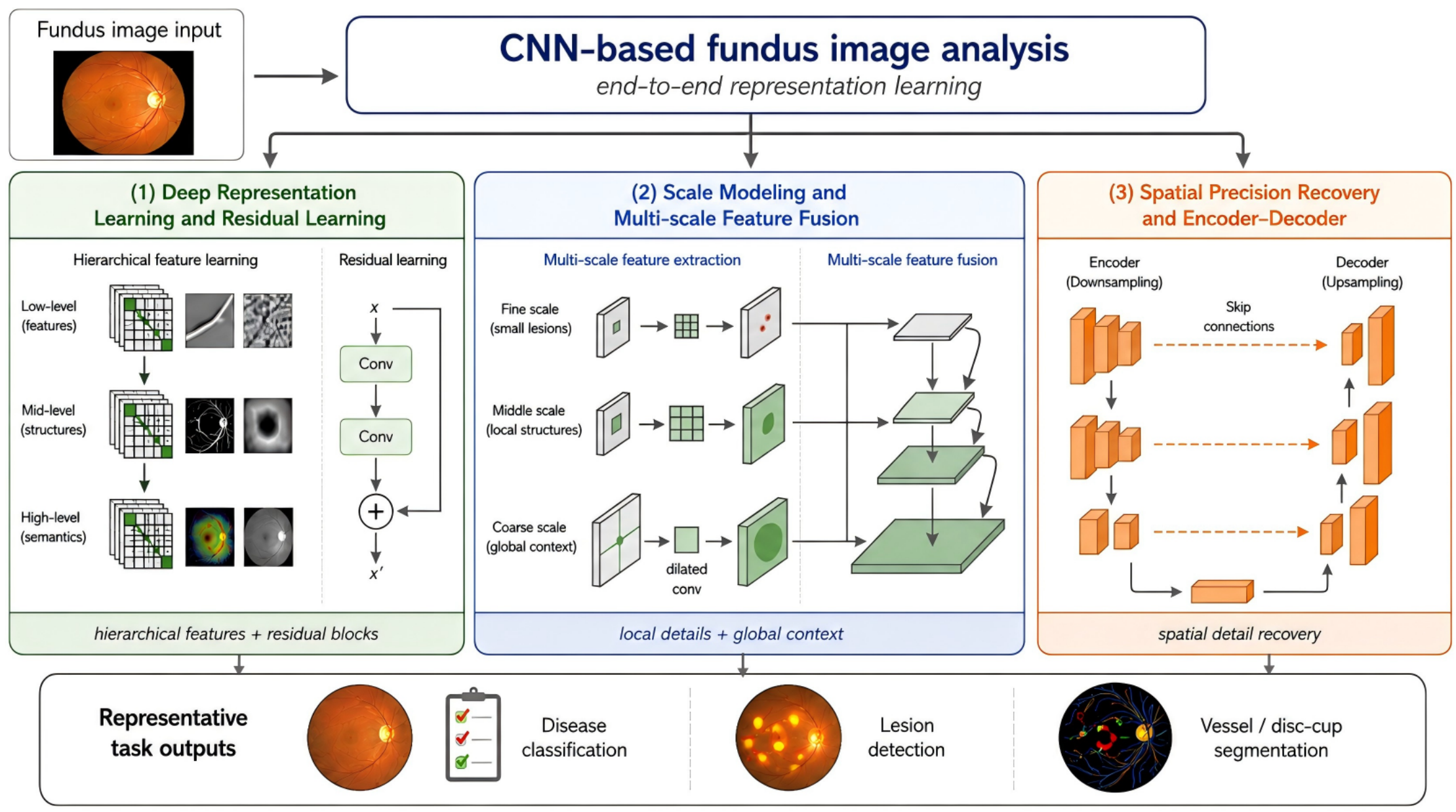

**Figure 16:** CNN-based End-to-end Representation Learning for CFP Analysis.

**Scale Modeling and Multi-Scale Feature Fusion**

Once CNN backbones had been widely adopted, scale heterogeneity became a central methodological challenge in CFP analysis. Micro-lesions, fine vascular structures, and larger anatomical targets such as the optic disc and optic cup often coexist within the same image, making single-path feature extraction insufficient for capturing both fine-grained local details and broader contextual cues [159, 160, 162]. This observation imposed more stringent requirements on receptive field design for small structures such as retinal vessels [159–161], feature pyramid-based modeling for lesion-related detection tasks such as diabetic macular edema [168], and cross-level feature fusion for larger anatomical targets such as the optic disc and optic cup [130, 162]. In response, multi-scale modeling became a key direction in CNN-based CFP analysis, with representative designs including bi-directional feature pyramid and attention-based fusion for diabetic macular edema detection [168], parallel or multi-resolution feature extraction pathways for retinal vessel representation [159, 160], and hierarchical feature fusion strategies for optic disc/cup analysis [130, 162]. These designs enabled CNNs to better integrate local structural details and global contextual information, thereby supporting scale-sensitive fundus analysis tasks, including lesion detection [168], retinal vessel segmentation [159–161], and joint analysis of the optic disc and optic cup [130, 162].

**Spatial Precision Recovery and Encoder–Decoder**

Beyond scale adaptation, pixel-level fundus image analysis further requires precise spatial recovery. CNNs were therefore increasingly extended from image-level recognition and detection to dense prediction tasks, including retinal vessel segmentation [169] and optic disc/cup segmentation [162, 163, 170]. Compared with classification tasks, these tasks impose substantially higher demands on spatial localization accuracy, because fine vascular structures and anatomical boundaries must be preserved at the pixel level [163, 169, 170]. However, standard CNN architectures inevitably lose spatial details during repeated downsampling operations, thereby limiting their effectiveness in modeling fine-grained structures [171–173]. To address this issue, encoder–decoder architectures were introduced into fundus image segmentation, building on symmetric downsampling–upsampling pathways and skip connections to combine high-level semantic information with progressively restored spatial resolution [171, 173]. In fundus applications, such designs and their variants have been used to preserve fine vascular structures [169, 174] and delineate the boundaries of the optic disc and optic cup [130, 162, 163]. As a result, CNN-based CFP analysis was extended from image-level recognition to pixel-level structural interpretation.

**Stage Summary and Methodological Boundaries**

The proliferation of convolutional neural networks (CNNs) fundamentally institutionalized a mature and efficient data-driven representation framework in CFP analysis, directly driven by critical breakthroughs in deep representation learning, scale adaptation, and spatial precision recovery. Through these architectural advances [156, 157], CNNs substantially elevated automated diagnostics from coarse image-level disease recognition [116, 130, 164] to fine-grained, pixel-level structural interpretation [162, 163, 169]. Bound to their local inductive biases and hierarchical feature extraction mechanisms, CNN-based topologies retain favorable computational efficiency compared with global modeling architectures, remaining highly practical for mainstream high-throughput screening and dense segmentation scenarios. Simultaneously, the operational reliability of these networks remained tightly coupled with scalable data engineering preprocessing pipelines, where deep automated image quality assessment (IQA) networks and localized region-of-interest (ROI) cropping served as essential components to standardize visual inputs and control clinical noise.

Nevertheless, the methodological boundaries of this paradigm remain distinctly constrained. At the algorithmic level, the core computational backbone relies on convolutional operations restricted by local receptive fields, which dictates that long-range spatial dependencies can only be captured indirectly through deep stacking or multi-scale feature aggregation [119]. This structural limitation becomes increasingly pronounced when critical diagnostic cues are loosely distributed across distinct anatomical regions, or when localized lesions necessitate joint contextual reasoning with the global retinal landscape. Parallelly, at the preprocessing level, conventional online data augmentations (e.g., rigid spatial rotations and basic intensity shifts) and heuristic quality-control filters fail to fundamentally resolve severe cross-center style variations and heterogeneous vendor domain shifts. Compounded by a rigid reliance on high-quality annotated cohorts, persistent sensitivity to population shifts, and constrained interpretability, these multi-axial bottlenecks collectively demarcate the performance ceiling of CNN-driven expert systems, inevitably motivating the subsequent paradigm transition toward global contextual modeling and self-adaptive foundation paradigms [51, 116, 151].

## 4.3. Vision-Based Unimodal Global Modeling Methods

While convolutional neural networks establish an efficient framework for localized retinal features, their reliance on local receptive fields introduces a locality bias, necessitating deep stacking or multi-scale aggregation to capture long-range spatial dependencies [116, 119]. This structural limitation becomes evident in complex diagnostic scenarios requiring cross-regional contextual reasoning, such as glaucoma staging or global lesion-to-context synthesis [119, 165]. Against this backdrop, the Vision Transformer (ViT) was introduced to reformulate fundus images as token sequences via self-attention mechanisms, shifting the algorithmic paradigm from local hierarchical representation toward explicit global modeling [175]. Crucially, this algorithmic transition has been heavily fueled by self-supervised visual pretraining on million-scale, unlabeled heterogeneous data resources (e.g., RETFound [54] and SynFundus-1M [20]), which enables modern ophthalmic vision foundation models (VFMs) to learn highly generalized retinal representations that transfer efficiently to diverse downstream clinical tasks.

Correspondingly, this algorithmic leap profoundly reframed the role of preprocessing pipelines, shifting the engineering focus from semantic image preservation to hardware-aware computational adaptation. In this paradigm, traditional explicit image enhancement (e.g., contrast tuning or structural filtering) became increasingly neutralized, as foundation architectures internalized domain and sensor variability through massive pretraining. Instead, the preprocessing axis pivoted toward token-level structural preparation, introducing specialized patch tokenization, dynamic multi-resolution scaling, and flexible token-pruning strategies designed explicitly to reconcile the catastrophic quadratic computational and memory overhead of processing high-resolution fundus photographs. Recent research within this vision-only foundation paradigm has further branched into recovering fine-grained structures via dense prediction [176–178], evaluating domain adaptation limits [179–181], and adopting state space models (SSMs) [120, 182]. In these emergent SSM/Mamba pipelines, the definition of preprocessing has uniquely expanded to encompass advanced multi-directional sequence scan trajectories to flatten two-dimensional fundus topological maps into linear tracking streams. Together, these cross-axial developments extend unimodal CFP analysis from local representations to self-adaptive global modeling, establishing the final performance ceiling of pure vision before transitioning to multimodal AI.

**Global Modeling as the Fundamental Computational Unit**

The key transformation introduced by ViT lies in reformulating two-dimensional images as sequences of “tokens". The standard pipeline typically involves partitioning a fundus image into non-overlapping patches of fixed size, projecting each patch linearly into a patch token, and incorporating positional encoding to preserve spatial structural information. The resulting token sequence is then fed into a Transformer encoder, where multi-head self-attention explicitly computes the relationships between any given token and all others, enabling the integration of local evidence and global context in a single step [175, 183]. This modeling paradigm provides a more direct mechanism for capturing cross-regional relationships in tasks such as glaucoma assessment, diabetic retinopathy grading, and multi-disease recognition, and further establishes global structural relationships as a central, learnable component within the network[119, 135, 184, 185].

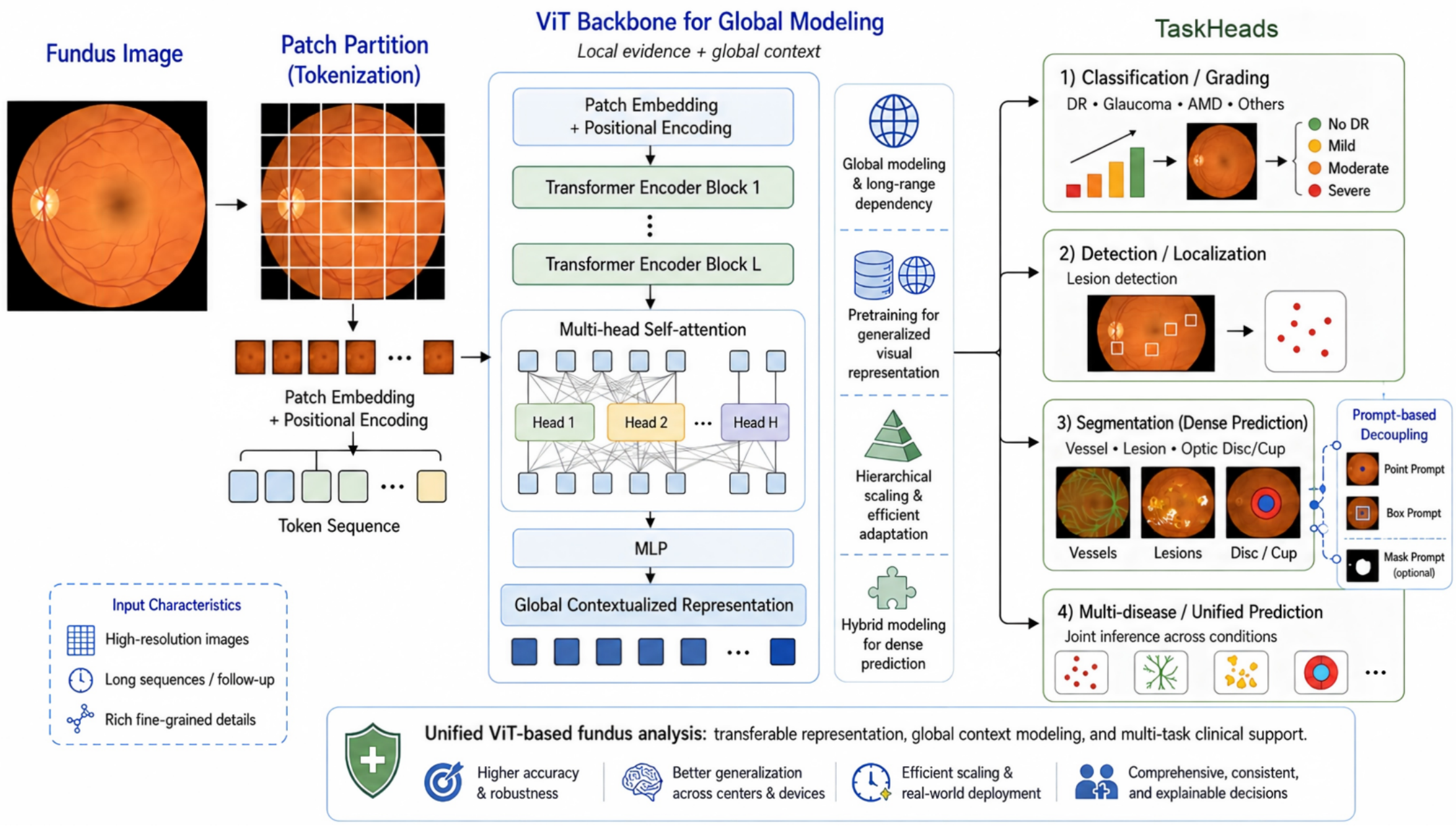

**Figure 17:** Vision-Based Unimodal Global Modeling Methods.

**Data Efficiency and Pretraining Paradigm**

Compared with CNNs, ViTs weaken strong inductive biases such as translation invariance, and therefore typically rely more heavily on large-scale data, data augmentation, and regularization strategies to learn stable visual priors [175, 186].

As a result, the “pretraining–fine-tuning" paradigm has become a crucial bridge for introducing ViTs into medical imaging and fundus-related tasks: models are first pretrained on large-scale natural image datasets or unlabeled retinal images to acquire general-purpose representations, and subsequently fine-tuned on target fundus datasets to adapt to specific tasks and label systems [54, 186, 187]. In scenarios characterized by limited annotations or long-tailed class distributions, self-supervised learning further amplifies the value of this paradigm. In particular, strategies such as masked image modeling—where patches are masked and reconstructed—encourage the model to learn more transferable structural and contextual semantics [188]. In the context of fundus imaging, such pretraining approaches provide more robust initialization for few-shot fine-tuning, reduce dependence on ImageNet-based pretrained weights, and enhance cross-dataset adaptability [54, 187].

**Computational Complexity Constraints and Hierarchical Adaptation**

The computational complexity of native self-attention typically grows quadratically with the number of tokens, leading to a substantial increase in inference cost and memory consumption when processing high-resolution fundus images [175, 183, 189]. In response to this practical constraint, a central direction of architectural evolution has been to preserve the “global relational modeling capability" while maintaining “acceptable computational cost". Representative approaches include dynamic computation compression based on adaptive tokens [189], complexity control through hierarchical or progressively structured token representations [183, 185], and the design of efficient Transformer variants tailored for real-world fundus imaging tasks [190]. These developments highlight that the practical deployment of Transformer-based models in fundus image analysis must jointly consider high-resolution inputs, resource constraints, and the requirements of clinical workflows [190].

**Local Detail Recovery and Dense Prediction**

In fundus imaging tasks, a substantial portion of critical evidence manifests as fine-grained structures or boundary information, such as thin vascular networks, optic disc and optic cup boundaries, and various micro-lesions associated with diabetic retinopathy. However, the patch-based tokenization in ViT may cause these subtle structures to be fragmented, diluted, or weakened at early stages of representation [175, 176]. To mitigate this issue, researchers have progressively integrated Transformer architectures into dense prediction frameworks. On one hand, encoder designs have been enhanced through local structure augmentation, texture-aware modeling, and finer-grained cross-layer feature fusion [177, 178, 191]; on the other hand, the U-shaped architectural paradigm has been adopted in the decoder to restore spatial resolution and boundary precision, enabling applications such as vessel segmentation, lesion segmentation, and joint optic disc/cup segmentation [176, 178, 191]. The central objective of this line of research is to ensure that global relational modeling capabilities effectively contribute to precise localization of fine anatomical and pathological structures [176–178].

**CNN–Transformer Hybrid Modeling**

Although ViTs demonstrate clear advantages in global modeling, their relatively weak local inductive biases often necessitate convolutional components to effectively capture fine-grained textures and subtle structural details; conversely, pure CNNs struggle to explicitly model long-range cross-regional dependencies [175, 192, 193]. To leverage the complementary strengths of both paradigms, CNN–Transformer hybrid architectures have gradually emerged as an important direction in fundus image analysis. Such approaches typically adopt a collaborative design in which convolutional modules are responsible for local representation, while Transformer modules capture global relationships. Through mechanisms such as dual-branch architectures, hierarchical coupling, or feature-level interactions, these models enable joint modeling of local details and global dependencies [175, 192, 193]. In tasks such as diabetic retinopathy grading and vessel segmentation, this hybrid strategy often demonstrates more robust and well-balanced performance [192–194].

**General Visual Representation and Task Decoupling**

Following the significant enhancement of global modeling capabilities brought by ViT and its variants, another key shift in unimodal vision lies in "task decoupling". Traditional deep learning approaches typically follow a paradigm of "one model per task," whereas the introduction of foundation models seeks to learn more generalizable visual representations through large-scale unified pretraining, and to partially decouple downstream tasks from model parameters via lightweight adaptation mechanisms [54, 181]. In CFP analysis, this direction is reflected both in retinal foundation models designed for disease detection [54], and in general-purpose vision models—exemplified by Segment Anything—that can perform segmentation via prompt-driven mechanisms, along with their adaptations to fundus imaging tasks [137, 179, 180]. It is important to note, however, that general-purpose representations do not necessarily guarantee stable performance gains in medical scenarios. When foundation models are primarily trained on natural image distributions, they may exhibit insufficient sensitivity to low-contrast regions, micro-scale targets, and domain-specific artifacts commonly found in medical images. Moreover, their performance often depends heavily on prompt quality, interaction strategies, and domain adaptation approaches [137, 180, 181]. Therefore, at the level of this review, such approaches are more appropriately positioned as a "paradigm upgrade in unimodal visual modeling" while they substantially enhance model generality and transferability, their effective integration into fundus imaging applications still requires domain-specific data, well-designed prompting protocols, and rigorous reliability evaluation to achieve a clinically viable closed-loop system [54, 137, 181].

**Efficient Sequence Modeling and Long-Range Dependency**

With global modeling becoming a foundational capability, efficiency has once again emerged as a critical constraint for clinical deployment. When processing high-resolution images or longer sequences, Vision Transformers incur rapidly increasing computational and memory costs as sequence length grows. In contrast, state space models (SSMs) and their visual variants (e.g., the Mamba family) provide an alternative pathway for long-range dependency modeling with near-linear complexity, enabled by selective state spaces and recurrent state updates [195]. In the context of fundus color image analysis, this line of research is currently reflected primarily in high-resolution vascular structure modeling and efficient long-range association scenarios. Works such as Serp-Mamba, HM-Mamba, and HCM have demonstrated their potential from perspectives including high-resolution ultra-widefield scanning laser ophthalmoscopy (UWF-SLO) vessel continuity, multi-scale tubular structure modeling, and semi-supervised hybrid CNN–Mamba segmentation, respectively [120, 182, 196]. However, as an emerging architectural paradigm, existing fundus-related studies remain largely concentrated on method-level performance comparisons within specific segmentation benchmarks. Evidence regarding cross-center generalization, external validation, and standardized engineering reporting remains limited [120, 182, 196]. Therefore, a more cautious positioning in this review is that SSM/Mamba-based approaches represent a promising new paradigm for high-resolution and long-range dependency modeling in fundus image analysis, yet their true clinical benefits still require further confirmation through more rigorous external validation and standardized evaluation frameworks [197–199].

**Capability Spillover Driven by Global Modeling**

Once global relational modeling becomes a foundational capability, the value of unimodal visual models extends beyond improvements in single-task accuracy, toward supporting more comprehensive functionalities aligned with clinical workflows. In multi-disease or comorbidity-related recognition scenarios, models are able to integrate diverse pathological cues distributed across the entire image within a single inference process, thereby enabling joint decision-making that more closely resembles real clinical practice [54, 200]. In tasks involving cross-regional structural evidence, such global representations are also more conducive to aggregating dispersed information and forming stable inferences [119]. In “gatekeeping" tasks such as image quality assessment, global representations facilitate the holistic evaluation of factors including blurring, occlusion, specular reflection, and field-of-view deficiencies, thereby reducing the likelihood of invalid samples entering downstream diagnostic pipelines [136, 201]. In disease progression prediction, long-term risk modeling based on unimodal visual inputs has begun to demonstrate potential, providing more stable visual cues for follow-up analysis and prognostic assessment [54, 139].

**Stage Summary and Methodological Boundaries**

In perspective, the architectural evolution of unimodal visual modeling driven by Vision Transformers does not represent a simplistic obsolescence of CNNs, but rather a profound paradigm maturation centered on the explicit formulation of global dependencies. This technological trajectory unfolds across multiple critical dimensions, including data efficiency via self-supervised pretraining, computational feasibility via streamlined attention, and fine-grained structural fidelity via advanced dense prediction frameworks [175, 177, 178, 190]. Against this background, vision foundation models (VFMs) have substantially expanded the horizons of generalized representation learning [54, 181], while state space models (SSMs) fundamentally restructure the computational paradigm to reconcile the efficiency bottlenecks associated with high-resolution inputs and long-range sequence modeling [120, 182], collectively establishing a new performance ceiling for unimodal CFP analysis. Crucially, this algorithmic progression was paralleled by a strategic redirection of preprocessing pipelines; as foundation architectures internalized domain and sensor variability through massive pretraining, the engineering focus shifted from traditional manual pixel enhancement to hardware-aware computational adaptation, introducing architectural-level patch tokenization, dynamic multi-resolution scaling, and sequence scan trajectories.

Nevertheless, the methodological boundaries of this paradigm remain distinct. The computational and memory footprints imposed by high-resolution CFP images persist as formidable barriers to large-scale deployment; although efficient attention mechanisms and linearized SSM preprocessing alleviate quadratic complexity to some extent, achieving deterministic performance gains across heterogeneous imaging devices still heavily relies on rigorous engineering reporting and external validation [120, 175, 190]. Furthermore, the intrinsic patch-based tokenization mechanism within ViTs inherently dampens sensitivity to minute pathological structures; while dense prediction decoders and hybrid topologies actively counteract this degradation, they inevitably incur substantial architectural overhead [175, 177, 178, 192, 193]. Most fundamentally, even when equipped with augmented global reasoning and robust token-level segmentation capacities, unimodal vision-centric systems operate under a closed-world assumption bounded strictly by intra-image evidence. They lack the capacity for explicit alignment with critical external clinical context—such as patient histories, systemic examination findings, and demographic risk factors. This systemic visual isolation, coupled with the inability of pure visual preprocessing to capture non-image risk markers, logically motivates the subsequent transition toward multimodal joint-modeling and heterogeneous data preparation paradigms [54, 202, 203].

### 4.4. Multimodal-Modeling Methods

As unimodal visual modeling approaches its structural and performance limits, its inherent constraints become evident. In real-world clinical settings, disease grading and risk stratification typically require integrating multimodal evidence—including imaging findings, patient history, laboratory results, and systemic biomarkers. Consequently, analytical paradigms relying solely on visual input are inherently limited in fully characterizing disease states [117, 204, 205]. To reconcile these boundaries, the research focus has shifted from "intra-visual modeling" toward the "joint modeling of heterogeneous information." This algorithmic transition is seamlessly coupled with an evolution in the data dimension toward text-paired and EHR-integrated ocular cohorts, as exemplified by the report-paired corpora in RET-CLIP [202] and EyeCLIP [203], and the longitudinal clinical risk factors in DeepDR Plus [139].

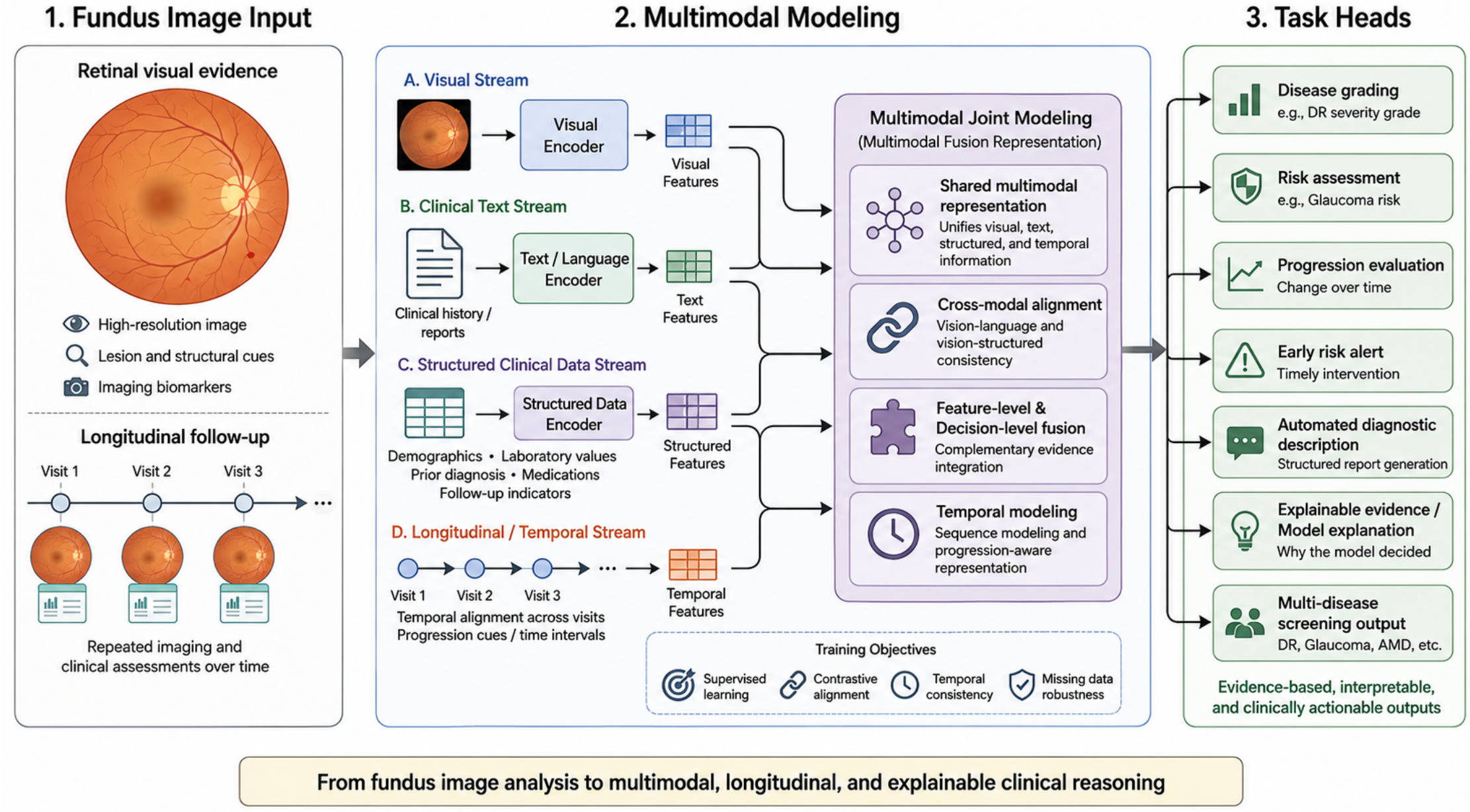

**Figure 18:** Multimodal Modeling Framework for CFP Analysis.

Inevitably, this multi-modal expansion has forced a profound paradigm shift in preprocessing pipelines, stretching the definition of data preparation far beyond explicit pixel manipulation to encompass heterogeneous cross-modal standardization. Within this advanced lifecycle, preprocessing has evolved into a critical multi-axial workflow to sustain semantic alignment. It demands text-level preprocessing—including text normalization and prompt template crafting—to standardize unstructured clinical diagnostic reports, alongside self-supervised missing-value imputation to resolve the severe sparseness inherent to tabular electronic health records (EHRs). By leveraging these robustly preprocessed data resources, the training objective transitions from isolated visual representation to cross-modal semantic alignment and multi-source collaborative inference [117, 202, 203]. Building upon this tri-axial transition, this section systematically reviews multimodal models according to their methodological evolution.

**Vision–Language Joint Modeling**

An important direction in multimodal fundus analysis is vision–language joint modeling. Such approaches integrate visual encoders with language models or text encoders, enabling fundus color images to be mapped into a shared representation space that supports text generation, semantic retrieval, and language-based reasoning [138, 203, 206]. Unlike conventional visual models that primarily output classification or grading results, vision–language models aim to explicitly organize visual evidence in natural language form, thereby providing outputs that are more aligned with clinical communication in terms of result presentation, interaction, and interpretability [206–208]. In fundus imaging tasks, these models have been applied to image–text alignment pretraining, automatic generation of diagnostic descriptions, knowledge-enhanced disease recognition, and the construction of analysis interfaces that approximate conversational usage scenarios [138, 202, 203, 208]. By structuring visual evidence through language, such models partially address the limitation of traditional deep learning systems that "produce conclusions without explaining the reasoning process," and lay the groundwork for downstream applications such as report generation, human–AI collaborative annotation, and clinical question answering [203, 206, 208].

From a methodological perspective, however, vision–language joint modeling still faces several key challenges. The quality of language outputs is highly dependent on the alignment between images and text, as well as on the organization of domain knowledge; when visual evidence is insufficient or training corpora are biased, models may generate descriptions that appear plausible but lack sound medical justification. In addition, the absence of unified standards for language output formats, evaluation metrics, and safety constraints across different studies makes it difficult to directly compare performance and reliability [206, 209]. Accordingly, in the context of CFP analysis, vision–language models are more appropriately regarded as an emerging direction for capability expansion rather than a fully mature and clinically reliable solution [203, 206, 209].

**Vision–Structured Clinical Data Fusion**

Beyond language information, the joint modeling of fundus color images with structured clinical data represents another important trajectory in the evolution of multimodal methodologies. Such approaches typically incorporate numerical and categorical variables—such as patient age, sex, disease duration, laboratory indicators, risk factors, and prior diagnoses—into the learning and inference process alongside visual features [117, 140, 204]. Compared with purely image-based analysis, this paradigm more closely reflects real-world clinical decision-making, where diagnostic judgments are formed through the integration of imaging evidence and patient-specific contextual information [117, 204]. In tasks such as diabetic retinopathy grading, risk assessment, and related predictive modeling, the fusion of visual and structured data often helps mitigate ambiguities inherent in image-only interpretation, and provides more robust decision support in borderline cases or risk stratification scenarios [117, 140, 204]. From an implementation perspective, existing studies primarily explore feature-level fusion and attention-guided fusion strategies, aiming to effectively exploit complementary information from different modalities within a unified latent space [117, 140, 204].

It should be noted, however, that structured clinical data exhibit substantial variability across centers in terms of availability, completeness, and recording standards. Issues such as missing data and modality absence may significantly affect model training and generalization performance [210]. Furthermore, multimodal learning may suffer from modality imbalance, where one modality dominates the optimization process, thereby limiting the effective utilization of others [211]. Consequently, while vision–structured clinical data fusion has demonstrated clear potential, its robustness under varying data quality conditions and degrees of modality completeness still requires more systematic evaluation.

**Longitudinal Multimodal Joint Modeling**

In the management and follow-up of chronic ocular diseases, diagnostic and intervention decisions often rely on the integrated analysis of information collected at multiple time points. Compared with single-visit assessments, longitudinal multimodal modeling seeks to jointly analyze fundus color images from different time points together with corresponding textual records, clinical indicators, or other auxiliary examinations, in order to capture the dynamic progression of disease [141, 212–214]. Such approaches typically treat the temporal dimension as an additional modeling axis, employing sequence modeling, temporal attention, or time-aware multimodal fusion mechanisms to align and jointly infer visual and non-visual information along the time axis [141, 213, 214]. In tasks such as diabetic retinopathy progression assessment, age-related macular degeneration (AMD) progression prediction, and glaucoma progression monitoring, longitudinal multimodal models have demonstrated potential advantages in trend identification, risk stratification, and early warning [141, 212–214]. However, longitudinal multimodal modeling imposes higher requirements on data completeness and consistency. Missing time points, absent modalities, and heterogeneity in clinical records within follow-up data may significantly affect model training and generalization performance [210]. In addition, modality imbalance in multimodal learning may weaken the effective collaboration among different information sources [211]. Consequently, compared with static multimodal fusion, longitudinal modeling is more dependent on standardized data construction, temporal external validation, and transparent reporting frameworks [197, 198].

**Stage Summary and Methodological Boundaries**

Ultimately, the transition into the multimodal artificial intelligence era establishes a critical inflection point in CFP analysis, effectively shifting the paradigm from "vision-driven perception" to "joint modeling of heterogeneous information". By incorporating language and structured clinical data, models demonstrate capabilities beyond unimodal approaches in result representation, risk stratification, and longitudinal assessment, thereby aligning automated analysis more closely with real-world clinical decision-making processes [117, 203, 204, 213, 214]. Simultaneously, this architectural leap relies on a fundamentally reshaped preprocessing infrastructure, where the robust standardization of non-image evidence – specifically advanced structured text normalization and self-supervised missing-value tabular imputation – serves as the mathematical prerequisite for coherent cross-modal alignment.

Nevertheless, this stage is characterized by clearly defined methodological boundaries. Multimodal models are highly sensitive to data alignment and quality, and their generalization across centers is typically more challenging than that of unimodal approaches [117, 210]. At the preprocessing level, the persistent sparsity, irregular sampling, and structural noise inherent to longitudinal electronic health records (EHRs) mean that sub-optimal data imputation or flawed prompt crafting can severely degrade downstream feature fusion. Furthermore, the complementary contributions of different modalities and the precise sources of performance gains are not always readily interpretable, and modality imbalance may further diminish the benefits of joint modeling [140, 211]. In addition, existing studies still exhibit limitations in reproducibility, evaluation design, and reporting standards, which constrain direct comparisons across methods [197–199]. Consequently, multimodal methodologies remain in a phase where exploration and standardization advance in parallel, and their maturation in practical applications depends on more systematic data construction, rigorous heterogeneous preprocessing frameworks, and comprehensive clinical validation.

## 4.5. Summary

This chapter systematically reviews the methodological evolution of color fundus photography (CFP) analysis algorithms since the early 2000s, tracing a historical trajectory from knowledge-driven explicit modeling to data-driven convolutional representations, global contextual modeling, and ultimately multimodal collaborative intelligence. This technological paradigm shift does not represent a simplistic linear replacement of obsolete techniques, but rather a continuous, tri-axial realignment of data foundations, preprocessing paradigms, and modeling priorities engineered to reconcile the intrinsic constraints of each developmental milestone.

Methodologically, this historical evolution unfolds through successive paradigm-level reconfigurations. The inaugural transition from handcrafted morphological priors to inductive-biased convolutional networks transferred the feature-extraction burden to deep models, simultaneously restructuring preprocessing from rigid manual pixel tuning—such as heuristic contrast enhancement and masking—into scalable, automated data-engineering pipelines. This convolutional dominance subsequently yielded to global relational modeling represented by Vision Transformers, a shift that systematically overcame local receptive field constraints. Within this unimodal global paradigm, traditional explicit image enhancement became increasingly internalized through massive self-supervised pretraining, shifting the preprocessing axis toward architectural-level structural preparation, including hardware-aware patch tokenization and dynamic multi-resolution scaling. As pure visual modeling approached its performance ceiling, the frontier inevitably expanded toward multimodal joint architectures by embedding text-paired and electronic health record (EHR) integrated ocular cohorts. Consequently, the definition of preprocessing underwent a radical abstraction beyond the visual domain to encompass structured clinical text standardization and self-supervised missing-value tabular imputation, ultimately transitioning the analytical focus from the isolated image toward the entirety of robustly prepared clinical evidence.

Despite these substantial architectural advances, the current technological landscape remains bounded by profound, interlocking methodological tensions. A primary discord exists between model expressiveness and clinical generalizability, where severe performance degradation under multi-center and multi-device conditions reveals that deep networks still capitalize on distribution-specific statistical artifacts rather than invariant pathophysiological representations, underscoring the necessity for domain-agnostic preprocessing safeguards. This bottleneck is further compounded by a critical tension between automated precision and decision trustworthiness, where emergent cross-modal and foundation architectures introduce risks of hallucinated outputs and uncalibrated uncertainty propagation, vulnerabilities heavily amplified by upstream structural noise and unaligned cross-modal preprocessing synchronizations. Most fundamentally, a severe friction persists between theoretical algorithmic capacity and practical deployment constraints, as parameter-heavy systems impose prohibitive computational footprints. This operational reality firmly establishes efficiency-oriented preprocessing, token-level data management, and resource awareness as intrinsic engineering design constraints rather than post hoc empirical optimizations.

In conclusion, future breakthroughs in CFP analysis will not stem from the continuous stacking of increasingly complex model architectures, but from a systematic reconfiguration of its collective methodological foundations. Achieving robust cross-domain deployment, interrogable decision mechanisms, and resource-efficient clinical translation fundamentally demands the coordinated optimization of data lifecycles, rigorous preprocessing frameworks, standardized evaluation protocols, and algorithmic backbones. As the discipline transitions toward paradigm-level consolidation, establishing unified, reproducible, and clinically aligned evaluation frameworks constitutes an absolute prerequisite for unlocking the long-term clinical utility and technological sustainability of artificial intelligence in real-world ophthalmic workflows.

## 5. Conclusion and Future Outlook

### 5.1. Conclusion

This review has systematically charted the multi-dimensional landscape of Color Fundus Photography (CFP) analysis, decoding its technological trajectory through the continuous, tri-axial interplay of dataset evolution, preprocessing paradigms, and algorithmic frameworks.

Initially, the domain was anchored by localized clinical tasks, operating on small-scale, public diagnostic datasets (e.g., DRIVE, MESSIDOR) with sparse annotations. To address fundamental image variability within these tightly bounded cohorts, traditional preprocessing techniques relied heavily on handcrafted pixel-level transformations—such as global histogram equalization and localized contrast-limited adaptive augmentation (CLAHE)—which served as a vital pipeline to feeding knowledge-driven, hand-crafted algorithms. These early methodologies laid the foundational structural understandings for retinal geometric descriptors but were chronically vulnerable to brittle cross-dataset generalization.

Concurrently, the sudden transition into the deep convolutional era was fueled by a landmark consolidation and scaling of multi-disease data repositories (e.g., EyePACS, APTOS). This data explosion necessitated a programmatic shift in data preparation, introducing robust semantic data augmentation and automated quality control pipelines to mitigate sensor artifacts and class imbalances. Backed by these rigorous data engineering pipelines, convolutional neural networks (CNNs) institutionalized end-to-end representation learning and multi-scale feature aggregation, successfully elevating automated CFP analysis from macro image-level screening to precise pixel-level structural segmentation.

More recently, the ongoing paradigm shift has pushed the boundaries toward open-world foundation learning and heterogeneous evidence synthesis. In the unimodal visual domain, the emergence of million-scale pretraining datasets has catalyzed a decisive departure from specialized pipelines. Facilitated by advanced token-level interactions and linear-complexity tracking, Vision Transformers (ViTs), Vision Foundation Models (VFMs), and State Space Models (SSMs) have effectively bypassed the localized inductive biases of CNNs, establishing a new performance ceiling capable of robust zero-shot downstream task adaptation.

At the contemporary frontier, the field has aggressively dismantled the systemic visual isolation inherent to unimodal systems. The data dimension has fundamentally evolved from isolated raw visual pixels into highly integrated, text-paired, and EHR-linked multi-modal registries (exemplified by advanced cross-modal frameworks like RET-CLIP, EyeCLIP, and DeepDR Plus). This data-dimension evolution has forced a paradigm shift in data preparation—transitioning from traditional spatial image enhancements to structured text standardization, automated prompt templates, and missing-value tabular imputation. Consequently, contemporary algorithmic frameworks have evolved from isolated visual representation to deep cross-modal semantic alignment and multi-source collaborative reasoning.

In synthesis, this comprehensive review reveals that the upper bound of clinical diagnostic performance in CFP analysis is no longer bounded strictly by pixel-resolution scaling or algorithmic parameter depth, but rather by the collaborative optimization across the entire life cycle of data curation, structural quality assurance, and multimodal context synthesis. Bridging the gap between benchmark-centric performance and real-world clinical deployment remains the ultimate milestone for the next generation of ophthalmic artificial intelligence.

### 5.2. Future Outlook

To further accelerate the transition of automated CFP analysis from theoretical benchmarks into safe, reliable, and holistic clinical deployment, future research must converge on multiple deeply interconnected, multi-dimensional frontiers.

Unified Multimodal Standardization and Missing-Value Robustness constitutes the primary imperative as the data dimension transitions from unimodal images to heterogeneous vision-language-EHR cohorts. The field currently suffers from acute non-standardization, siloed registries, and severe missingness in clinical variables, demanding rigorous benchmarks for multimodal data curation that focus on standardized text normalization for expert diagnostic reports and robust cross-modal alignment protocols. Methodologically, instead of relying on brittle heuristic imputations for tabular electronic health records, future architectures must integrate deep generative imputation and self-supervised masked modeling to natively handle highly incomplete clinical data streams without sacrificing diagnostic certainty.

This heterogeneous data alignment is inextricably linked to the need for High-Fidelity Localized Tokenization Balanced with Edge-Computing Preprocessing. While VFMs and SSM variants offer unprecedented global contextual reasoning, their intrinsic patch-based tokenization mechanisms naturally dilute fine-grained sensitivity to minute micro-pathological structures, such as early-stage microaneurysms or subtle drusen clusters. Conversely, feeding ultra-high-resolution fundus images directly into deep foundation backbones incurs prohibitive computational overhead. Future engineering must therefore develop joint pipelines where lightweight, hardware-aware preprocessing modules—such as localized region-of-interest cropping or automated high-pass contrast enhancement—work in synergy with hybrid, multi-granularity model architectures. Exploring token pruning, dynamic sparse attention, and tailored linear Mamba topologies is paramount to facilitating resource-constrained edge-device deployment in high-throughput community screening scenarios.

Beyond the refinement of baseline architectures, Generative Counterfactual Augmentation for Data-Scarce Ocular Pathologies represents a critical frontier to remedy the persistent data scarcity for rare ophthalmic diseases and specific systemic cardiovascular or renal manifestations in the fundus. Future paradigms should shift from basic spatial data augmentation toward state-of-the-art latent diffusion models (LDMs) capable of generating anatomically coherent, high-fidelity synthetic fundus photographs. Crucially, these generative frameworks should be deployed as tools for counterfactual evaluation. By synthesizing controllable, step-by-step lesion progressions on specific longitudinal patient backgrounds, researchers can build bias-free causal validation frameworks to systematically assess model fairness across diverse ethnic populations and rare clinical phenotypes.

Ultimately, the technical viability of these paradigms must be validated through Explainable Cross-Modal Synergy and Causal Clinical Alignment. Ophthalmic multimodal AI must transcend simplistic late-fusion heuristics or uninterpretable black-box token concatenation. Future research must prioritize the development of deeply aligned, cross-modal generative architectures—such as specialized ophthalmic Large Multimodal Models (LMMs)—capable of generating structured, clinically rigorous diagnostic rationales. This requires implementing explicit cross-attention mechanisms that anchor textual clinical tokens directly to pixel-level visual evidence. More importantly, future visual-clinical architectures should integrate causal graphical models to ensure that the alignment between ocular phenotypes and systemic risks reflects true pathological causation rather than superficial statistical correlations, thereby fostering absolute clinical trust in real-world decision-making workflows.

## Declaration of generative AI and AI-assisted technologies in the writing process

During the preparation of this work, the author(s) used generative AI tools in order to check and improve the grammar and clarity of the text. After using this tool/service, the author(s) reviewed and edited the content as needed and take(s) full responsibility for the content of the published article.

## Acknowledegments

The authors are grateful to the support of the Guangdong Key Disciplines Project (2024ZDJS137).

## CRediT authorship contribution statement

**Yongsheng Luo**: Data curation, Formal analysis, Visualization, Writing – original draft, Writing – review & editing, Conceptualization. **Wengan He**: Data curation, Formal analysis, Visualization, Writing – original draft, Writing – review & editing, Supervision, Project administration, Conceptualization. **Wenhui Xu**: Data curation, Formal analysis, Visualization, Writing – original draft, Writing – review & editing. **Lihong Jiang**: Data curation, Formal analysis, Visualization, Writing – original draft, Writing – review & editing. **Fan Xiao**: Data curation, Writing – original draft. **Zhuohang Huang**: Data curation, Writing – original draft. **Yuanzhu Liang**: Data curation, Writing – original draft. **Jiayi Liu**: Data curation, Writing – original draft. **Yuxi Chen**: Data curation, Writing – original draft. **Yu Li**: Funding acquisition, Supervision, Project administration, Conceptualization.

# A. Summary of the Dataset

Table 3: Summary of the Dataset

| No. | Dataset | Size | Task | Annotation | Label | Year | Ref. |
|---|---|---|---|---|---|---|---|
| 1 | STARE | 400 | Vessel segmentation | Vessel contour | Pixel-level | 2000 | [13, 14] |
| 2 | DRIVE | 40 | Vessel segmentation | Vessel contour | Pixel-level | 2004 | [28] |
| 3 | Messidor | 1.2K | DR grading, DME risk assessment | DR grade, DME risk | Image-level | 2008 | [15] |
| 4 | Messidor-2 | 1,748 | DR grading, DME classification | Five-level DR grades, Presence of DME | Image-level | 2014 | [15, 215] |
| 5 | DIARETDB0 | 130 | DR classification, lesion detection | MA, HE, EX, SE references, DR label | Point-level, Image-level | 2006 | [41, 42] |
| 6 | DIARETDB1 | 89 | DR classification, lesion detection | MA, HE, EX, SE references, DR label | Polygon-level, Image-level | 2007 | [42, 43] |
| 7 | DIARETDB1 v2.1 | 89 | DR classification, lesion detection | MA, HE, EX, SE references, DR label | Polygon-level, Image-level | 2012 | [216] |
| 8 | DRIONS-DB | 110 | ONH segmentation | OD contours | Pixel-level | 2008 | [217] |
| 9 | REVIEW | 16 | Vessel segmentation, width measurement | Vessel contour | Pixel-level | 2008 | [218] |
| 10 | ROC | 100 | MA detection | Location and radius of MA | Bounding box (circular) | 2009 | [58] |
| 11 | ORIGA | 650 | OD and OC segmentation | OC and OD contours, CDR, ISNT rule, disc size/tilt, RNFL defects, disc HE, PPA ($\alpha/\beta$), disc notching | Image-level, Pixel-level | 2010 | [219] |
| 12 | CHASE_DB1 | 28 | Vessel segmentation | Vessel contour | Pixel-level | 2012 | [44] |
| 13 | HRF (High-Resolution Fundus) | 45 | Vessel segmentation, OD localization | Vessel contours, OD location and diameter | Pixel-level, Bounding box-level (circular) | 2013 | [45] |
| 14 | ACHIKO-K | 258 | Glaucoma classification, OD and OC segmentation, pathological sign detection, longitudinal analysis | Glaucoma label, OC and OD contour, HE, ONH drusen; disc notching, PPA, optic nerve cupping, lamina cribrosa visibility, etc. | Image-level, Pixel-level | 2013 | [220] |
| 15 | RITE (derived from DRIVE) | 40 | Artery and vein segmentation | Artery and vein contours | Pixel-level | 2013 | [221] |
| 16 | e-ophtha | 463 | DR lesion segmentation | MA and EX contours | Pixel-level | 2013 | [215] |
| 17 | Drishti-GS | 101 | OD and OC segmentation, glaucoma classification | OC and OD contours, Presence of glaucoma | Pixel-level, Image-level | 2014 | [222] |
| 18 | Diabetic Retinopathy Detection (EyePACS/Kaggle) | 88,696 | DR grading | Five-level DR grades | Image-level | 2015 | [46] |
| 19 | Diabetic Retinopathy 2015 Data Colored Resized (derived from EyePACS) | 35,126 | DR grading | Five-level DR grades | Image-level | 2015 | [223] |
| 20 | FIRE (Fundus Image Registration Dataset) | 129 | Image registration | Corresponding control point pairs | Point-level | 2017 | [47] |
| 21 | DR HAGIS | 39 | Vessel segmentation | Vessel contours | Pixel-level | 2017 | [48] |
| 22 | IDRiD (Indian Diabetic Retinopathy Image Dataset) | 516 | Lesion segmentation, OD segmentation, fovea detection, DR grading; DME grading | MA, HE, EX, SE contours, OD contour, disc center, fovea localization, DR and DME grades | Image-level, Pixel-level, Point-level | 2018 | [16] |
| 23 | APTOS-2019 Blindness Detection | 5,590 | DR grading | Five-level DR grades | Image-level | 2019 | [49] |

Table 3: Summary of the Dataset (continued)

| No. | Dataset | Size | Task | Annotation | Label | Year | Ref. |
|---|---|---|---|---|---|---|---|
| 24 | Retinal Fundus Multi-disease Image Dataset (RFMiD) | 3,200 | Multi-disease classification | Normal, DR, glaucoma, and other diseases (46 classes) | Image-level | 2019 | [224] |
| 25 | OIA-ODIR | 10,000 | Multi-disease classification | Normal, DR, glaucoma, cataract, AMD, HR, myopia, and other diseases | Image-level | 2019 | [50] |
| 26 | Diabetic Retinopathy 224×224 Gaussian Filtered | 3,662 | DR grading | Five-level DR grades | Image-level | 2019 | [225] |
| 27 | Diabetic Retinopathy (resized) | 35,126 | DR grading | Five-level DR grades | Image-level | 2019 | [226] |
| 28 | Diabetic Retinopathy Diagnosis Dataset | 1,505 | DR classification | Presence of DR | Image-level | 2019 | [227] |
| 29 | AVRDB | 100 | Artery and vein segmentation | Artery and vein contours | Pixel-level | 2019 | [228] |
| 30 | RIM-ONE v1 | 169 | Glaucoma grading, ONH segmentation | ONH contours, glaucoma grades | Pixel-level, Image-level | 2011 | [229] |
| 31 | RIM-ONE v2 | 455 | OD segmentation | Disc contours | Pixel-level | 2014 | [229] |
| 32 | RIM-ONE v3 | 159 | OD and OC segmentation | OD and OC contours | Pixel-level | 2015 | [229] |
| 33 | RIM-ONE DL | 485 | Glaucoma classification | Presence of glaucoma | Image-level | 2020 | [229] |
| 34 | IOSTAR | 30 | Vessel segmentation | Vessel contours | Pixel-level | 2015 | [230] |
| 35 | RC-SLO | 40 | Vessel segmentation | Vessel contours | Pixel-level | 2015 | [230] |
| 36 | KORA-S4 | Not publicly available (2,840 participants) | AMD grading | AMD grades | Image-level | 2016 | [231] |
| 37 | REFUGE | 1,200 | Glaucoma classification, OD and OC segmentation | Presence of glaucoma, OD and OC contours, fovea location | Image-level, Pixel-level | 2018 | [232] |
| 38 | LAG | 5,824 | Glaucoma classification | Presence of glaucoma, attention maps | Image-level, Attention-level | 2019 | [233] |
| 39 | G1020 | 1,020 | Glaucoma classification, OD and OC segmentation, CDR computation, neuroretinal rim analysis | Glaucoma labels, OD and OC contours, OD bounding box; vertical CDR, neuroretinal rim in four quadrants | Image-level, Pixel-level, Bounding box-level | 2020 | [234] |
| 40 | ADAM | 1,200 | AMD classification, OD and lesion segmentation | Presence of AMD, OD contours, fovea coordinates, five lesion contours | Image-level, Pixel-level | 2020 | [57] |
| 41 | Hypertension & Hypertensive Retinopathy | 1,424 | HR classification | Presence of HR | Image-level | 2020 | [235] |
| 42 | Diabetic Retinopathy Balanced (derived from Diabetic Retinopathy (resized)) | 34,792 | DR grading | Five-level DR grades | Image-level | 2020 | [236] |
| 43 | FIVES | 800 | Vessel segmentation, multi-disease classification, image quality assessment | Vessel contours, disease labels (healthy, AMD, DR, glaucoma), image quality scores (illumination/color distortion, blur, low contrast) | Pixel-level, Image-level | 2021 | [237] |
| 44 | Predicting Human Eye Diseases | 601 | Multi-disease classification | Normal, cataract, glaucoma, and other diseases | Image-level | 2021 | [238] |
| 45 | TREND Database | 82 | Vessel segmentation, image quality assessment | Vessel contours, image quality grades, clinical information | Pixel-level, Image-level | 2021 | [62] |

Table 3: Summary of the Dataset (continued)

| No. | Dataset | Size | Task | Annotation | Label | Year | Ref. |
|---|---|---|---|---|---|---|---|
| 46 | Eye Diseases Classification (derived from IDRiD, Ocular Recognition, HRF, etc.) | 4,217 | Multi-disease classification | Normal, DR, cataract, glaucoma | Image-level | 2021 | [239] |
| 47 | ODIR5K Classification (derived from OIA-ODIR) | 6,393 | Multi-disease classification | Normal, DR, glaucoma, cataract, AMD, HR, myopia, and other diseases | Image-level | 2021 | [240] |
| 48 | Deep-learning-based Segmentation of Fundus Photographs | Segmentation: 369; Classification: 287 | Subretinal fluid segmentation, CSC classification | SRF contours, Presence of CSC | Pixel-level, Image-level | 2021 | [241] |
| 49 | Brasil Glaucoma BrG | 2,000 | Glaucoma classification, OD and OC segmentation | Presence of Glaucoma, OD and OC contours | Image-level, Pixel-level | 2021 | [242] |
| 50 | SMDG-19 (Standardized Multi-Channel Dataset for Glaucoma) | Fundus: 12,449; OCT: 80 | Glaucoma classification, vessel segmentation, OD and OC segmentation | Glaucoma labels, vessel contours, OD and OC contours, clinical data, OCT-based OD and OC boundaries | Image-level, Pixel-level | 2023 | [18] |
| 51 | Retinal Fundus Image 50k | 57,110 | Multi-disease classification | Normal, cataract, DR, glaucoma, HR, myopia, AMD | Image-level | 2024 | [243] |
| 52 | Fundus Glaucoma Detection Data (derived from SMDG-19) | 17,242 | Glaucoma classification | Presence of glaucoma | Image-level | 2023 | [244] |
| 53 | Ocular Toxoplasmosis Fundus Images Dataset | 412 | Toxoplasmosis classification; lesion segmentation | Healthy, active, inactive lesions; lesion contours | Image-level; Pixel-level | 2023 | [245] |
| 54 | Retina Blood Vessel Dataset | 100 | Vessel segmentation | Vessel contours | Pixel-level | 2023 | [246] |
| 55 | DeepDRiD | 2,000 | DR grading; image quality assessment | Five-level DR grades; quality scores; clinical information | Image-level | 2022 | [247] |
| 56 | Eye Disease Image Dataset | 5,318 | Multi-disease classification | DR, glaucoma, high myopia, etc. (10 classes) | Image-level | 2024 | [248] |
| 57 | Evaluation benchmark for natural robustness evaluation of retinal vessel segmentation models (derived from DRIVE, STARE, and CHASE_DB1) | 46,750 | Vessel segmentation | Vessel contours | Pixel-level | 2024 | [249] |
| 58 | Diagnosis of Diabetic Retinopathy | 5,556 | DR classification | Presence of DR | Image-level | 2024 | [250] |
| 59 | Fundus Image Dataset for Retinopathy of Prematurity (ROP) | 1,099 | ROP grading | Five-level ROP grades | Image-level | 2024 | [251] |
| 60 | Fundus-AVSeg | 100 | Vessel segmentation | Vessel contours | Pixel-level | 2024 | [252] |

Table 3: Summary of the Dataset (continued)

| No. | Dataset | Size | Task | Annotation | Label | Year | Ref. |
|---|---|---|---|---|---|---|---|
| 61 | AIROGS (Artificial Intelligence for Robust Glaucoma Screening) | 113,893 | Glaucoma grading | Referable glaucoma, non-referable glaucoma, ungradable | Image-level | 2024 | [51] |
| 62 | PAPILA | 488 | Glaucoma classification; OD and OC segmentation | Presence of glaucoma; OD and OC contours | Image-level; Pixel-level | 2022 | [253] |
| 63 | GAMMA (Glaucoma Grading from Multi-Modality Images Challenge) | 300 pairs (fundus + OCT) | Glaucoma grading; OD and OC segmentation; fovea localization | Normal, early glaucoma, progressive glaucoma; fovea coordinates; OD and OC contours | Image-level; Pixel-level; Point-level | 2022 | [52] |
| 64 | LMOD | Multi-modal (surgical: 2,256; OCT: 3,859; SLO: 10,000; eye photos: 2,432; fundus: 3,386) | Anatomical recognition; disease diagnosis; visual-language question answering | Multi-source annotations including anatomical structures, lesion localization, free-text descriptions, disease grading, biomarkers | Image-level; Pixel-level; Bounding box-level; Text-level | 2025 | [17] |
| 65 | NTUH-Hsinchu Glaucoma Dataset | 1,155 | Glaucoma grading; CDR and RNFL prediction | Normal, pre-perimetric glaucoma, glaucoma; demographic data | Image-level | 2022 | [254] |
| 66 | OCTA-Fundus-DR | 222 (111 OCTA + 111 fundus) | DR grading | No DR, mild DR, moderate DR | Image-level | 2023 | [53] |
| 67 | FundusTuning-CN (paired with ODIR-5K) | 6,393 images ( 100k text instructions) | Visual question answering; visual dialogue | Image labels; diagnostic, treatment, and prevention text | Image-level; Text-level | 2024 | [34] |
| 68 | SynFundus-1M | 1,000,000 | Foundation model pretraining; disease diagnosis | Natural language diagnostic descriptions; region readability labels | Image-level; Text-level | 2023 | [20] |
| 69 | Eyecare-100K | 102,000 image–text pairs | Visual question answering; report generation; diagnosis and grading; lesion localization | Classification/grading labels; QA pairs | Image-level; Text-level | 2025 | [19] |
| 70 | EyeCLIP Pretraining Data | 2,777,593 images; 11,180 reports | Vision–language pretraining | Mostly unlabeled images; angiography reports with structured keywords | Image-level; Text-level | 2024 | [35] |
| 71 | RETFound Pretraining Data | Fundus: 904,170; OCT: 736,442 | Retinal foundation model pretraining | Unlabeled | None | 2023 | [54] |
| 72 | VisionFM Pretraining Data | 3,396,918 | General ophthalmic foundation model pretraining | Unlabeled | None | 2023 | [55] |
| 73 | ACRIMA | 705 | Glaucoma classification | Presence of glaucoma | Image-level | 2021 | [255] |
| 74 | Chaksu | 1,345 | OC and OD segmentation; glaucoma classification | Presence of glaucoma; OC and OD boundaries | Pixel-level; Image-level | 2022 | [256] |

## B. Early-Stage Fundus Datasets Before 2014

Table 4: Early-Stage Fundus Datasets (Before 2014)

| No. | Dataset | Source (Centers) | Source (Country / Institution) | Acquisition Device | Resolution | Disease Coverage | Limitations | Ref |
|---|---|---|---|---|---|---|---|---|
| 1 | STARE | Single-center | USA (UCSD, California) | TopCon TRV-50 fundus camera | 605 × 700 | Vessel-related | Low resolution, subjective annotations, limited sample size | [13, 14] |
| 2 | Messidor | Multi-center | France (Lariboisière Hospital Paris; St Étienne Medical Faculty; LaTIM-CHRU Brest) | Topcon TRC NW6 fundus camera with 3CCD video camera | 1440×960; 2240×1488; 2304×1536 | DR and DME | Large resolution variation, limited lesion annotations | [15] |
| 3 | Messidor-2 | Multi-center | France (Lariboisière Hospital Paris; St Étienne Medical Faculty; LaTIM-CHRU Brest; Brest University Hospital) | Topcon TRC NW6 | 1440×960; 2240×1488; 2304×1536 | DR and DME | Class imbalance; limited population diversity | [15, 215] |
| 4 | DRIVE | Single-center | Netherlands | Canon CR5 | 768 × 584 | Vessel-related | limited sample size, limited disease types | [28] |
| 5 | DIARETDB0 | Single-center | Finland (Kuopio University Hospital) | Not reported | 1500 × 1152 | DR | limited sample size, inconsistent annotations, variable image quality, limited population diversity | [41, 42] |
| 6 | DIARETDB1 | Single-center | Finland (Kuopio University Hospital) | ZEISS FF 450plus with Nikon F5 digital camera | 1500 × 1152 | DR | limited sample size, inconsistent annotations, variable quality, limited population diversity | [42, 43] |
| 7 | DRIONS-DB | Single-center | Spain (Miguel Servet Hospital) | Not reported | 600 × 400 | OD-related | limited sample size, limited annotation diversity | [217] |
| 8 | ROC | Multi-center | NR | Topcon NW100, NW200; Canon CR5-45NM | 768×576; 1058×1061; 1389×1383 | MA-related | Heterogeneous image quality, cross-device variability | [58] |
| 9 | CHASE_DB1 | Single-center | UK (London) | Nidek NM-200-D fundus camera | 1280 × 960 | Vessel-related | Uneven illumination, low vessel-background contrast, central vessel reflex in arterioles, limited sample size | [44] |
| 10 | e-ophtha | Multi-center | France (OPHDIAT network) | Not reported | 1440×960 to 3504×2336 | DR-related lesions | Inconsistent resolution, limited sample size | [215] |
| 11 | HRF (High-Resolution Fundus) | Single-center | Germany (University of Erlangen) | Canon CR-1 | 3504 × 2336 | Vessel- and OD-related | Limited sample size, limited acquisition diversity | [45] |
| 12 | RITE | Single-center | Netherlands | Canon CR5 | 768 × 584 | Vessel-related | Low resolution, limited sample size, limited disease types | [221] |
| 13 | Drishti-GS1 | Single-center | India (Aravind Eye Hospital) | Not reported | 2896 × 1944 | Glaucoma | Limited sample size, limited disease types | [222] |
| 14 | ORIGA | Single-center | Singapore Eye Research Institute | Not reported | Not reported | Glaucoma | Limited sample size, limited annotation diversity | [219] |
| 15 | ACHIKO-K | Single-center | South Korea (Asan Medical Center) | Nikon D80, D90 | 640×480; 2144×1424; 3216×2136 | Glaucoma | Limited sample size, limited population diversity | [220] |
| 16 | REVIEW | Single-center | UK (Sunderland Eye Infirmary) | Canon 60UV film camera; Zeiss fundus camera; JVC 3CCD camera | Multiple (3584×2438, etc.) | Vessel-related | Limited sample size, partial vessel annotation (not full network) | [218] |
| 17 | RIM-ONE v1 | Multi-center | Spain | Nidek AFC-210 with Canon EOS 5D Mark II | 2144 × 1424 | Glaucoma | Limited dataset size, limited population diversity | [229] |
| 18 | RIM-ONE v2 | Multi-center | Spain | Nidek AFC-210 with Canon EOS 5D Mark II | 2144 × 1424 | Glaucoma | Limited dataset size, limited population diversity | [229] |

Table 4: Early-Stage Fundus Datasets (Before 2014) (continued)

| No. | Dataset | Source (Centers) | Source (Country / Institution) | Acquisition Device | Resolution | Disease Coverage | Limitations | Ref |
|---|---|---|---|---|---|---|---|---|
| 19 | DiaRetDB1 v2.1 | Multi-center | Finland | Not reported | Not reported | DR | Limited dataset size, not strictly pixel-level annotations | [216] |

## C. Mid-Stage Fundus Datasets (2015–2020)

Table 5: Mid-Stage Fundus Datasets (2015–2020)

| No. | Dataset | Source (Centers) | Source (Country / Institution) | Acquisition Device | Resolution | Disease Coverage | Limitations | Ref. |
|---|---|---|---|---|---|---|---|---|
| 1 | Diabetic Retinopathy 2015 Data Colored Resized (derived from DR Detection) | Multi-center | EyePACS | Not reported | 224 × 224 | DR | Low image resolution | [223] |
| 2 | Diabetic Retinopathy Detection (EyePACS/Kaggle) | Multi-center | EyePACS | Not reported | Multi-resolution | DR | Cross-device variability; image artifacts; defocus; under/over-exposure | [46] |
| 3 | DR HAGIS | Multi-center | UK (NHS) | Canon CR DGI; Topcon TRC-NW6s; TRC-NW8 | Multiple (4752×3168, etc.) | Vessel-related | Limited sample size; no healthy samples; single-expert annotation | [48] |
| 4 | FIRE | Single-center | Greece (Papageorgiou Hospital) | Nidek AFC-210 fundus camera | 2912 × 2912 | Non-disease-oriented | Limited sample size | [47] |
| 5 | IDRiD | Single-center | India (Sushrusha Eye Clinic) | Kowa VX-10a fundus camera | 4288 × 2848 | DR and DME | Limited sample size; single-source data | [16] |
| 6 | APTOS-2019 Blindness Detection | Multi-center | India (Aravind Eye Care System) | Not reported | Multi-resolution | DR | Only image-level DR grading; lack of lesion-level annotations | [49] |
| 7 | Diabetic Retinopathy (resized) | Multi-center | EyePACS | Not reported | 1024 × 1024 | DR | Resizing may degrade fine lesion details | [226] |
| 8 | Diabetic Retinopathy 224×224 Gaussian Filtered | Multi-center | EyePACS | Not reported | 224 × 224 | DR | Gaussian filtering and resizing may cause detail loss | [225] |
| 9 | Diabetic Retinopathy Diagnosis Dataset | Not reported | Not reported | Not reported | Multi-resolution | DR | Lack of disease labels | [227] |
| 10 | OIA-ODIR | Multi-center | China (Shanggong Medical Technology) | Canon, Zeiss, Kowa, etc. | Multi-resolution | Multi-disease (normal, DR, glaucoma, cataract, AMD, HR, myopia, etc.) | Class imbalance; cross-device variability | [50] |
| 11 | RFMiD | Single-center | India (Sushrusha Eye Clinic) | Topcon OCT-2000; Kowa VX-10; Topcon TRC-NW300 | 2144 × 1424; 4288 × 2848; 2048 × 1536 | Multi-disease (46 diseases) | Cross-device variability; inconsistent resolution | [224] |
| 12 | AVRDB | Single-center | Pakistan (AFIO) | Topcon TRC-NW8 | 1504 × 1000 | Vessel-related | Extremely limited sample size; single-region population; no disease grading | [228] |
| 13 | RIM-ONE DL | Multi-center | Spain | Nidek AFC-210 fundus camera with Canon EOS 5D Mark II | Not reported | Glaucoma | Limited population diversity; no glaucoma severity grading | [229] |
| 14 | Hypertension & Hypertensive Retinopathy | Not reported | Not reported | Not reported | 800 × 800 | HR | Lack of dataset description; no severity grading | [235] |
| 15 | Diabetic Retinopathy Balanced (derived from Diabetic Retinopathy resized) | Multi-center | EyePACS | Not reported | 512 × 512 | DR | Resizing may cause detail loss | [236] |
| 16 | REFUGE | Multi-center | Spain (Granada) | Zeiss Visucam 500; Canon CR-2 | 2124×2056; 1634×1634 | Glaucoma | Large resolution variation; low proportion of glaucoma samples | [232] |
| 17 | G1020 | Single-center | Germany (private clinic) | Not reported | 1944×2108 to 2426×3007 | Glaucoma | Single-center data; limited diversity | [234] |

Table 5: Mid-Stage Fundus Datasets (2015–2020) (continued)

| No. | Dataset | Source (Centers) | Source (Country / Institution) | Acquisition Device | Resolution | Disease Coverage | Limitations | Ref. |
|---|---|---|---|---|---|---|---|---|
| 18 | LAG | Single-center | China (Beijing Tongren Hospital) | Not reported | 500 × 500 | Glaucoma | Single-center data; limited generalizability; no pixel-level annotation | [233] |
| 19 | KORA-S4 | Single-center | Germany (Augsburg region) | Topcon TRC-NW5S | 768 × 576 | AMD | Limited ethnic diversity; lack of longitudinal data | [231] |
| 20 | ADAM | Single-center | China (Zhongshan Ophthalmic Center) | Zeiss Visucam 500; Canon CR-2 | 2124×2056; 1444×1444 | AMD | Imbalanced subtype distribution | [57] |
| 21 | IOSTAR | Multi-center | China and Netherlands | Scanning Laser Ophthalmoscopy | 1024 × 1024 | Vessel-related | Limited sample size; high acquisition cost | [230] |
| 22 | RC-SLO | Multi-center | China and Netherlands | Scanning Laser Ophthalmoscopy | 360 × 320 | Vessel-related | Complex image conditions (artifacts, crossings, reflections) | [230] |
| 23 | RIM-ONE v3 | Multi-center | Spain | Nidek AFC-210 fundus camera with Canon EOS 5D Mark II | 2144 × 1424 | Glaucoma | Limited sample size; limited population diversity | [229] |

## D. Recent Stage Fundus Datasets (2021–Present)

Table 6: Recent Stage Fundus Datasets (2021–Present)

| No. | Dataset | Source (Centers) | Source (Country / Institution) | Acquisition Device | Resolution | Disease Coverage | Limitations | Ref. |
|---|---|---|---|---|---|---|---|---|
| 1 | FIVES | Single-center | China (Second Affiliated Hospital, Zhejiang University) | Topcon TRC-NW8 fundus camera | 2048 × 2048 | Multi-disease (AMD, DR, glaucoma) | Limited population diversity; single acquisition setting | [237] |
| 2 | Predicting Human Eye Diseases | Not reported | Not reported | Not reported | 2464 × 1632 | Multi-disease (cataract, glaucoma) | Unclear data source; limited sample size; uncertain annotation quality; narrow disease spectrum | [238] |
| 3 | TREND Database | Single-center | Montenegro (University of Montenegro) | MiIS HORUS Scope DEC 200 | 2560 × 1920 | Non-disease-oriented | Limited sample size; only young healthy subjects; limited population diversity | [62] |
| 4 | eye_diseases_classification (derived from IDRiD, Oculur recognition, HRF etc.) | Multi-center | Not reported | Not reported | Multi-resolution | Multi-disease (DR, cataract, glaucoma, normal) | Heterogeneous sources; inconsistent annotation standards | [239] |
| 5 | ODIR5K_Classification | Multi-center | China (487 hospitals across 26 provinces) | Not reported | 224 × 224 | Multi-disease (DR, glaucoma, cataract, AMD, hypertension, myopia, etc.) | Resolution reduction may cause detail loss; class imbalance; no pixel-level annotations | [240] |
| 6 | Deep-learning-based Segmentation of Fundus Photographs (CSCR) | Single-center | South Korea (Aerospace Medical Center) | Not reported | 256 × 256 | Single disease focus (CSC) | Limited sample size; weak annotation characteristics | [241] |
| 7 | Brasil Glaucoma (BrG) | Multi-center | Brazil | Welch Allyn 11820 ophthalmoscope | 400 × 400 | Single disease focus (glaucoma) | Single device; low resolution; no disease stage annotation | [242] |
| 8 | SMDG-19 | Multi-center | Not reported | Not reported | 512 × 512 | Single disease focus (glaucoma) | Class imbalance; inconsistent annotation quality | [18] |
| 9 | Fundus Dataset | Multi-center | Not reported | Not reported | 1596×990; 3680×3288 | Multi-disease (DR, glaucoma, cataract, AMD, hypertension, myopia, etc.) | Unclear data source; insufficient dataset description | [257] |
| 10 | Retinal Fundus Image 50K | Multi-center | Not reported | Not reported | 512–1024 | Multi-disease (DR, glaucoma, cataract, AMD, hypertension, myopia) | Unclear source; class imbalance | [243] |
| 11 | Fundus Glaucoma Detection Data (PyTorch format, derived from SMDG-19) | Multi-center | Not reported | Not reported | 512 × 512 | Single disease focus (glaucoma) | Class imbalance; inconsistent annotation quality | [244] |
| 12 | Ocular Toxoplasmosis Fundus Images | Multi-center | Paraguay (Hospital de Clínicas; Pediatric Hospital) | ZEISS VISUCAM 500; Pictor Plus-Portable Ophthalmic Camera | 2124×2056; 1536×1152 | Single disease focus (toxoplasmosis) | Limited sample size; limited diversity; class imbalance | [245] |
| 13 | Retina Blood Vessel Dataset | Multi-center | Not reported | Not reported | 512 × 512 | Multi-disease (DR, AMD) | Low resolution; limited sample size | [246] |
| 14 | Eye Disease Image Dataset | Multi-center | Bangladesh | Topcon TL-211; TRC-50DX | 2004 × 1690 | Multi-disease (DR, glaucoma, macular scar, optic disc edema, CSC, retinal detachment, retinitis pigmentosa, myopia) | Majority augmented data; regional bias | [248] |

Table 6: Recent Stage Fundus Datasets (2021–Present) (continued)

| No. | Dataset | Source (Centers) | Source (Country / Institution) | Acquisition Device | Resolution | Disease Coverage | Limitations | Ref. |
|---|---|---|---|---|---|---|---|---|
| 15 | Evaluation benchmark for natural robustness evaluation of retinal vessel segmentation models (derived from DRIVE, STARE, and CHASE_DB1) | Multi-center | Not reported | Not reported | Multiple | Multi-disease (hypertension, DR, glaucoma) | Inconsistent resolution; limited diversity | [249] |
| 16 | ROP Dataset | Single-center | China (Shenzhen Eye Hospital) | RetCam; SW-8000; Nautilus system | 512 × 512 | Single disease focus (ROP) | Class imbalance | [251] |
| 17 | Diagnosis of Diabetic Retinopathy | Not reported | Not reported | Not reported | 224 × 224 | Single disease focus (DR) | Lack of lesion-level annotations | [250] |
| 18 | Fundus-AVSeg | Single-center | China (Shenzhen Eye Hospital) | ZEISS VISUCAM200; Canon fundus cameras | 2656×1992; 1280×1280 | Multi-disease (DR, AMD, glaucoma) | No disease labels; limited clinical applicability | [252] |
| 19 | AIROGS | Multi-center | USA (500 centers) | Canon, Topcon, Nidek, etc. | 120–1024 | Multi-disease (DR, AMD, glaucoma) | Geographic bias; domain shift between training and testing | [51] |
| 20 | PAPILA | Single-center | Spain | Topcon TRC-NW400 | 2576 × 1934 | Single disease focus (glaucoma) | Missing clinical variables; class imbalance | [253] |
| 21 | GAMMA | Single-center | China (Zhongshan Ophthalmic Center) | Topcon OCT; Kowa; TRC-NW400 | Multi-modal | Single disease focus (glaucoma) | Limited population diversity | [52] |
| 22 | LMOD | Multi-center | Not reported | Not reported | Multi-resolution | Multi-disease (glaucoma, macular hole, cataract, DR, among others) | Incomplete modality coverage; potential bias | [17] |
| 23 | NTUH-Hsinchu Glaucoma Dataset | Single-center | China (NTUH Hsinchu Branch) | Zeiss VISUCAM 524 | 1620 × 1440 | Single disease focus (glaucoma) | Limited sample size; demographic imbalance | [254] |
| 24 | OCTA-Fundus-DR | Single-center | India | Centervue Eidon; Optovue OCTA | Multi-modal | Single disease focus (DR) | Limited sample size | [53] |
| 25 | FundusTuning-CN (used in conjunction with ODIR-5K) | Single-center | China (Beijing Tongren Hospital) | Not reported | Multi-resolution | Multi-disease (DR, AMD, glaucoma, myopia, cataract) | Non-clinical-grade annotations; limited label consistency | [34] |
| 26 | SynFundus-1M | Multi-center | China | Not reported | Multi-resolution | Multi-disease (DR, AMD, optic nerve abnormalities, chorioretinal diseases, degenerative myopia, DME, epiretinal membrane, glaucoma, hypertensive retinopathy, pathological myopia, RVO) | Noisy labels (AI-generated); limited rare cases | [20] |
| 27 | Eyecare-100K | Multi-center | Global | Not reported | Multi-resolution | Multi-disease (common fundus diseases) | Modality imbalance; LLM-generated text may introduce hallucination | [19] |
| 28 | EyeCLIP Pretraining Data | Multi-center | China and global datasets | Multiple devices | Multi-resolution | Multi-disease (common ocular diseases and 17 rare diseases) | Text generation noise; long-tail scarcity; missing systemic labels | [35] |
| 29 | RETFound Pretraining Data | Single-center | UK (Moorfields Eye Hospital) | Topcon; Optos; Heidelberg | Multi-resolution | Multi-disease (DR, glaucoma, AMD, among others) | Limited population diversity; limited modalities | [54] |
| 30 | VisionFM Pretraining Data | Multi-center | Global (26 countries) | Multiple devices | Not reported | Multi-disease (DR, glaucoma, AMD, cataract, myopia, RVO, RD, HR) | Synthetic data artifacts; loss of 3D information; limited rare disease coverage | [55] |

Table 6: Recent Stage Fundus Datasets (2021–Present) (continued)

| No. | Dataset | Source (Centers) | Source (Country / Institution) | Acquisition Device | Resolution | Disease Coverage | Limitations | Ref. |
|---|---|---|---|---|---|---|---|---|
| 31 | ACRIMA | Single-center | Spain | Not reported | 178–1420 | Single disease focus (glaucoma) | Limited sample size; lack of detailed annotations | [255] |
| 32 | Chaksu | Single-center | India | Mobile fundus cameras | Multiple | Single disease focus (glaucoma) | Limited population diversity | [256] |

## E. Comprehensive Summary of Preprocessing Methods for Fundus Color Images

Table 7: Comprehensive Summary of Preprocessing Methods for Fundus Color Images

| No. | Category | Method | Principle | Advantages | Limitations | References |
|---|---|---|---|---|---|---|
| 1 | Traditional preprocessing methods | Contrast enhancement (HE) | Adjusts the image histogram to make gray-level distribution more uniform | Enhances overall image contrast | May cause grayscale distortion and detail loss | [73] |
| 2 | Traditional preprocessing methods | Contrast enhancement (AHE) | Adaptively computes histograms in local regions | Enhances local contrast | Over-amplifies noise in homogeneous regions | [73] |
| 3 | Traditional preprocessing methods | Contrast enhancement (CLAHE) | Limits local histogram contrast and interpolates results | Suppresses over-enhancement and stabilizes local contrast improvement | Depends on parameter settings | [73] |
| 4 | Traditional preprocessing methods | Contrast enhancement (ESIHE spatial histogram equalization) | Optimizes clipping threshold to balance gray-level stretching | Improves adaptability to low exposure and suppresses noise | Limits contrast enhancement range | [258] |
| 5 | Traditional preprocessing methods | Contrast enhancement (Rayleigh distribution CLAHE) | Dynamically adjusts clipping threshold based on Rayleigh distribution | Enhances illumination adaptability while preserving color and reducing artifacts and noise amplification | Depends on parameter settings | [12, 75] |
| 6 | Traditional preprocessing methods | Contrast enhancement (Channel extraction) | Extracts the most informative color channel | Focuses on key information and simplifies processing, improving lesion discriminability | May lose multispectral information | [154] |
| 7 | Traditional preprocessing methods | Contrast enhancement (Normalization) | Standardizes brightness and color distribution | Reduces imaging variation and unifies data scale | Parameter-dependent and difficult to eliminate complex noise | [79] |
| 8 | Traditional preprocessing methods | Brightness adjustment (Random Brightness Adjustment, RBA) | Randomly adjusts global brightness | Enhances illumination diversity and robustness | May cause feature distortion and ignore spatial variation, leading to local information loss | [78] |
| 9 | Traditional preprocessing methods | Brightness adjustment (Hue Saturation Value, HSV) | Adjusts hue and saturation distribution | Reduces color bias and enhances regional contrast | May weaken bright lesions or marker features | [76] |
| 10 | Traditional preprocessing methods | Brightness adjustment (Iterative Robust Homomorphic Surface Fitting, IRHSF) | Separates illumination and reflectance components, with channel-wise correction to remove wavelength-dependent attenuation | Suppresses illumination interference and enhances structural stability | May introduce reflectance estimation bias | [77] |
| 11 | Traditional preprocessing methods | Smoothing & denoising – spatial filtering – linear (Mean filter) | Replaces central pixel with average gray value of neighborhood | Efficiently reduces Gaussian noise with low computational cost | Blurs vessels and lesion edges | [79] |
| 12 | Traditional preprocessing methods | Smoothing & denoising – spatial filtering – linear (Gaussian filter) | Applies distance-weighted averaging for smoothing | Robust to uniform noise | Weakens contrast of fine structures such as vessels | [79] |
| 13 | Traditional preprocessing methods | Smoothing & denoising – spatial filtering – linear (Wiener filter) | Dynamically adjusts filter coefficients based on minimum mean square error | Balances denoising and deblurring while preserving details | Strongly depends on noise power estimation | [79, 81] |
| 14 | Traditional preprocessing methods | Smoothing & denoising – spatial filtering – nonlinear (Median filter) | Sorts neighborhood values and replaces center with median | Effectively removes salt-and-pepper noise while preserving edges | Weak for Gaussian noise; may damage details at high noise density | [12, 22] |

Table 7: Comprehensive Summary of Preprocessing Methods for Fundus Color Images (continued)

| No. | Category | Method | Principle | Advantages | Limitations | References |
|---|---|---|---|---|---|---|
| 15 | Traditional preprocessing methods | Smoothing & denoising – spatial filtering – nonlinear (Weighted median filter) | Assigns weights to neighborhood pixels and replaces center with weighted median after removing extremes | Better preservation of retinal structures and effective salt-and-pepper noise removal | May lose fine structures at high noise density; weak for Gaussian noise; increased complexity with larger windows | [12, 259] |
| 16 | Traditional preprocessing methods | Smoothing & denoising – spatial filtering – nonlinear (Adaptive median filter) | Dynamically adjusts filtering window size | Effective denoising while preserving details under high noise density | Enlarged window causes blurring under high noise density | [12, 82] |
| 17 | Traditional preprocessing methods | Smoothing & denoising – spatial filtering – nonlinear (Anisotropic diffusion filter) | Controls diffusion based on image gradient; denoises flat regions while preserving edges | Removes noise while preserving edges and details; improves uneven illumination | Depends on parameter settings | [12] |
| 18 | Traditional preprocessing methods | Smoothing & denoising – frequency filtering (Gabor filter) | Uses Gaussian-modulated sinusoidal functions to extract directional texture features | Enhances vessel directional features | Depends on parameter settings | [154] |
| 19 | Traditional preprocessing methods | Smoothing & denoising – Stationary Wavelet Transform denoising | Uses translation-invariant multi-scale wavelet decomposition without downsampling and threshold reconstruction | Reduces translation artifacts and preserves structural continuity | Improper threshold selection may weaken fine vessels | [87] |
| 20 | Traditional preprocessing methods | Structure & edge enhancement (Single-scale Hessian eigenvalue analysis) | Computes Hessian eigenvalues to enhance ridge-like vessel structures | Improves vessel-background contrast without complex parameter tuning | Weak response to low-contrast vessels; cannot distinguish noise from vessels; affected by illumination and pathology | [90] |
| 21 | Traditional preprocessing methods | Structure & edge enhancement (Emboss) | Simulates embossed effect using grayscale gradient | Intuitively visualizes spatial structure | Produces pseudo-edges and reduces real details | [91] |
| 23 | Traditional preprocessing methods | Structure & edge enhancement (Laplacian of Gaussian, LoG) | Combines Gaussian smoothing with second-order derivative for edge detection | Improves edge localization accuracy | Sensitive to noise; prone to double edges | [69] |
| 24 | Traditional preprocessing methods | Structure & edge enhancement (Canny operator) | Uses gradient computation and non-maximum suppression for edge detection | Produces continuous and refined edges | Threshold-dependent; unstable for weak edges | [69, 260] |
| 25 | Traditional preprocessing methods | Combination methods (Morphological reflection suppression, Wiener filtering, contrast consistency enhancement, dual-threshold strategy) | Applies morphological reflection suppression, homomorphic filtering, and Wiener filtering to correct illumination and noise; enhances vessel continuity using contrast consistency and dual-threshold strategy | Addresses uneven illumination and low contrast; unsupervised adaptability; balances denoising and detail preservation; computationally efficient | Parameter-dependent; limited performance on severe pathology; weak for non-Gaussian noise; may cause local imbalance under extreme illumination | [261] |
| 26 | Traditional preprocessing methods | Combination methods (Gamma correction and CLAHE) | Combines gamma correction and CLAHE to balance illumination and enhance low-light details | Low computational cost; suitable for large-scale processing; avoids color distortion; improves low-light visibility | Fixed gamma limits adaptability; ineffective for extreme blur or noise; limited vessel/lesion enhancement | [93] |
| 27 | Traditional preprocessing methods | Combination methods (Illumination correction, Mahalanobis distance discrimination, and global spatial entropy-based contrast enhancement) | Based on scattering imaging model; integrates illumination correction, Mahalanobis discrimination, and contrast enhancement for joint optimization | Solves uneven illumination, low contrast, and color distortion; enhances fine vessels and lesions; strong generalization | Depends on threshold selection; may lose foreground in very low contrast; computationally complex | [262] |
| 28 | Traditional preprocessing methods | Combination methods (NLM filtering and CLAHE) | Uses NLM to suppress Gaussian noise while preserving microaneurysms, then applies CLAHE for contrast enhancement | Effective for microaneurysm detection; simple pipeline | Depends on channel selection; weak for multiplicative speckle noise; improper CLAHE parameters may cause over-enhancement | [97] |

Table 7: Comprehensive Summary of Preprocessing Methods for Fundus Color Images (continued)

| No. | Category | Method | Principle | Advantages | Limitations | References |
|---|---|---|---|---|---|---|
| 29 | Traditional preprocessing methods | Combination methods (ES-BM3D, CLAHE and coherence enhancement) | Applies ES-BM3D for denoising, CLAHE for contrast enhancement, and coherence enhancement for vessel continuity | Performs well under mixed noise; preserves vessel details and continuity | Sensitive to parameters; poor adaptability to severe pathology | [98] |
| 30 | Traditional preprocessing methods | Combination methods (Structure-adaptive filtering (SAF) and mathematical morphology (MM)) | Combines guided filtering and coherence-enhancing diffusion; enhances vessels via morphological closing and removes background noise | Effective for vessels of different scales; suitable for low-contrast images | Parameter-dependent; limited performance on pathological images | [263] |
| 31 | Traditional hybrid preprocessing methods | Combination methods (PPB, NLM, SVD, and GLM) | Combines PPB and NLM for denoising; uses SVD and GLM to enhance vessel segmentation | Strong mixed-noise suppression; preserves fine vessels; high segmentation accuracy | Long iterative process; depends on noise model; weak robustness under extreme illumination | [99] |
| 32 | Traditional hybrid preprocessing methods | Combination methods (Median filtering, matched filter/PSO+matched filter/matched filter-FDOG) | Uses median filtering for denoising; PSO optimizes matched filters (MF or MF-FDOG) for vessel enhancement | Automatic parameter optimization; improves detection of fine vessels | Poor performance for low-contrast, broken, or extremely thin vessels | [95] |
| 33 | Traditional hybrid preprocessing methods | Combination methods (Gamma correction, CLAHE, Gaussian/Median filtering) | Multi-stage pipeline for brightness and contrast optimization followed by smoothing | Significantly improves contrast and detail visibility | Limited enhancement for fine vessels; parameter-dependent; slight edge blurring; weak robustness for low-quality images | [94] |
| 34 | Traditional hybrid preprocessing methods | Combination methods (NAFSM/BM3D and HE/AHE/CLAHE/ESIHE) | Combines denoising (NAFSM/BM3D) with histogram-based contrast enhancement | Strong denoising and improved low-contrast visibility | Complex pipeline; fixed parameters; lacks adaptability | [96] |
| 35 | Traditional hybrid preprocessing methods | Combination methods (Gamma correction and HE in Lab space) | Performs gamma correction and HE in Lab color space | Preserves color consistency; enhances global and local contrast | CDF sensitive in severe pathology; gamma adaptability reduced; requires color space conversion; may amplify background noise in dark regions | [92] |
| 36 | Deep learning-assisted preprocessing methods | GAN-based (CycleGAN) | Utilizes two inverse generators to translate images between domains without paired data, while enforcing cycle-consistency loss to preserve content | Effectively preserves retinal anatomical structures and reduces interference from artifacts and domain discrepancies in downstream lesion detection tasks; enables flexible unsupervised learning | Limited by resolution; limited capability in handling severe artifacts; strong dependence on training data; high computational cost | [23, 24] |
| 37 | Deep learning-assisted preprocessing methods | GAN-based (StillGAN) | Treats low-quality and high-quality images as two independent domains and introduces local structural and illumination constraints to jointly learn global features and local details | Enables effective enhancement of multiple medical image types under unpaired settings while preserving structure and illumination | High computational cost; may introduce artifacts and color shifts; sensitive to parameter settings | [103] |
| 38 | Deep learning-assisted preprocessing methods | GAN-based (OTRE framework) | Employs optimal transport (OT) to preserve structure and regularized enhancement (RE) to optimize outputs, achieving unsupervised retinal image enhancement | Strong structural preservation; end-to-end unsupervised learning; good generalization and robustness | High computational cost; assumes all input images are usable and does not consider severely corrupted image screening | [104] |

Table 7: Comprehensive Summary of Preprocessing Methods for Fundus Color Images (continued)

| No. | Category | Method | Principle | Advantages | Limitations | References |
|---|---|---|---|---|---|---|
| 39 | Deep learning-assisted preprocessing methods | GAN-based (OTEGAN framework) | Based on optimal transport theory, maximizes information preservation and introduces lesion-consistent re-sampling strategy; adopts a U-shaped generator and efficient channel attention modules to improve image quality | Effectively avoids lesion distortion and false vessel generation; lightweight architecture with lower training and deployment cost | High computational cost; assumes all input images are usable and does not consider severely corrupted image screening | [105] |
| 40 | Deep learning-assisted preprocessing methods | GAN-based (P-GAN) | Starts from low resolution and progressively increases network depth with smooth layer transitions; incorporates normalization and minibatch discrimination to stabilize high-resolution image generation | Enables ultra-high magnification super-resolution reconstruction and clearly restores microaneurysm morphology | High computational cost; limited by resolution; limited capability in handling severe artifacts; strong dependence on training data | [106] |
| 41 | Deep learning-assisted preprocessing methods | GAN-based (ESRGAN) | Utilizes residual-in-residual dense blocks (RRDB) without batch normalization to extract features, combined with relativistic discriminator and improved perceptual loss for adversarial training-based super-resolution | Improves image quality discrimination and generates higher-quality images | High computational cost; may introduce structural distortions | [102] |
| 42 | Deep learning-assisted preprocessing methods | GAN-based (P-SRGAN) | Employs progressive training to generate high-resolution images from low resolution and introduces triplet loss to optimize image quality | Gradually reconstructs high-resolution images in super-resolution tasks and restores more details and critical features such as microaneurysms | High computational cost; limited capability in reconstructing low-quality images | [102] |
| 43 | Deep learning-assisted preprocessing methods | GAN-based (RankSRGAN) | Introduces ranking constraints based on perceptual metrics and a learned ranker to optimize the generator on top of SRGAN, producing more visually natural super-resolution images | Improves visual quality and perceptual optimization | High computational cost; dependent on data quality; may introduce artifacts | [102] |
| 44 | Deep learning-assisted preprocessing methods | GAN-based (DUS-GAN) | Enhances low-resolution images using GAN without supervised training and introduces mean opinion score (MOS) loss to improve perceptual quality | Unsupervised training; adaptable to various degradations; improves visual quality | High computational cost; long training time; may introduce artifacts; dependent on data quality | [102] |
| 45 | Deep learning-assisted preprocessing methods | GAN-based (IGAN) | Constructs a deep network using Inception-style parallel convolutions and dilated convolutions; incorporates spectral normalization to stabilize training and mitigate mode collapse while improving fidelity | Multi-scale feature extraction enhances detail representation; alleviates mode collapse and improves generation performance | High computational cost; complex model architecture; limited performance in complex scenarios | [102] |
| 46 | Deep learning-assisted preprocessing methods | GAN-based (Real-ESRGAN) | Builds a high-order degradation model to simulate real-world distortions; adopts RRDB generator and spectral-normalized U-Net discriminator; trained on purely synthetic data for real-world blind super-resolution | Strong generalization in real scenarios; no reliance on paired data; excellent visual quality | High computational cost; limited robustness under severe degradation; risk of over-sharpening details | [102] |
| 47 | Deep learning-assisted preprocessing methods | Non-GAN (A decomposition-based network) | Based on Retinex theory, decomposes images into illumination and reflectance components to correct non-uniform illumination while preserving details | Effectively corrects uneven illumination while preserving details and color information; improves vessel segmentation accuracy | High computational cost; dependent on supervision quality; limited to illumination decomposition and does not address other degradations | [101] |

Table 7: Comprehensive Summary of Preprocessing Methods for Fundus Color Images (continued)

| No. | Category | Method | Principle | Advantages | Limitations | References |
|---|---|---|---|---|---|---|
| 48 | Deep learning-assisted preprocessing methods | Non-GAN (DTRSRN) | Integrates Swin Transformer for global context modeling and residual CNN for detail extraction; employs fractal dimension as quantitative supervision to preserve vascular morphology for high-precision super-resolution | Effectively preserves fine vascular morphology | Complex model architecture; large number of parameters; high computational cost | [109] |
| 49 | Deep learning-assisted preprocessing methods | Non-GAN (COFE-Net) | Simulates real degradation processes in fundus images and constructs a clinically oriented enhancement network | Suppresses uneven illumination and blur while preserving anatomical structures and pathological features | Strong dependence on training data; limited capability in handling severely degraded images; high computational cost | [108] |
| 50 | Deep learning-assisted preprocessing methods | Non-GAN (A deep learning-based framework for retinal fundus image enhancement) | Constructs datasets using real paired high- and low-quality fundus images; simulates degradation via data augmentation and applies customized CNN for automatic enhancement | Automated processing; adaptable to multiple degradations | Strong dependence on training data; limited capability in handling severely degraded images; high computational cost | [107] |
| 51 | Deep learning-assisted preprocessing methods | Non-GAN (ArcNet) | Requires no labeled data; extracts high-frequency components to preserve retinal structure; trained on synthetic data and adapted to real cataract fundus images | Eliminates dependence on annotations; effectively restores clarity in cataract fundus images | Dependent on synthetic data and high-frequency priors; limited performance for severely opaque images; high computational cost | [110] |
| 52 | Deep learning-assisted preprocessing methods | Non-GAN (Context-Aware OT) | Models fundus image enhancement as an unpaired distribution alignment problem; leverages deep contextual features and optimal transport distance to map low-quality images to high-quality ones | Better preserves local structures; reduces artifacts; supports unpaired training | Limited by resolution; limited capability in handling severe artifacts; strong dependence on training data; high computational cost | [111] |
| 53 | Hybrid preprocessing methods | Sequential (CLAHE and ESRGAN-based preprocessing) | Applies CLAHE to enhance image contrast, followed by ESRGAN for super-resolution reconstruction | Enhances details and contrast, effectively improving model learning capability | High computational cost; slow inference speed; may introduce hallucination artifacts | [112] |
| 54 | Hybrid preprocessing methods | Sequential (CLAHE and GAN-based preprocessing) | CLAHE enhances local contrast, while GAN further restores details | Improves visibility of DR-related features and enhances classification performance | May introduce over-enhancement and artifacts; high computational cost | [113] |
| 55 | Hybrid preprocessing methods | Embedded (PCFlow) | Embeds traditional high-frequency filtering operators into a deep learning network, extracts retinal structures as constraints, and employs invertible pyramid normalizing flows for end-to-end frequency-domain hierarchical enhancement | Combines advantages of traditional operators and deep learning methods | Limited capability in handling extremely low-quality images; strong dependence on high-quality annotated datasets | [114] |
| 56 | Hybrid preprocessing methods | Embedded (GFENet) | Uses Gaussian high-pass filtering to extract high-frequency maps (HFM), which are embedded as frequency-domain self-supervised signals into a deep enhancement network | Preserves vascular structures; does not require paired data; strong generalization capability | High computational cost; limited capability in handling severely degraded images | [115] |

## F. Performance Comparison of Preprocessing Techniques

Table 8: Performance Comparison of Preprocessing Techniques

| Category | Method | Dataset | PSNR (dB) | AMBE | SNR | Entropy | MSE | SSIM | Sen (%) | Spec (%) | Acc (%) | Ref. |
|---|---|---|---|---|---|---|---|---|---|---|---|---|
| Traditional preprocessing methods | HE | Not reported | 7.0327 | – | −2.1674 | 4.1383 | – | – | – | – | – | [68] |
| Traditional preprocessing methods | ADHE | Not reported | 22.1576 | – | −0.1772 | 5.1255 | – | – | – | – | – | [68] |
| Traditional preprocessing methods | CLAHE | DRIVE | 29.6413 | – | – | 4.8494 | 255.02 | 0.86 | – | – | – | [21] |
| Traditional preprocessing methods | Rayleigh CLAHE | DIARETDB | – | – | – | 5.9846–6.1790 | – | – | – | – | – | [75] |
| Traditional preprocessing methods | ESIHE | Not reported | 15.5006 | – | −0.1570 | 4.5129 | – | – | – | – | – | [68] |
| Traditional preprocessing methods | Combination methods (BM3D and ESIHE) | DIARETDB1 | 16.6203 | 26.9654 | 4.0196 | 6.4381 | – | – | – | – | – | [96] |
| Traditional preprocessing methods | Combination methods (Median filtering and matched filter-FDOG) | DRIVE | 58.33 | – | – | – | 0.093 | – | 16.639 | 96.85 | 89.233 | [95] |
| Traditional preprocessing methods | Combination methods (Median filtering, matched filter, and PSO) | DRIVE | 59.35 | – | – | – | 0.071 | – | 34.516 | 97.161 | 91.105 | [95] |
| Traditional preprocessing methods | Combination methods (Median filtering and matched filter) | DRIVE | 58.17 | – | – | – | 0.095 | – | 17.54 | 95.976 | 88.363 | [95] |
| Deep learning-based preprocessing | Retinex decomposition network | Kaggle Diabetic Retinopathy Detection Challenge | 25.05 | – | – | – | – | 0.9119 | – | – | – | [101] |
| Deep learning-based preprocessing | P-SRGAN | EYEPACS | 46.1 | – | – | – | – | 0.91 | 0.89 | – | 0.96 | [106] |

*Abbreviations:* PSNR, peak signal-to-noise ratio; AMBE, absolute mean brightness error; SNR, signal-to-noise ratio; MSE, mean squared error; SSIM, structural similarity index; Sen, sensitivity; Spec, specificity; Acc, accuracy.

## G. Explicit Structural Modeling Methods with Mathematical Priors

Table 9: Explicit Structural Modeling Methods with Mathematical Priors

| Method | Task | Target | Preprocessing | Dataset | Performance |
|---|---|---|---|---|---|
| 2D Gaussian Matched Filtering [121] | Detection | Retinal vessels | Mean filtering for noise suppression; field-of-view masking; automatic threshold segmentation using between-class variance maximization | DRIVE | Acc = 0.877, AUC = 0.788 |
| Matched Gaussian Filtering with Kalman Tracking [264] | Detection | Retinal vessels | Gaussian smoothing combined with second-order derivative-based response enhancement | NR | NR |
| Piecewise Threshold Probing of Matched Filter Response [14] | Segmentation | Retinal vessels | Response normalization; global threshold segmentation; morphological thinning and endpoint initialization | In-house dataset | At TPR ≈ 0.750, FPR reduced by 15×; standard Acc/AUC not reported |
| Mathematical Morphology with Curvature Evaluation [122] | Segmentation | Retinal vessels | Noise suppression; connectivity-based morphological reconstruction; intensity normalization | STARE | Sen = 0.675, Spe = 0.957, AUC = 0.904 |
| Ridge-Based Vessel Segmentation Using Profile-Based Measurement [28] | Segmentation | Retinal vessels | Green-channel intensity extraction; local mean-variance normalization; multi-scale Gaussian derivative computation | Utrecht; Hoover | Utrecht: Acc = 0.944, AUC = 0.952; Hoover: Acc = 0.952, AUC = 0.961 |
| Likelihood Ratio Vesselness Filtering [147] | Detection | Retinal vessels | Multi-scale Gaussian derivative analysis; orientation and scale selection; non-maximum suppression | DRIVE | Acc = 0.880 |
| Centerline Detection with Morphological Reconstruction [145] | Segmentation | Retinal vessels | Illumination normalization using large-kernel mean filtering; directional linear filtering for vessel enhancement | DRIVE; STARE | DRIVE: Acc = 0.946; STARE: Acc = 0.952 |
| Multiscale Differential Geometry with Region Growing [265] | Segmentation | Retinal vessels | Green-channel contrast enhancement; multi-scale Gaussian convolution for differential structure analysis; adaptive threshold determination | STARE | Acc = 0.960 ± 0.010, Sen (TPR) = 0.720 ± 0.070, FPR = 0.020 ± 0.010 |
| Ribbon of Twins Active Contour Model [123] | Segmentation | Retinal vessels | Morphological centerline initialization; removal of spurious branches and junction artifacts | DRIVE; STARE; REVIEW | DRIVE: Sen = 0.732; HRIS: vessel width error = 0.420 pixels |
| Fourier Cross-Sectional Profile-Based Vessel Detection [266] | Detection | Retinal vessels | Green-channel extraction for vessel-to-background contrast enhancement | STARE; ARIA | ROC analysis reported; quantitative metrics unavailable |
| Morphological Bit-Plane-Based Vessel Segmentation [150] | Segmentation | Retinal vessels | Region-of-interest masking; background normalization via mean filtering; vessel skeleton initialization | DRIVE; STARE | Average: Acc = 0.942; DRIVE: Sen = 0.730, Spe = 0.974; STARE: Sen = 0.732, Spe = 0.966 |
| Finite Element Binary Level Set Method [267] | Segmentation | Retinal vessels | Gaussian matched filtering for vessel enhancement; morphological opening for removal of nonvascular structures | DRIVE; STARE | DRIVE: Acc = 0.951; STARE: Acc = 0.952, Sen = 0.753, Spe = 0.981; Kappa = 0.740 |
| Graph-Cut Segmentation with Vector Flux Constraint [124] | Segmentation | Retinal vessels and optic disc | Intensity inversion; adaptive histogram equalization; morphological opening; distance-transform-based vector field initialization | DRIVE; STARE | DRIVE: Sen = 0.751, Acc ≈ 0.947; STARE: Sen = 0.789, Acc = 0.944, FPR = 0.037 |
| Trainable Combination of Shifted Filter Responses [149] | Segmentation | Retinal vessels | Green-channel extraction; region-of-interest masking; contrast enhancement using adaptive histogram equalization | DRIVE; STARE; CHASE_DB1 | DRIVE: Acc = 0.944, Sen = 0.766, Spe = 0.970, AUC = 0.961; STARE: Acc = 0.950, Sen = 0.772, Spe = 0.970, AUC = 0.956; CHASE_DB1: Acc = 0.939, Sen = 0.759, Spe = 0.959, AUC = 0.949 |
| Cognitive Hierarchical Graph-Based Segmentation [268] | Segmentation | Retinal vessels | Illumination correction using Gaussian low-pass filtering; multi-scale black top-hat enhancement | DRIVE; STARE | DRIVE: Acc = 0.934, Sen = 0.850, Spe = 0.944; STARE: Acc = 0.924, Sen = 0.633, Spe = 0.950 |
| Improved Multiscale Matched Filtering [269] | Segmentation | Retinal vessels | Illumination correction using spline-based modeling; contrast equalization; green-channel extraction | DRIVE; STARE; HRF | DRIVE: Acc = 0.934, Sen = 0.706, Spe = 0.969, AUC = 0.952; STARE: Acc = 0.934, Sen = 0.785, Spe = 0.951, AUC = 0.957; HRF-Healthy: Acc = 0.954, AUC = 0.974 |

Table 9: Explicit Structural Modeling Methods with Mathematical Priors (continued)

| Method | Task | Target | Preprocessing | Dataset | Performance |
|---|---|---|---|---|---|
| Adaptive Morphological Optic Disc Segmentation [148] | Segmentation | Optic disc | Bright-lesion suppression using morphological reconstruction; top-hat contrast enhancement; morphological opening and closing operations | DRIVE; DIARETDB1 | DRIVE: localization success rate = 1.000, Overlap = 0.415; DIARETDB1: success rate = 0.978, Overlap = 0.437, mean absolute distance = 8.310 ± 4.050 pixels |

*Abbreviations:* Acc, accuracy; AUC, area under the receiver operating characteristic curve; Sen, sensitivity; Spe, specificity; FPR, false positive rate; TPR, true positive rate; NR, not reported.
*Notes:* Results are reported in a unified format for consistency across studies, with only representative metrics retained for comparison. Some performance metrics (e.g., [121], [122], [147]) were not reported in the original papers and were extracted from the review study [143].

# H. Implicit Pattern Learning Methods with Handcrafted Features

Table 10: Implicit Pattern Learning Methods with Handcrafted Features

| Method | Task | Target | Preprocessing | Dataset | Performance |
|---|---|---|---|---|---|
| Wavelet Transform with k-Nearest Neighbor Classification [125] | Classification | Diabetic retinopathy | Median filtering; adaptive histogram equalization; green-channel extraction; optic disc removal | DIARETDB1 | Acc = 0.950, Sen = 0.926, Spe = 0.876, Precision = 0.961 |
| Machine Learning Ensemble with Majority Voting [270] | Classification | Age-related macular degeneration | Adaptive contrast enhancement; histogram equalization; resolution interpolation; region-of-interest extraction; contour detection; image scaling | Private dataset (n = 864) | AUC = 0.959, Acc = 0.969, F1 = 0.937, Sen = 0.937, Spe = 0.979 |
| Multilevel Segmentation with GLCM and Entropy Features [271] | Classification | Age-related macular degeneration | Image resizing to 256 × 256; brightness adjustment; contrast enhancement; Gaussian filtering | REFUGE2018 | Acc = 0.937, F1 = 0.937, Sen = 0.933, Spe = 0.940 |
| Logistic Regression Classification [272] | Classification | Glaucoma | Image resizing to 800 pixels; optic disc cropping to 240 × 240 | Internal dataset (n = 464) | Acc (overall) = 0.772, Acc (advanced) = 0.986, Acc (early) = 0.730 |
| Pixel-Wise k-Nearest Neighbor Classification [273] | Segmentation | Retinal vessels | Green-channel extraction; feature normalization | DRIVE | AUC = 0.929, Acc = 0.942, Kappa = 0.715 |
| Ridge-Based Vessel Segmentation Using Profile-Based Measurement [152] | Segmentation | Retinal vessels | Green-channel extraction; local intensity normalization | Utrecht; Hoover | Utrecht: AUC = 0.952, Acc = 0.944; Hoover: AUC = 0.961, Acc = 0.952 |
| Morlet Wavelet and Gaussian Mixture Model Vessel Segmentation [154] | Segmentation | Retinal vessels | Green-channel inversion; aperture correction; intensity normalization | DRIVE; STARE | DRIVE: AUC = 0.960, Acc = 0.947; STARE: AUC = 0.965, Acc = 0.947 |
| Line Operator with Linear SVM Classification [126] | Segmentation | Retinal vessels | Inverted green-channel extraction; field-of-view correction | DRIVE; STARE | DRIVE: AUC = 0.963, Acc = 0.960; STARE: AUC = 0.968, Acc = 0.965 |
| Scale-Space Vessel Segmentation [274] | Segmentation | Retinal vessels | Adaptive histogram equalization; field-of-view masking | DRIVE | Spe = 0.948, Kappa = 0.807 |
| Multiscale Gabor Filtering with GMM Segmentation [127] | Segmentation | Retinal vessels | Green-channel extraction; contrast enhancement | DRIVE | AUC = 0.974, Sen = 0.965, Spe = 0.971 |
| Wavelet, Curvelet, and Hessian Features with SVM [275] | Segmentation | Retinal vessels | Green-channel extraction; intensity normalization; median filtering; adaptive thresholding | DRIVE | Acc = 0.933, Sen = 0.776 |
| Fuzzy Artificial Bee Colony Optimization Segmentation [276] | Segmentation | Retinal vessels | Field-of-view masking | DRIVE | AUC = 0.956, Acc = 0.960 |
| Radial Projection with Semi-Supervised SVM Segmentation [277] | Segmentation | Retinal vessels | Green-channel extraction; contrast enhancement | DRIVE; STARE | DRIVE: Acc = 0.943, Sen = 0.741, Spe = 0.975; STARE: Acc = 0.950, Sen = 0.729, Spe = 0.976 |
| Leung–Malik Filter Bank with Hierarchical Learning [278] | Segmentation | Retinal vessels | Green-channel extraction | DRIVE; STARE | DRIVE: AUC = 0.955, Acc = 0.903; STARE: AUC = 0.971, Acc = 0.927 |
| Major Vessel Extraction with Subimage Classification [152] | Segmentation | Retinal vessels | Green-channel extraction; high-pass filtering; top-hat reconstruction | DRIVE; STARE; CHASE_DB1 | DRIVE: AUC = 0.962, Acc = 0.952, Sen = 0.725, Spe = 0.983; STARE: AUC = 0.969, Acc = 0.951, Sen = 0.772, Spe = 0.973 |
| Fully Connected Conditional Random Field Segmentation [128] | Segmentation | Retinal vessels | NR | DRIVE; STARE; CHASE_DB1; HRF | DRIVE: Sen = 0.790, Spe = 0.968; STARE: Sen = 0.768, Spe = 0.974; CHASE_DB1: Sen = 0.728, Spe = 0.951; HRF: Sen = 0.787, Spe = 0.958 |
| Ensemble Classification-Based Vessel Segmentation [129] | Segmentation | Retinal vessels | Green-channel extraction; contrast normalization | DRIVE; STARE | DRIVE: AUC = 0.975, Acc = 0.948, Sen = 0.741, Spe = 0.981; STARE: AUC = 0.977, Acc = 0.953, Sen = 0.755, Spe = 0.976 |

Table 10: Implicit Pattern Learning Methods with Handcrafted Features (continued)

| Method | Task | Target | Preprocessing | Dataset | Performance |
|---|---|---|---|---|---|
| Hierarchical Feature Extraction with Ensemble Learning [279] | Segmentation | Retinal vessels | Illumination correction; vessel contrast enhancement | DRIVE; STARE | DRIVE: Acc = 0.977, Sen = 0.817, Spe = 0.973, AUC = 0.948; STARE: Acc = 0.981, Sen = 0.810, Spe = 0.979, AUC = 0.975 |
| Automated Retinal Image Quality Assessment [280] | Classification | Fundus image quality | Green-channel extraction; intensity normalization | UK Biobank | AUC = 0.903, Sen = 0.953, Spe = 0.913 |
| Ensemble Retinal Vessel Segmentation [281] | Segmentation | Retinal vessels | Green-channel extraction | DRIVE; RIS | DRIVE: Acc = 0.962, Sen = 0.746, Spe = 0.984; RIS: Acc = 0.954, Sen = 0.832, Spe = 0.961 |

*Abbreviations:* Acc, accuracy; AUC, area under the receiver operating characteristic curve; Sen, sensitivity; Spe, specificity; Precision, positive predictive value; F1, F1-score; IoU, intersection over union; DSC, Dice similarity coefficient; MCC, Matthews correlation coefficient; Kappa, Cohen's kappa coefficient; NR, not reported.
*Notes:* Results are reported in a unified format for consistency across studies; only representative metrics are retained for comparison.

# I. Deep Learning Methods with Convolutional Neural Networks

Table 11: Deep Learning Methods with Convolutional Neural Networks

| Method | Task | Target | Preprocessing | Dataset | Performance |
|---|---|---|---|---|---|
| MPCNN [169] | Segmentation | Retinal vessels | No explicit preprocessing; green-channel extraction | DRIVE | Acc = 0.947, AUC = 0.975, TPR = 0.728, FPR = 0.022 |
| Asymmetric deep learning network [163] | Segmentation | Optic disc | Background removal; field-of-view cropping and resizing to 256 × 256; image normalization; data augmentation using random flipping, translation, and rotation | MESSIDOR; ORIGA; REFUGE | Global: DSC = 0.938, IoU = 0.885, MCC = 0.938; Local disc region: DSC = 0.974, IoU = 0.949, MCC = 0.959 |
| W-Net [130] | Segmentation | Optic disc and optic cup | Region-of-interest cropping; contrast enhancement using CLAHE; Gaussian filtering; polar coordinate transformation | REFUGE; DRISHTI-GS; RIM-ONE r3; Private dataset | REFUGE: OD (OE = 0.068, F1 = 0.964, Spe = 0.986, Sen = 0.974, Acc = 0.982), OC (OE = 0.178, F1 = 0.900, Spe = 0.995, Sen = 0.905, Acc = 0.987); DRISHTI-GS: OD (OE = 0.054, F1 = 0.971, Spe = 0.994, Sen = 0.958, Acc = 0.981), OC (OE = 0.233, F1 = 0.858, Spe = 0.983, Sen = 0.925, Acc = 0.972); RIM-ONE r3: OD (OE = 0.049, F1 = 0.975, Spe = 0.986, Sen = 0.981, Acc = 0.984), OC (OE = 0.237, F1 = 0.863, Spe = 0.986, Sen = 0.895, Acc = 0.972); Private dataset: OD (OE = 0.051, F1 = 0.974, Spe = 0.990, Sen = 0.967, Acc = 0.983), OC (OE = 0.255, F1 = 0.843, Spe = 0.989, Sen = 0.879, Acc = 0.976) |
| Coarse-to-fine deep learning framework [131] | Segmentation | Optic disc | Removal of non-retinal background; field-of-view cropping; image normalization; data augmentation using image flipping and rotation | CFI; DIARETDB0; DIARETDB1; DRIONS-DB; DRIVE; MESSIDOR; ORIGA | Overall (n = 2978): IoU = 0.891, DSC = 0.939, Acc = 0.970, Sen = 0.944 |
| SU-Net [170] | Segmentation | Optic disc and optic cup | Coarse segmentation; contour coordinate transformation; cropping of contour neighborhood regions; geometric parameter perturbation for data augmentation; random cropping and scaling | MESSIDOR; DRISHTI-GS | MESSIDOR: Disc (E = 0.040, D = 0.979), Cup (E = 0.145, D = 0.920); DRISHTI-GS: Disc (E = 0.047, D = 0.976), Cup (E = 0.163, D = 0.907), CDR error = 0.047, AUC = 0.935 |
| MGAN [282] | Segmentation | Arteries and veins | Image resizing to 512 × 512; random patch extraction of size 256 × 256; data augmentation using flipping, rotation, and translation | LEI-CENTRAL; AV-DRIVE | LEI-CENTRAL: BACC = 0.932, Sen = 0.922, Spe = 0.941, DiceV = 0.666, DiceA = 0.624; AV-DRIVE: BACC = 0.955, Sen = 0.936, Spe = 0.973, DiceV = 0.794, DiceA = 0.755 |
| Bayesian U-Net [132] | Segmentation | Optic disc | Pseudo-label generation using Hough transform; Gaussian filtering; morphological opening; Canny edge detection; morphological dilation; zero padding and resizing to 640 × 640 | DRISHTI-GS; RIM-ONE; REFUGE validation set; REFUGE test set | DRISHTI-GS: Acc = 0.997, Sen = 0.983, Spe = 0.997, IoU = 0.895, DSC = 0.944, HD = 3.615, ASD = 3.394; RIM-ONE: Acc = 0.992, Sen = 0.974, Spe = 0.992, IoU = 0.783, DSC = 0.876, HD = 4.993, ASD = 8.886; REFUGE validation set: Acc = 0.997, Sen = 0.965, Spe = 0.997, IoU = 0.829, DSC = 0.903, HD = 4.043, ASD = 7.343; REFUGE test set: Acc = 0.997, Sen = 0.973, Spe = 0.997, IoU = 0.819, DSC = 0.896, HD = 4.049, ASD = 6.954 |
| PY-Net [162] | Segmentation | Optic disc and optic cup | NR | REFUGE; DRISHTI-GS; RIM-ONE r3 | RIM-ONE r3: OD (DSC = 0.961, IoU = 0.926, Sen = 0.958, Acc = 0.968), OC (DSC = 0.929, IoU = 0.874, Sen = 0.932, Acc = 0.990); DRISHTI-GS: OD (DSC = 0.971, IoU = 0.944, Sen = 0.969, Acc = 0.998), OC (DSC = 0.934, IoU = 0.876, Sen = 0.946, Acc = 0.998); REFUGE: OD (IoU = 0.932, DSC = 0.965, Acc = 0.999, Sen = 0.975), OC (IoU = 0.885, DSC = 0.939, Acc = 1.000, Sen = 0.937) |
| Joint segmentation model [283] | Segmentation | Arteries and veins | Illumination correction using median filtering; construction of six-channel input representation | DRIVE; MESSIDOR; CT-DRIVE | A/V classification: Sen = 0.937, Spe = 0.929; Vessel segmentation: Acc ≈ 0.948 |

Table 11: Deep Learning Methods with Convolutional Neural Networks (continued)

| Method | Task | Target | Preprocessing | Dataset | Performance |
|---|---|---|---|---|---|
| BCE3-U-Net [284] | Segmentation | Arteries and veins | Global contrast enhancement; local intensity normalization; Gaussian low-pass filtering | RITE test set; DRIVE test set | RITE test set: arteries (AUC-ROC = 0.974 ± 0.002, AUC-PR = 0.813 ± 0.007), veins (AUC-ROC = 0.980 ± 0.001, AUC-PR = 0.870 ± 0.004), vessels (AUC-ROC = 0.983 ± 0.000, AUC-PR = 0.928 ± 0.001), A/V classification (Sen = 0.875 ± 0.021, Spe = 0.909 ± 0.007, Acc = 0.892 ± 0.007, AUC-ROC = 0.956 ± 0.004), vessel/background (Sen = 0.791 ± 0.012, Spe = 0.987 ± 0.001, Acc = 0.962 ± 0.001, AUC-ROC = 0.983 ± 0.000); DRIVE test set: AUC-ROC = 0.978 ± 0.000, AUC-PR = 0.910 ± 0.001 |
| MCU-Net [285] | Segmentation | Retinal vessels | Intensity normalization; median filtering; contrast enhancement; multi-scale feature representation construction | Duke; Private dataset; DRIVE | Duke: F1 = 0.799, Sen = 0.801, Spe = 0.986, Acc = 0.975, AUC = 0.987; Private dataset: F1 = 0.753, Sen = 0.749, Spe = 0.988, Acc = 0.976, AUC = 0.986; DRIVE: F1 = 0.819, Sen = 0.802, Spe = 0.985, Acc = 0.969, AUC = 0.983 |
| NUA-Net [160] | Segmentation | Retinal vessels | Color space transformation from RGB to Lab; CLAHE applied to luminance channel; conversion back to RGB; green-channel extraction; normalization to [0,1]; random patch cropping; data augmentation using flipping and rotation | DRIVE; STARE; CHASE_DB1; HRF; IOSTAR | DRIVE: Sen = 0.806, Spe = 0.986, Acc = 0.971, AUC = 0.988; STARE: Sen = 0.851, Spe = 0.990, Acc = 0.979, AUC = 0.994; CHASE_DB1: Sen = 0.811, Spe = 0.986, Acc = 0.975, AUC = 0.988; HRF: Sen = 0.855, Spe = 0.974, Acc = 0.965, AUC = 0.982; IOSTAR: Sen = 0.839, Spe = 0.975, Acc = 0.965, AUC = 0.982; PR-AUC: DRIVE = 0.916, STARE = 0.947 |
| BTS-DSN [174] | Segmentation | Retinal vessels | Data augmentation using rotation, flipping, and scaling; patch-based training by dividing images into multiple regions; dataset-specific resolution adjustment | DRIVE; STARE; CHASE_DB1 | DRIVE: Sen = 0.789, Spe = 0.980, Acc = 0.956, AUC = 0.981, MCC = 0.796, F1 = 0.825; STARE: Sen = 0.821, Spe = 0.984, Acc = 0.967, AUC = 0.986, MCC = 0.822, F1 = 0.842; CHASE_DB1: Sen = 0.789, Spe = 0.980, Acc = 0.963, AUC = 0.984, MCC = 0.773, F1 = 0.798 |
| SSANet [159] | Segmentation | Retinal vessels | Zero-mean unit-variance normalization; data augmentation using random rotation, brightness adjustment, contrast scaling, and color augmentation | DRIVE; STARE; CHASE_DB1; HRF | DRIVE: Acc = 0.957, Spe = 0.975, Sen = 0.835, AUC = 0.982; STARE: Acc = 0.976, Spe = 0.986, Sen = 0.854, AUC = 0.992; CHASE_DB1: Acc = 0.978, Spe = 0.987, Sen = 0.852, AUC = 0.992; HRF: AUC = 0.910, F1 = 0.822 |
| SID2Net [161] | Segmentation | Retinal vessels | Green-channel extraction; mean intensity subtraction; global threshold post-processing | DRIVE; STARE | DRIVE: Acc = 0.952, Sen = 0.843, Spe = 0.968, F1 = 0.816, MCC = 0.791, G = 0.903, FS = 0.834, AUC = 0.975; STARE: Acc = 0.962, Sen = 0.863, Spe = 0.973, F1 = 0.823, MCC = 0.804, G = 0.916, FS = 0.845, AUC = 0.982 |
| Four-class U-Net [286] | Classification | Arteries and veins | Region-of-interest padding; local contrast enhancement; background suppression using Gaussian smoothing; dataset mean-standard deviation normalization; data augmentation using cropping, rotation, flipping, and elastic deformation | DRIVE; HRF | DRIVE: vessel segmentation Acc = 0.968; A/V discrimination zone 3 (Acc = 0.943, F1 = 0.943), zone 1 (Acc = 0.939, F1 = 0.938), zone 4 (Acc = 0.944, F1 = 0.944); HRF: Det% vessel = 0.807, full-image Acc = 0.968, full-image F1 = 0.969 |
| SEGAN [133] | Segmentation | Retinal vessels | Intensity normalization; dataset-specific resolution adjustment using downsampling; spatial padding; data augmentation using rotation and flipping | DRIVE; STARE; CHASE_DB1; HRF | DRIVE: Sen = 0.829, Spe = 0.981, Precision = 0.840, Acc = 0.956, AUC = 0.983, G = 0.902, F1 = 0.835; STARE: Sen = 0.881, Spe = 0.978, Precision = 0.795, Acc = 0.967, AUC = 0.986, G = 0.928, F1 = 0.836; CHASE_DB1: Sen = 0.844, Spe = 0.978, Precision = 0.801, Acc = 0.963, AUC = 0.987, G = 0.908, F1 = 0.822; HRF: Sen = 0.831, Spe = 0.973, Precision = 0.812, Acc = 0.956, AUC = 0.969, G = 0.899, F1 = 0.821 |
| DUNet [287] | Segmentation | Retinal vessels | Conversion to single-channel representation; intensity normalization; CLAHE; gamma correction; random patch cropping | DRIVE; STARE; CHASE_DB1; HRF; SYNTHE | DRIVE: PPV = 0.853, TPR = 0.796, TNR = 0.980, Acc = 0.957, F1 = 0.824, AUC = 0.980; STARE: PPV = 0.878, TPR = 0.760, TNR = 0.988, Acc = 0.964, F1 = 0.814, AUC = 0.983; CHASE_DB1: PPV = 0.763, TPR = 0.816, TNR = 0.975, Acc = 0.961, F1 = 0.788, AUC = 0.980; HRF: PPV = 0.859, TPR = 0.746, TNR = 0.987, Acc = 0.965, AUC = 0.983; SYNTHE: PPV = 0.851, TPR = 0.649, TNR = 0.982, Acc = 0.935, F1 = 0.736, AUC = 0.940 |

Table 11: Deep Learning Methods with Convolutional Neural Networks (continued)

| Method | Task | Target | Preprocessing | Dataset | Performance |
|---|---|---|---|---|---|
| Strided-CNN [288] | Segmentation | Retinal vessels | Illumination correction using morphological closing; channel-wise background correction; PCA-based grayscale conversion; contrast enhancement; noise suppression | DRIVE; STARE; CHASE_DB1; HRF | DRIVE: Sen = 0.870, Spe = 0.985, Acc = 0.956, AUC = 0.986, DSC = 0.985, IoU = 0.948; STARE: Sen = 0.848, Spe = 0.986, Acc = 0.968, AUC = 0.988, DSC = 0.981, IoU = 0.946; HRF: Sen = 0.829, Spe = 0.961, Acc = 0.962, AUC = 0.985, DSC = 0.979, IoU = 0.939; CHASE_DB1: Sen = 0.886, Spe = 0.982, Acc = 0.976, AUC = 0.985, DSC = 0.981, IoU = 0.939 |
| DA-Res2UNet [134] | Segmentation | Retinal vessels | Removal of non-retinal borders; gamma correction; grayscale conversion; CLAHE; Z-score normalization | DRIVE; CHASE_DB1; STARE | DRIVE: Acc = 0.970, Spe = 0.985, Sen = 0.820, AUC = 0.987, MCC = 0.812, F1 = 0.827; CHASE_DB1: Acc = 0.977, Spe = 0.989, Sen = 0.808, AUC = 0.991, MCC = 0.804, F1 = 0.816; STARE: Acc = 0.976, Spe = 0.989, Sen = 0.814, AUC = 0.988, MCC = 0.823, F1 = 0.835 |

*Abbreviations:* Acc, accuracy; AUC, area under the receiver operating characteristic curve; AUC-PR, area under the precision-recall curve; Sen, sensitivity; Spe, specificity; Precision, positive predictive value when reported as precision; DSC, Dice similarity coefficient; IoU, intersection over union; MCC, Matthews correlation coefficient; HD, Hausdorff distance; ASD, average surface distance; BACC, balanced accuracy; PPV, positive predictive value; TPR, true positive rate; TNR, true negative rate; PR-AUC, area under the precision-recall curve; NR, not reported.
*Notes:* Results are reported in a unified format for consistency across studies; only formatting and notation have been standardized, without altering the original reported content.

# J. Representation Learning Methods with Visual Unimodal Data

Table 12: Representation Learning Methods with Visual Unimodal Data

| Method | Task | Target | Preprocessing | Dataset | Performance |
|---|---|---|---|---|---|
| ViT-based DR grading model [135] | Classification | DR | Histogram equalization; image translation augmentation; random rotation | Kaggle Diabetic Retinopathy Detection | Acc = 0.914, Sen = 0.926, Spe = 0.977, AUC = 0.988, F1 = 0.923, Precision = 0.928, QWK = 0.935 |
| Vision Transformer with Residual Attention [184] | Classification | DR | Random resizing to 512 × 512; random horizontal flipping; random vertical flipping; random rotation | DDR; IDRiD | DDR: Acc (Class 0/1/2/3/4/5) = 0.828/0.964/0.779/0.988/0.958/0.975, AUC = 0.902, AA = 0.915, AUC (Class 0/1/2/3/4/5) = 0.998/0.613/0.951/0.946/0.974/0.929; IDRiD: Acc (Class 0/1/2/3/4) = 0.806/0.981/0.728/0.913/0.971 |
| VMLRI [187] | Classification | DR | Image cropping; uniform resizing; random horizontal flipping; random rotation; consistent augmentation for pretraining and fine-tuning | APTOS | Acc = 0.934, Sen = 0.973, Spe = 0.954, AUC = 0.985 |
| ViT-AMD [289] | Classification | AMD | Image resizing to 224 × 224 | Retinal Image Bank; iChallenge-AMD | Acc = 0.742; Normal: Sen = 0.824, Spe = 0.880, F1 = 0.799, Precision = 0.774; Dry AMD: Sen = 0.760, Spe = 0.794, F1 = 0.700, Precision = 0.649; Wet AMD: Sen = 0.640, Spe = 0.938, F1 = 0.726, Precision = 0.838 |
| STMF-DRNet [290] | Classification | DR | Removal of black background; image resizing to 512 × 512; CLAHE; Gaussian smoothing; class-specific augmentation using flipping and scaling | DDR; EyePACS; APTOS-2019; Clinical dataset | DDR: Acc = 0.879, Sen = 0.816, Spe = 0.849, F1 = 0.891, Kappa = 0.890; EyePACS: Acc = 0.863, Sen = 0.808, Spe = 0.896, F1 = 0.877, Kappa = 0.841; APTOS-2019: Acc = 0.855, Sen = 0.810, Spe = 0.904, F1 = 0.859, Kappa = 0.872; Clinical dataset: Acc = 0.770, Sen = 0.770, Spe = 0.942, F1 = 0.770, Kappa = 0.877 |
| Swin-MCSFNet [136] | Classification | Fundus image quality | Removal of non-retinal borders; image resizing to 256 × 256; random vertical flipping; random horizontal flipping; random rotation from −180° to 180°; intensity normalization | EyeQ; OIA-ODIR; EyePACS | EyeQ: Acc = 0.877, F1 = 0.776, Precision = 0.792, Recall = 0.766, AUC (micro) = 0.930, AUC (Good) = 0.960, AUC (Usable) = 0.810, AUC (Reject) = 0.960; OIA-ODIR: Acc = 0.798, F1 = 0.644, Precision = 0.639, Recall = 0.665; EyePACS: Acc = 0.749, F1 = 0.627, Precision = 0.616, Recall = 0.664 |
| DR-MobileViT [190] | Classification | DR | Image resizing to 224 × 224; normalization to [0,1]; CLAHE; random rotation within ±30°; horizontal flipping; brightness adjustment | APTOS-2019; MESSIDOR-2 | APTOS-2019: Sen = 0.920, Spe = 0.880, AUC = 0.950, QWK = 0.890; MESSIDOR-2: Sen = 0.890, Spe = 0.860, AUC = 0.930, QWK = 0.870 |
| DR-EfficientFormer [190] | Classification | DR | Image resizing to 224 × 224; normalization to [0,1]; CLAHE; random rotation within ±30°; horizontal flipping; brightness adjustment | APTOS-2019; MESSIDOR-2 | APTOS-2019: Sen = 0.900, Spe = 0.870, AUC = 0.940, QWK = 0.870; MESSIDOR-2: Sen = 0.870, Spe = 0.850, AUC = 0.920, QWK = 0.850 |
| DR-SwinTiny [190] | Classification | DR | Image resizing to 224 × 224; normalization to [0,1]; CLAHE; random rotation within ±30°; horizontal flipping; brightness adjustment | APTOS-2019; MESSIDOR-2 | APTOS-2019: Sen = 0.910, Spe = 0.890, AUC = 0.950, QWK = 0.880; MESSIDOR-2: Sen = 0.880, Spe = 0.870, AUC = 0.930, QWK = 0.860 |
| Swin Transformer with Multi-Wavelet [291] | Classification | DR | Data augmentation; image resizing; automatic cropping | APTOS-2019 | Binary classification: Acc = 0.970, Sen = 0.952, Spe = 0.952, F1 = 0.987; Five-class classification: Acc = 0.820, Sen = 0.802, Spe = 0.902 |
| TCDDU-Net [178] | Segmentation | Retinal vessels | Grayscale conversion; zero-mean normalization; CLAHE; gamma correction; normalization by division by 255; random cropping into 48 × 48 patches | DRIVE; STARE; CHASE_DB1 | DRIVE: Acc = 0.970, Sen = 0.826, Spe = 0.984, AUC = 0.987, F1 = 0.827; STARE: Acc = 0.974, Sen = 0.792, Spe = 0.988, AUC = 0.986, F1 = 0.816; CHASE_DB1: Acc = 0.972, Sen = 0.819, Spe = 0.982, AUC = 0.985, F1 = 0.784 |
| SURVS [292] | Segmentation | Retinal vessels | Grayscale conversion; gamma correction; Laplacian-of-Gaussian filtering; Otsu thresholding; morphological opening and closing; dataset-specific cropping, padding, and resizing | DRIVE; STARE; CHASE_DB1 | DRIVE: Acc = 0.967, Sen = 0.855, Spe = 0.976; STARE: Acc = 0.967, Sen = 0.858, Spe = 0.974; CHASE_DB1: Acc = 0.957, Sen = 0.845, Spe = 0.967 |
| TD Swin-UNet [177] | Segmentation | Retinal vessels | Image resizing to 448 × 448; CLAHE; gamma correction; normalization to [-1,1]; random horizontal and vertical flipping | DRIVE; CHASE_DB1 | DRIVE: Acc = 0.966, Sen = 0.848, Spe = 0.984, F1 = 0.865; CHASE_DB1: Acc = 0.976, Sen = 0.852, Spe = 0.987, F1 = 0.852 |

Table 12: Representation Learning Methods with Visual Unimodal Data (continued)

| Method | Task | Target | Preprocessing | Dataset | Performance |
|---|---|---|---|---|---|
| PCAT-UNet [191] | Segmentation | Retinal vessels | Uniform resizing to 512 × 512; field-of-view masking; histogram stretching; Gaussian transformation; random rotation; random horizontal flipping; gamma-based contrast enhancement | DRIVE; STARE; CHASE_DB1 | DRIVE: Acc = 0.962, Sen = 0.858, Spe = 0.993, AUC = 0.987, F1 = 0.816; STARE: Acc = 0.980, Sen = 0.870, Spe = 0.994, AUC = 0.995, F1 = 0.884; CHASE_DB1: Acc = 0.981, Sen = 0.849, Spe = 0.997, AUC = 0.993, F1 = 0.827 |
| SGAT-Net [293] | Segmentation | Retinal vessels | Image resizing to 2048 × 2048; partition into 512 × 512 patches; grayscale conversion and normalization; CLAHE; gamma correction; flipping; rotation; translation; scaling; additive Gaussian noise; random brightness adjustment | FIVES; DRIVE; STARE; CHASE_DB1 | FIVES: Acc = 0.989, AUC = 0.947, F1 = 0.905, IoU = 0.835, Sen = 0.916, Spe = 0.993; DRIVE: Acc = 0.966, AUC = 0.911, F1 = 0.833, Sen = 0.863, Spe = 0.977; STARE: Acc = 0.976, F1 = 0.851, Sen = 0.893, Spe = 0.983; CHASE_DB1: Acc = 0.978, F1 = 0.849, Sen = 0.870, Spe = 0.986 |
| HT-Net [192] | Segmentation | Retinal vessels | Dataset-specific augmentation using random rotation, color jittering, additive Gaussian noise, and flipping; dataset-specific cropping and resizing | DRIVE; CHASE_DB1; STARE | DRIVE: Acc = 0.970, Sen = 0.826, Spe = 0.984, AUC = 0.987, F1 = 0.828; CHASE_DB1: Acc = 0.975, Sen = 0.848, Spe = 0.983, AUC = 0.990, F1 = 0.815; STARE: Acc = 0.977, Sen = 0.848, Spe = 0.987, AUC = 0.993, F1 = 0.846 |
| CoVi-Net [193] | Segmentation | Retinal vessels | Reshaping to 512 × 512 during training; random rotation; horizontal flipping; vertical flipping; random erasing; additive Gaussian noise; gamma correction; histogram equalization; image sharpening; color, contrast, and brightness enhancement | DRIVE; CHASE_DB1; STARE | DRIVE: Acc = 0.970, Sen = 0.835, Spe = 0.983, AUC = 0.988, F1 = 0.828, F = 0.875, Ccon = 0.999, A = 0.956, L = 0.916; CHASE_DB1: Acc = 0.976, Sen = 0.845, Spe = 0.984, AUC = 0.990, F = 0.835, Ccon = 0.999, A = 0.929, L = 0.900; STARE: Acc = 0.976, Sen = 0.831, Spe = 0.989, AUC = 0.992, F = 0.903, Ccon = 0.999, A = 0.968, L = 0.934 |
| RetinaLiteNet [294] | Segmentation | Retinal vessels and optic disc | Extensive augmentation including dropout, additive white noise, gamma correction, histogram equalization, blurring, elastic deformation, flipping, shearing, sharpening, saturation enhancement, rotation, compression simulation, zooming, contrast enhancement, shifting, random cropping, padding, grid distortion, optical distortion, CLAHE, and random brightness adjustment; image resizing to 512 × 512 | DRIVE; IOSTAR | DRIVE-vessel: Sen = 0.784, Spe = 0.980, AUC = 0.970, F1 = 0.806, IoU = 0.675; DRIVE-optic disc: Sen = 0.940, Spe = 0.970, AUC = 0.990, F1 = 0.933, IoU = 0.880; IOSTAR-vessel: Sen = 0.775, Spe = 0.972, AUC = 0.980, F1 = 0.801, IoU = 0.676; IOSTAR-optic disc: Sen = 0.765, Spe = 0.999, AUC = 0.990, F1 = 0.854, IoU = 0.749 |
| EfficientNet-Swin Transformer Hybrid Network [295] | Classification | DR | Image cropping; Gaussian smoothing; color-space cropping with Gaussian smoothing; automatic cropping; center cropping; random rotation; translation; flipping; scaling; random cropping; brightness and contrast adjustment; resizing to 256 × 256 × 3 followed by center cropping to 224 × 224 × 3 | APTOS-2019 | Acc = 0.970, Sen = 0.950, Spe = 0.980, AUC = 0.970, F1 = 0.960, Precision = 0.980 |
| CNN-Transformer Fusion Network [194] | Classification | DR | Removal of low-quality and noisy images; resizing to 224 × 224; unified color format; rotation; scaling; random cropping; brightness adjustment; noise perturbation; Gaussian filtering; histogram equalization; intensity thresholding; morphological processing; connected-component analysis; region alignment to 7 × 7 | APTOS-2019; IDRiD | APTOS-2019: Acc = 0.943, Sen = 0.910, Spe = 0.958, AUC = 0.964, F1 = 0.924; IDRiD: Acc = 0.959, Sen = 0.939, Spe = 0.977, AUC = 0.958, F1 = 0.957 |
| RETFound [54] | Classification / Prediction | Ophthalmic and systemic diseases | NR | APTOS-2019; IDRiD; MESSIDOR-2; Glaucoma Fundus; PAPILA; Retina; JSIEC; OCTID; UK Biobank; MEH-AlzEye | DR classification: APTOS-2019 (AUC = 0.943), IDRiD (AUC = 0.822), MESSIDOR-2 (AUC = 0.884); Wet AMD 1-year conversion prediction: CFP (AUC = 0.862), OCT (AUC = 0.799); 3-year systemic disease prediction: myocardial infarction (CFP AUC = 0.737), heart failure (CFP AUC = 0.794), ischemic stroke (CFP AUC = 0.754), Parkinson's disease (CFP AUC = 0.669) |
| GlanceSeg [296] | Segmentation | Microaneurysms | Gaze-area generation using Gaussian function; binary attention-map generation for ROI extraction; ROI cropping and enhancement; fusion of feature texture map and morphological boundary descriptor saliency map; dataset-specific resizing | IDRiD; Retinal-Lesions | IDRiD: zero-shot (DSC = 0.368, AUPR = 0.552), fine-tuning (DSC = 0.394, AUPR = 0.571); Retinal-Lesions: AUPR = 0.682 |

Table 12: Representation Learning Methods with Visual Unimodal Data (continued)

| Method | Task | Target | Preprocessing | Dataset | Performance |
|---|---|---|---|---|---|
| FunduSAM [137] | Segmentation | Optic disc and optic cup | Image centering; optic-disc ROI cropping; alignment; polar coordinate transformation; inverse polar-coordinate transformation of output mask | REFUGE | REFUGE: Optic disc (DSC = 0.961, IoU = 0.882), optic cup (DSC = 0.867, IoU = 0.789) |
| Serp-Mamba [120] | Segmentation | Retinal vessels | Exclusion of invalid regions using official masks; uniform resizing to $1024 \times 1024$ | PRIME-FP20; MU-VS Center A; MU-VS Center B | PRIME-FP20: DSC = 0.690, IoU = 0.529, MCC = 0.684, BM = 0.658; MU-VS Center A: DSC = 0.611, IoU = 0.440, MCC = 0.601, BM = 0.584; MU-VS Center B: DSC = 0.571, IoU = 0.401, MCC = 0.563, BM = 0.539 |
| HM-Mamba [182] | Segmentation | Retinal vessels | Overlapping tile-based patch extraction with patch size $256 \times 256$ and stride 128; color perturbation; random horizontal flipping; random vertical flipping; random rotation | DRIVE; CHASE_DB1; STARE; IOSTAR; LES-AV | DRIVE: Acc = 0.975, Sen = 0.832, Spe = 0.990, AUC = 0.913, F1 = 0.833, IoU = 0.716; CHASE_DB1: Acc = 0.976, Sen = 0.821, Spe = 0.987, AUC = 0.909, F1 = 0.820, IoU = 0.684; STARE: Acc = 0.979, Sen = 0.815, Spe = 0.986, AUC = 0.900, F1 = 0.824, IoU = 0.689; IOSTAR: Acc = 0.978, Sen = 0.839, Spe = 0.985, AUC = 0.915, F1 = 0.831, IoU = 0.712; LES-AV: Acc = 0.982, Sen = 0.830, Spe = 0.993, AUC = 0.914, F1 = 0.843, IoU = 0.729 |
| CRMA-UNet [297] | Segmentation | Retinal vessels | NR | DRIVE; STARE; CHASE_DB1; HRF; IOSTAR | DRIVE: Acc = 0.972, Sen = 0.846, Spe = 0.989, F1 = 0.839, MSE = 0.031; STARE: Acc = 0.974, Sen = 0.885, Spe = 0.983, F1 = 0.833, MSE = 0.032; CHASE_DB1: Acc = 0.981, Sen = 0.860, Spe = 0.991, F1 = 0.859, MSE = 0.012; HRF: Acc = 0.976, Sen = 0.859, Spe = 0.993, F1 = 0.862; IOSTAR: Acc = 0.976, Sen = 0.894, Spe = 0.987, F1 = 0.847 |
| MM-UNet [298] | Segmentation | Retinal vessels | Dataset-specific resizing | DRIVE; STARE | DRIVE: Acc = 0.983, Sen = 0.893, Spe = 0.991, F1 = 0.896; STARE: Acc = 0.988, Sen = 0.918, Spe = 0.994, F1 = 0.918 |
| HCM [196] | Segmentation | Retinal vessels | Cropping into $224 \times 224$ patches; random horizontal flipping; random vertical flipping; additive Gaussian noise; grid distortion; optical distortion | DRIVE; CHASE_DB1; STARE | DRIVE: Sen = 0.822, Spe = 0.985, DSC = 0.825, IoU = 0.702; CHASE_DB1: Sen = 0.852, Spe = 0.985, DSC = 0.814, IoU = 0.686; STARE: Sen = 0.783, Spe = 0.989, DSC = 0.805, IoU = 0.676 |

*Abbreviations:* Acc, accuracy; AUC, area under the receiver operating characteristic curve; AUPR, area under the precision-recall curve; Sen, sensitivity; Spe, specificity; F1, F1-score; Precision, positive predictive value; Recall, recall; IoU, intersection over union; DSC, Dice similarity coefficient; QWK, quadratic weighted kappa; MCC, Matthews correlation coefficient; MSE, mean squared error; NR, not reported.

*Notes:* Results are reported in a unified format for consistency across studies. Only the metrics explicitly provided in the source table are retained and standardized.

## K. Multimodal Fusion Modeling Methods with Multimodal Data

Table 13: Multimodal Fusion Modeling Methods with Multimodal Data

| Method | Task | Target | Preprocessing | Dataset | Performance |
| --- | --- | --- | --- | --- | --- |
| FLAIR [138] | Detection / Classification | Retinal pathology and disease categories | Image resizing to 512 × 512; zero padding for rectangular images; intensity normalization to [0, 1]; data augmentation using horizontal flipping, random rotation within [-5°, 5°], scaling within [0.9, 1.1], and color jittering | MESSIDOR; FIVES; REFUGE; 20×3; ODIR200×3; MMAC; ACRIMA; DeepDRiD; MMAC-B; RP-MHL-2; CAT-MYA-2 | Domain-shift zero-shot: MESSIDOR (ACA = 0.604, Kappa = 0.772), FIVES (ACA = 0.735), REFUGE (ACA = 0.920); unseen-category zero-shot: 20×3 (ACA = 0.983), ODIR200×3 (ACA = 0.667), MMAC (ACA = 0.400); linear-probe ACA: MESSIDOR = 0.719, FIVES = 0.879, REFUGE = 0.843, 20×3 = 1.000, ODIR = 0.935, MMAC = 0.740, average = 0.852 |
| RET-CLIP [202] | Classification | DR, glaucoma, and ocular diseases | Image resizing to 512 × 512; random cropping followed by resizing to 224 × 224; random horizontal flipping; color jittering; intensity normalization; text correction including punctuation normalization, abbreviation restoration, language standardization, and manual grammatical revision | IDRiD; APTOS-2019; PAPILA; Glaucoma Fundus; JSIEC; Retina; RFMID; ODIR | IDRiD: AUC = 0.863, AUPR = 0.630; APTOS-2019: AUC = 0.951, AUPR = 0.748; PAPILA: AUC = 0.853, AUPR = 0.754; Glaucoma Fundus: AUC = 0.958, AUPR = 0.889; JSIEC: AUC = 0.999, AUPR = 0.972; Retina: AUC = 0.942, AUPR = 0.871; RFMID: AUC = 0.946, AUPR = 0.581; ODIR: AUC = 0.917, AUPR = 0.715 |
| EyeCLIP [203] | Classification / Retrieval / Visual Question Answering | Ocular and systemic diseases | Field-of-view-based cropping followed by resizing to 224 × 224; random resized cropping; color jittering; horizontal flipping; pretraining data quality control based on separable-vessel ratio thresholds; hierarchical medical concept extraction from reports using keyword mapping and regular-expression parsing | IDRiD; APTOS-2019; MESSIDOR-2; PAPILA; Glaucoma Fundus; JSIEC; Retina; OCTID; OCTDL; AngioReport; Retina Image Bank; OphthalVQA; UK Biobank | Zero-shot classification: DR (AUC = 0.681–0.757), glaucoma (AUC = 0.721, AUPR = 0.684), multi-disease (AUC = 0.660, AUPR = 0.688), OCTID (AUC = 0.800), OCTDL (AUC = 0.776); systemic disease prediction: stroke (AUC = 0.641, AUPR = 0.627), dementia (AUC = 0.536, AUPR = 0.572), PD (AUC = 0.580, AUPR = 0.616), MI (AUC = 0.596, AUPR = 0.582); cross-modal retrieval: AngioReport (t2i = 0.441, i2i = 0.407, i2t = 0.443), Retina Image Bank (t2i = 0.502, i2i = 0.433, i2t = 0.509) |
| RetiZero [118] | Classification / Retrieval | Common and rare fundus diseases | Text standardization to a unified prompt template; retention of only text precisely matched to color fundus photographs; exclusion of blurred, overexposed, underexposed, montage, local-field, monochromatic, and externally annotated images; label denoising | EYE-15; EYE-52; H1; H2; H3; rH1; rH2; rH3; SIC; SUSTech; SiMES; SINDI | Zero-shot classification: EYE-15 (Top-1/Top-3/Top-5 = 0.442/0.702/0.840), EYE-52 (Top-1/Top-3/Top-5 = 0.360/0.626/0.756); retrieval: EYE-15 (Top-1/Top-3/Top-5 = 0.854/0.928/0.950), EYE-52 (Top-1/Top-3/Top-5 = 0.726/0.843/0.886); internal-domain AUC: H1/H2/H3 = 0.997/0.980/0.993; few-shot AUC = 0.967/0.859/0.942; cross-domain internal AUC: rH1/rH2/rH3 = 0.998/0.986/0.990; external test AUC ≥ 0.912; clinical assistance accuracy: 0.552–0.628 |
| Meta-EyeFM [207] | Multitask | Ocular and systemic diseases with related fundus signs | NR | SEED; external datasets | Router: Acc = 0.968, Macro F1 = 0.921; disease detection AUC: DR = 0.974, AMD = 0.912, MMD = 0.988, glaucoma = 0.942, cataract = 0.939, diabetes = 0.808, hypertension = 0.798, CKD = 0.848; severity grading AUC: mild NPDR = 0.820, moderate/severe NPDR = 0.964, PDR = 0.989, early AMD = 0.916, late AMD = 0.992, MMD Cat1 = 0.954, Cat2 = 0.986, Cat3/4 = 0.887, cataract = 0.917, visually significant cataract = 0.960; sign recognition AUC: microaneurysm = 0.820, laser scar = 0.990, hard exudates = 0.956, soft exudates = 0.934, macular edema = 0.987, drusen = 0.779, pigmentary abnormalities = 0.893, tessellation = 0.953, diffuse chorioretinal atrophy = 0.971 |
| MDF-MLLM [299] | Classification | Acquired and inherited retinal diseases | Image resizing to 3 × 512 × 512; element-wise normalization by division by 255; watermark-region processing using grayscale conversion, threshold segmentation, morphological dilation, mask removal, and fast marching inpainting; unified prompt initialization | Hold-out test set | Acc = 0.940; acquired diseases: Precision = 1.000, Recall = 0.920, F1 = 0.960; inherited diseases: Precision = 1.000, Recall = 0.970, F1 = 0.990 |

Table 13: Multimodal Fusion Modeling Methods with Multimodal Data (continued)

| Method | Task | Target | Preprocessing | Dataset | Performance |
|---|---|---|---|---|---|
| DeepDR Plus [139] | Prediction | DR progression risk | Exclusion of ungradable images; image resizing to 512 × 512 for pretraining and fine-tuning; extraction of two 512 × 512 crops per iteration; data augmentation using random compression, random blurring, brightness jittering, contrast jittering, random gamma correction, additive Gaussian noise, and random rotation | DRPS; ECHM; WTHM; NDSP; CUHK-STDR; PUDM; SEED; SiDRP; BJHC | Internal validation: fundus model (C-index = 0.823), combined model (C-index = 0.833), metadata model (C-index = 0.696); external cohorts: fundus model (C-index = 0.786–0.802); 1-year progression prediction: internal fundus model (C-index = 0.823–0.862, IBS = 0.049–0.161), external cohorts (C-index = 0.804–0.837, IBS = 0.066–0.170); 1-year AUC: internal = 0.826–0.865, external = 0.722–0.863; $R^2 = 0.678$; mean individualized screening interval = 31.970 months; screening frequency reduction = 0.625; VTDR delayed detection rate = 0.002 |
| Multimodal Bi-LSTM-ConvNeXt-V2 [204] | Prediction | Glaucoma progression | Visual field and retinal nerve fiber layer image resizing to 224 × 224; normalization using ImageNet mean and standard deviation; red-green-blue image storage; intensity scaling; one-hot encoding of race; zero-mean unit-variance normalization for age, mean deviation, and retinal nerve fiber layer thickness; nearest-neighbor temporal matching within a ±6-month tolerance for missing visits | Five-fold cross-validation test fold | Acc = 0.947, AUC = 0.975 ± 0.015, F1 = 0.887 |
| MLPResNet50 [117] | Classification | DR | Image preprocessing using random cropping with variable scale and aspect ratio, resizing to 448 × 448, random horizontal flipping, color jittering, and normalization; tabular preprocessing using last-observation-carried-forward imputation, global median imputation, zero filling for binary features, separate encoding for missing categories, removal of features with missing rate greater than 90%, standardization of numerical features, and ordinal encoding of categorical features; weighted random sampling during training | EVIRED; BRset; mBRset | EVIRED best fusion: Acc = 0.634, one-vs-rest AUC = 0.892/0.786/0.803/0.869/0.578/0.986, Weighted F1 = 0.626, Kappa = 0.807; BRset: Acc = 0.659, AUC = 0.912/0.836/0.777/0.910/0.913, Weighted F1 = 0.659, Kappa = 0.764; mBRset: Acc = 0.632, AUC = 0.850/0.683/0.843/0.969/0.958, Weighted F1 = 0.601, Kappa = 0.771 |
| Multimodal LBP-LSTM [205] | Detection / Classification | DR | Local binary pattern texture extraction from fundus and optical coherence tomography images; data augmentation for all modalities; feature-dimension normalization to 2000; multimodal feature concatenation with electronic health record features | NR | Acc = 0.960, AUC = 0.990, Sen = 0.950, Spe = 0.970 |
| FedMHA-DR [140] | Classification | DR | Image resizing to 224 × 224; minority-class augmentation using rotation, flipping, and mild color jittering; additional augmentation during training using rotation within ±15°, flipping, and mild brightness and contrast adjustment; class-weighted loss; per-round feature regeneration and normalization for electronic health record variables | RetinaMNIST | AUC = 0.764, F1 = 0.711, Precision = 0.705, Recall = 0.718 |
| Time-Series VLM for DR Forecasting [141] | Prediction | Referable DR or maculopathy progression | Use of current macula-centered fundus image; automatic generation of twenty semantically equivalent prompt templates; random sampling of historical sequence length; appending prior best-corrected visual acuity and previous diabetic retinopathy grade when available | Internal hold-out test set | 1-year: AUC = 0.691, Acc = 0.631, F1 = 0.365, Sen = 0.725, Spe = 0.535; 2-year: AUC = 0.674, Acc = 0.611, F1 = 0.357, Sen = 0.609, Spe = 0.591; 3-year: AUC = 0.671, Acc = 0.610, F1 = 0.382, Sen = 0.595, Spe = 0.543 |
| TV-LSTM [214] | Prediction | Late AMD progression | Selection of AREDS Field 2 images; random selection of the left image from stereo pairs; exclusion of ungradable images; cropping of macula-centered square regions; resizing to 224 × 224; recoding of the age-related macular degeneration severity scale into a binary outcome; additive encoding of fifty-two single-nucleotide polymorphisms by minor allele count | AREDS | AUC-ROC = 0.948, AUC-PR = 0.859 |

Table 13: Multimodal Fusion Modeling Methods with Multimodal Data (continued)

| Method | Task | Target | Preprocessing | Dataset | Performance |
| --- | --- | --- | --- | --- | --- |
| TIMM-ProRS [212] | Classification / Prediction | DR severity and 5-year progression risk | Contrast enhancement using CLAHE; Gaussian denoising; image resizing to 224 × 224; min-max normalization of biomarkers; data augmentation using rotation within ±15°, flipping, and brightness variation within ±20%; exclusion of blurred images using variance-of-Laplacian thresholding | APTOS-2019; Messidor-2; RFMiD; EyePACS; Messidor-1 | Classification: APTOS-2019 (AUC = 0.970, Acc = 0.978, F1 = 0.960, Sen = 0.989, Spe = 0.993, QWK = 0.950), Messidor-2 (AUC = 0.950, Acc = 0.964, F1 = 0.950, Sen = 0.976, Spe = 0.989, QWK = 0.930), EyePACS (AUC = 0.960, Acc = 0.967, F1 = 0.960, Sen = 0.981, Spe = 0.990, QWK = 0.940), Messidor-1 (AUC = 0.940, Acc = 0.962, F1 = 0.950, Sen = 0.973, Spe = 0.987, QWK = 0.920), RFMiD (AUC = 0.970, Acc = 0.972, F1 = 0.960, Sen = 0.980, Spe = 0.985, QWK = 0.950); progression prediction: C-index = 0.890, Brier score = 0.100 |
| Fundus-CRF Multimodal Fusion Model [213] | Classification / Prediction | Cardiovascular disease risk | Exclusion of images with retinal or vitreous pathology and low-quality images; conversion of full-resolution images to JPEG format; center cropping; random horizontal and vertical flipping; random rotation within 30°; brightness adjustment; scaling augmentation; resizing to 448 × 448; normalization of age, systolic blood pressure, total cholesterol, and high-density lipoprotein | SMC; UK Biobank | Internal validation: AUC = 0.781, Acc = 0.683, Sen = 0.871, Spe = 0.508, PPV = 0.621, NPV = 0.810; external validation: AUC = 0.872; external validation after training on full SMC: AUC = 0.882; external validation with uncertainty quantification: AUC = 0.905; future event association: HR = 6.280 |

*Abbreviations:* Acc, accuracy; AUC, area under the receiver operating characteristic curve; AUPR, area under the precision-recall curve; Sen, sensitivity; Spe, specificity; Precision, positive predictive value when reported as precision in the original study; Recall, recall; F1, F1-score; IoU, intersection over union; DSC, Dice similarity coefficient; Kappa, Cohen's kappa; QWK, quadratic weighted kappa; C-index, concordance index; IBS, integrated Brier score; HR, hazard ratio; PPV, positive predictive value; NPV, negative predictive value; NR, not reported; DR, diabetic retinopathy; AMD, age-related macular degeneration; MMD, myopic macular degeneration; CKD, chronic kidney disease; PD, Parkinson's disease; MI, myocardial infarction; CLAHE, contrast-limited adaptive histogram equalization.
*Notes:* Results are presented in a unified format for cross-study comparison. Only representative metrics are retained for methods with extensive multi-task evaluations. NR indicates that the corresponding information was not reported in the original source.